\documentclass[11pt,english]{article}
\usepackage[T1]{fontenc}
\usepackage[utf8]{inputenc}
\usepackage{geometry}
\usepackage{microtype}
\usepackage{graphicx}
\usepackage{subfigure}
\usepackage{booktabs} % for professional tables
\usepackage{stmaryrd}
\usepackage{hyperref}

\usepackage{amsmath}
\usepackage{amssymb}
\usepackage{mathtools}
\usepackage{amsthm}
\usepackage{bm}
\usepackage{natbib}

\renewcommand{\cite}[1]{\citep{#1}}
\usepackage[capitalize,noabbrev]{cleveref}

\theoremstyle{plain}
\newtheorem{theorem}{Theorem}[section]

\newtheorem{lemma}[theorem]{Lemma}
\newtheorem{corollary}[theorem]{Corollary}
\theoremstyle{definition}

\theoremstyle{remark}

\usepackage[textsize=tiny]{todonotes}
\RequirePackage{algorithm}
\RequirePackage{algorithmic}

\newcommand{\pr}{\mathbb{P}}

\newcommand{\ba}{\bm{a}}

\newcommand{\bX}{\bm{X}}
\newcommand{\bA}{\bm{A}}
\newcommand{\bY}{\bm{Y}}

\newcommand{\bbA}{\mathbb{A}}

\newcommand{\bV}{\mathbf{V}}

\newcommand{\bu}{\bm{u}}

\newcommand{\T}{\mathrm{\scriptscriptstyle T}}

\newcommand{\E}{\mathbb{E}}
\newcommand{\ind}{\mbox{$\perp\!\!\!\perp$}}
\begin{document}
	% It is OKAY to include author information, even for blind
	% submissions: the style file will automatically remove it for you
	% unless you've provided the [accepted] option to the icml2024
	% package.
	
	% List of affiliations: The first argument should be a (short)
	% identifier you will use later to specify author affiliations
	% Academic affiliations should list Department, University, City, Region, Country
	% Industry affiliations should list Company, City, Region, Country
	
	% You can specify symbols, otherwise they are numbered in order.
	% Ideally, you should not use this facility. Affiliations will be numbered
	% in order of appearance and this is the preferred way.
	% \icmlsetsymbol{equal}{*}
	\graphicspath{{figures}}
	
	\title{Causal Customer Churn Analysis with Low-rank Tensor Block Hazard Model}

	\author{Chenyin Gao\thanks{ Department of Statistics, North Carolina State University}, Zhiming Zhang\thanks{Independent scholar, Iowa State University}, and Shu Yang$^*$\thanks{Corresponding author: syang24@ncsu.edu,  Department of Statistics, North Carolina State University, Raleigh, NC 27695, U.S.A.}}
	
	% \icmlcorrespondingauthor{Firstname2 Lastname2}{first2.last2@www.uk}
	
	% You may provide any keywords that you
	% find helpful for describing your paper; these are used to populate
	% the "keywords" metadata in the PDF but will not be shown in the document
	\date{}
	\maketitle

	% this must go after the closing bracket ] following \twocolumn[ ...
	
	% This command actually creates the footnote in the first column
	% listing the affiliations and the copyright notice.
	% The command takes one argument, which is text to display at the start of the footnote.
	% The \icmlEqualContribution command is standard text for equal contribution.
	% Remove it (just {}) if you do not need this facility.
	
	% \printAffiliationsAndNotice{}  % leave blank if no need to mention equal contribution
	% \printAffiliationsAndNotice{\icmlEqualContribution} % otherwise use the standard text.

	\begin{abstract}
		This study introduces an innovative method for analyzing the impact of various interventions on customer churn, using the potential outcomes framework. We present a new causal model, the tensorized latent factor block hazard model, which incorporates tensor completion methods for a principled causal analysis of customer churn. A crucial element of our approach is the formulation of a 1-bit tensor completion for the parameter tensor. This captures hidden customer characteristics and temporal elements from churn records, effectively addressing the binary nature of churn data and its time-monotonic trends. Our model also uniquely categorizes interventions by their similar impacts, enhancing the precision and practicality of implementing customer retention strategies. For computational efficiency, we apply a projected gradient descent algorithm combined with spectral clustering. We lay down the theoretical groundwork for our model, including its non-asymptotic properties. The efficacy and superiority of our model are further validated through comprehensive experiments on both simulated and real-world applications.
		
	\end{abstract}
	
	\section{Introduction}
	
	Customer retention, loyalty, and churn are increasingly important
	topics across industries. Customer churn, or attrition, occurs when
	customers disengage from a company by ending subscriptions, moving
	to competitors, or changing their buying habits \citep{buckinx2007predicting}.
	Analyzing and preventing churn is critical for sustainable growth
	and profitability. Companies often explore diverse retention strategies,
	such as customized incentives, to extend customer lifetimes. Understanding the causality behind customer churn helps companies adopt proper retention strategies and essentially increase the customer lifetime value and the company's business value. 
	
	In our approach to analyzing causal churn, we define potential outcomes
	as the indicators of churn over time under different
	levels of treatment. Thus, the potential outcomes can be envisioned
	as a three-dimensional tensor, indexed by the customer, time, and intervention,
	where each entry represents whether a customer would churn at a particular
	time under a specific treatment (see Figure \ref{fig:binary-tensor}).
	In practice, only actual churn trajectories are observed. To address
	this, tensor completion methods can be employed to fill in all missing
	potential outcomes and facilitate the analysis of various treatment
	strategies. Tensors effectively uncover the hidden multiway data structure,
	often using low-rankness, which decomposes the tensor into a low-dimensional
	core tensor and matrix factors for each dimension. However, applying
	these existing tensor completion methods presents several challenges
	in our context. 
	
	Firstly, unlike tensors with continuous values, our churn analysis
	tensor is binary, with entries as $0/1$ indicators. Moreover, once
	a customer churns at a certain time, it remains churned, resulting in
	a monotone churn pattern. Therefore, applying low-rankness to the
	original potential outcome tensor may not be suitable in this case.
	Secondly, many retention interventions might have similar effects
	on customers, suggesting a potential benefit in grouping these interventions
	for more accurate estimation and streamlined implementation in practice.
	Previous research has proposed integrating interventions based on
	prior knowledge \citep{laber2014set,liu2018augmented,pan2021improved}.
	However, a data-driven approach to identify and cluster interventions
	with similar effects is desirable. For instance, \citet{ma2022learning}
	introduced a method using adaptive fusion penalty for clustering interventions,
	but it is limited to single-stage outcomes and parametric treatment
	effect models. Exploring and autonomously grouping interventions with
	similar effects over time to streamline the dimension space is therefore
	a significant area of interest. Finally, a major challenge arises
	from the uniform missing mechanisms in current matrix/tensor completion
	methods, which fail to consider the endogeneity in treatment assignment.
	As a result, directly applying these methods could lead to confounding
	biases.
	
	We develop a novel tensorized latent factor block hazard model in
	(\ref{eq:tensorM}) to conduct causal analysis on customer churn trajectories
	in relation to various treatment strategies. This model redefines
	the problem into a 1-bit tensor completion problem for the parameter
	tensor, leveraging structural information, such as low rankness and
	clustering blocks. In particular, it captures the customer attributes
	and temporal features by the latent factors inferred from the churn
	history. Besides, a block structure is adopted for the interventions
	based on their impact on the churn statuses to reduce the number of
	interventions. This data-driven clustering allows us to automatically
	identify the homogeneous intervention groups. To our knowledge, little work has explored the causality of customer churn analysis with grouped latent factors and we are among the first to provide non-asymptotic error analysis for this low-rank tensor block hazard model. 
	
	Our contributions are summarized as follows: 
	\begin{itemize}
		\item [1)] Tensorized Latent Factor Block Hazard Model: We introduce a model
		applying low-rankness to the parameter tensor (rather than the data
		tensor) to leverage latent unit and temporal factors and identify
		homogenous intervention groups more effectively.
		\item [2)] Computational Methodology: We use a projected gradient descent
		algorithm with spectral clustering to solve the inverse probability
		treatment weighted (IPTW) loss, adjusting for confounding effects
		and scaling well to large datasets.
		\item [3)] Optimal Treatment Search and Learning: The proposed framework provide
		the survival probabilities under all interventions and thus enables
		one to identify the individual optimal interventions that maximize retention
		time, closely related to the optimal policy search and Q-learning in
		causal inference \citep{qian2011performance}; and
		\item [4)] Theoretical Underpinnings and Empirical Evidence: We have established the non-asymptotic
		properties of our proposed model, including the upper bound on the
		tensor recovery accuracy and the clustering misclassification rate. Besides, we demonstrate the practical benefits and effectiveness of our framework via comprehensive synthetic experiments
		and a real-data application. Our implementation codes will be made publicly available after the acceptance of this manuscript.
	\end{itemize}
	
	\section{Related Work}
	Customer churn prediction can be naturally perceived as a classification task. With the rise of machine learning algorithms, various methods have been proposed for churn analysis, including support vector machines \citep{coussement2008churn}, random forest \citep{xie2009customer}, and other ensemble methods such as bagging and boosting \citep{lu2012customer}. Deep neural networks (DNN) have also been employed to extract valuable features related to customer churn, which significantly improve the performance on real-world datasets \citep{mishra2017novel, zhang2017deep,umayaparvathi2017automated, rudd2021causal}. However, these classification models mainly focus on determining churn status at a fixed time point, often overlooking the information contained in time to churn. 
	
	Considering customer
	lifetime as a time-to-churn metric and using survival analysis offers deeper insights into customer-company engagement \citep{lu2002predicting,lariviere2004investigating}. Traditional
	survival analysis, though, relies on potentially restrictive assumptions
	about hazard functions. Advanced survival models like survival random forests
	\citep{ishwaran2008random}, Cox boosting
	\citep{binder2009boosting}, survival Super-Learner \citep{van2011targeted}, and time-to-event reinforcement learning \citep{maystre2022temporally} have been developed to better handle more complex data. Similarly, DNNs have also been utilized for survival analysis to handle the special loss function induced by the censored data \citep{zhu2016deep, katzman2018deepsurv, ching2018cox, zhao2020deep}. However, these models may not accurately reflect the causal impact of interventions on churn
	due to confounding biases \citep{yang2018semiparametricAOS}. For example, if incentives are offered only to at-risk customers, it may be falsely ineffective as this group naturally shows higher retention and shorter churn times compared to others. Therefore, a more principled approach is necessary for the causal analysis of retention interventions.
	
	We use the potential outcomes framework of different
	intervention strategies for churn analysis. Here, the potential outcome is defined as the potential outcome (possibly contrary to fact) had the unit (customer)
	received a specific treatment (intervention). The fundamental problem
	of causal inference is that each unit receives only one treatment,
	leaving other potential outcomes unknown. Thus, the causal analysis
	problem is essentially a missing data problem. To address this, matrix
	or tensor completion techniques \citep{davenport20141,mao2023matrix}, commonly used for filling in missing data, are applicable. Tensor completion problem is initially addressed by unfolding tensors into matrices \citep{tomioka2010estimation, gandy2011tensor, liu2012tensor}. However, such unfolding-based methods might discard the multi-way structure of the tensor, rendering them less efficient. A non-convex approach is motivated to solve this problem in \citet{xia2017polynomial,xia2021statistically,cai2021nonconvex}, where the authors apply low-rank tensor factorization to enforce the low-rankness and update the factors iteratively. 
	
	Most of these matrix or tensor completion methods assume that the missingness occurs completely at random. However, the treatment assignment is typically an endogenous process, as it may be affected by some prognostic factors of that unit \cite{pearl2009causal}. A stream of work employing the debiased matrix/tensor completion method has been proposed in \citet{mandal2019weighted,agarwal2020synthetic,athey2021matrix,agarwal2021causal, mao2023matrix}, which re-weights each unit inversely to its probability of being assigned the actualized treatments. Admittedly, it is more challenging to recover the low-rank latent parameter matrix/tensor from the binary outcomes. A few methods have been proposed to handle such quantized and possibly corrupted outcomes, utilizing the properties of the matrix/tensor norm \cite{davenport20141,cai2013max} or enforcing the latent parameters lying in a low-rank matrix/tensor subspace \cite{wang2020learning,ashraphijuo2020union, mao2024mixed}.

	\section{Basic setup}
	\subsection{Notation and preliminaries}
	
	Before presenting our framework, let us present some basics of tensor algebra.
	Scalars are denoted by lowercase letters (e.g., $x,y$), while vectors
	and matrices use bold lowercase (e.g., $\boldsymbol{x},\boldsymbol{y}$)
	and uppercase letters (e.g., $\boldsymbol{X},\boldsymbol{Y}$), respectively.
	The outer product of vectors $\boldsymbol{x}\in\mathbb{R}^{p_{1}}$
	and $\boldsymbol{y}\in\mathbb{R}^{p_{2}}$ is $\boldsymbol{x}\otimes\boldsymbol{y}\in\mathbb{R}^{p_{1}\times p_{2}}$.
	Higher-order tensors are represented by calligraphic letters (e.g.,
	$\mathcal{X},\mathcal{Y}$). The entry-wise and inner products of
	tensors $\mathcal{X}$ and $\mathcal{Y}$
	are $\mathcal{X}\odot\mathcal{Y}$ and $\langle\mathcal{X},\mathcal{Y}\rangle=\sum_{i_{1},i_{2},i_{3}}\mathcal{X}_{i_{1},i_{2},i_{3}}\mathcal{Y}_{i_{1},i_{2},i_{3}}$,
	respectively.
	
	Tensors are unfolded into matrices on mode-$k$ using operator $\mathcal{M}_{(k)}(\cdot)$.
	For example, after unfolding the tensor $\mathcal{X}$ on its first
	mode, we have $\mathcal{M}_{(1)}(\mathcal{X})\in\mathbb{R}^{p_{1}\times p_{2}p_{3}}$,
	where $[\mathcal{M}_{(1)}(\mathcal{X})]_{i_{1},i_{2}+p_{2}(i_{3}-1)}=\mathcal{X}_{i_{1},i_{2},i_{3}}$.
	Also, mode-$k$ Tensor-matrix products are denoted by $\mathcal{X}\times_{k}\boldsymbol{U}_{k}$.
	For example, let $\boldsymbol{U}_{1}\in\mathbb{R}^{r_{1}\times p_{1}}$,
	the mode-$1$ tensor-matrix multiplication is $\mathcal{X}\times_{1}\boldsymbol{U}_{1}\in\mathbb{R}^{r_{1}\times p_{2}\times p_{3}}$,
	where $(\mathcal{X}\times_{1}\boldsymbol{U}_{1})_{i_{1},i_{2},i_{3}}=\sum_{j_{1}=1}^{p_{1}}\mathcal{X}_{j_{1},i_{2},i_{3}}\boldsymbol{U}_{i_{1},j_{1}}$.
	The multi-linear rank of a tensor $\mathcal{X}$ is $(r_{1},r_{2},r_{3})=\{\text{rank}(\mathcal{M}_{(1)}(\mathcal{X}))$,
	$\text{rank}(\mathcal{M}_{(2)}(\mathcal{X}))$, $\text{rank}(\mathcal{M}_{(3)}(\mathcal{X}))\}$.
	Tucker decomposition \citep{tucker1966some} factorizes a tensor $\mathcal{X}$
	into a core tensor $\mathcal{S}\in\mathbb{R}^{r_{1}\times r_{2}\times r_{3}}$
	and orthogonal matrices $\boldsymbol{U}_{i}\in\mathbb{R}^{p_{i}\times r_{i}}$
	as
	\begin{equation}
		\mathcal{X}=\mathcal{S}\times_{1}\boldsymbol{U}_{1}\times_{2}\boldsymbol{U}_{2}\times_{3}\boldsymbol{U}_{3}=[\![\mathcal{S};\boldsymbol{U}_{1},\boldsymbol{U}_{2},\boldsymbol{U}_{3}]\!].\label{eq:Tucker}
	\end{equation}
	In this decomposition, $\mathcal{S}$ can be considered as the principal
	components with $\boldsymbol{U}_{i}$ being the mode-$i$ loading
	matrices. Tensor norms including spectral norm $\|\mathcal{X}\|$, nuclear norm $\|\mathcal{X}\|_{*}$, Frobenius norm $\|\mathcal{X}\|_{F}$, and max norm $\|\mathcal{X}\|_{\max}$ are used;
	see \citet{kolda2009tensor} for a comprehensive review. Finally,
	$c_{0},C_{0}$ represent generic positive constants, $\asymp$ ($\lesssim$
	and $\gtrsim$) indicates equality (inequality) up to multiplicative
	numerical constants, and $[r]$ denotes the $r$-set $\{1,\cdots,r\}$.

	\subsection{Potential outcomes and causal assumptions}
	
	For each unit $i,$ $\bV_{i}=(\bX_{i},\bA_{i},\delta_{i},\bY_{i})$
	represent a $d$-dimensional covariate vector, a $k$-dimensional
	binary treatment vector $\bA_{i}=(A_{i,1},\cdots,A_{i,k})$, a churn
	indicator $\delta_{i}$ taking values from $\{0,1\}$, and a $T$-dimensional
	vector of retention statutes with possible censoring over $T$ time
	points $\bY_{i}=(Y_{i,1},\ldots,Y_{i,T})^{\T}$, respectively. The censorship of
	churn statuses is determined by the churn indicator
	$\delta_{i}$. For examples, $(\bY_{i},\delta_{i})=\{(1,1,0,\ldots,0)^{\T},1\}$
	implies that customer $i$ is retained until time point $2$ and churns
	at time point $3$, i.e., the churn status is observed; $(\bY_{i},\delta_{i})=\{(1,1,0,\ldots,0)^{\T},0\}$
	implies that customer $i$ stays for the first two time points and
	does not churn, i.e., the churn status is censored.
	
	Under the potential outcomes framework \citep{rubin1974estimating},
	$\bY_{i}^{(\ba)}$ denotes the potential churn trajectory of unit
	$i$ had the treatment be set to $\ba\in\bbA$, where $\bbA$ contains
	all possible binary treatment vectors of dimension $k$, i.e., a exhaustive set of size $2^k$. The
	"survival" function for retention at each time point $t$ is defined
	as $S_i^{(\ba)}(t)=\E(Y_{i,t}^{(\ba)})=\pr(Y_{i,t}^{(\ba)}=1)$ under treatment
	$\ba$. Similarly, the treatment-specific lifetime is defined by $\sum_{t=1}^{T}Y_{i,t}^{(\ba)}$. Since each
	unit receive only one treatment, not all potential outcomes are observable,
	necessitating certain assumptions for causal analysis:
	\begin{itemize}
		\item[A1)]  (Stable Unit Treatment Value) $\bY_{i}=\bY_{i}^{(\ba)}$ for $\bA_{i}=\ba$;
		\item[A2)]  (No Unmeasured Confounders) $\bA_{i}\ind\bY_{i}^{(\ba)}\mid\bX_{i}$
		for all $\ba\in\bbA$
		\item[A3)]  (Non-informative Censoring) $\delta_{i}\ind\bY_{i}^{(\ba)}\mid\bX_{i},\bA_{i}$;
		and 
		\item[A4)]  (Positivity) The generalized propensity score $\pi(\ba\mid\bX_{i})=\pr(\bA_{i}=\ba\mid\bX_{i})$
		for any $\ba\in\bbA$ is bounded away from $0$ and 1 almost surely. 
	\end{itemize}
	A1) rules out interference between units and multiple versions of
	treatment, A2) requires that $\bX_{i}$ accounts for all variables
	influencing both the treatment uptake and outcome, A3) is a common
	censoring at random assumption for survival analysis, which is a special
	case of the coarsening at random \citep{tsiatis2006semiparametric};
	and A4) ensures that every unit has a non-zero probability of receiving
	each level of treatment. Under Assumptions A1)--A4), the survival
	probability $S^{(\ba)}(t)$ are identifiable under all interventions.
	Moreover, in discrete survival analysis, we approximate the expected treatment-specific lifetime
	by the sum of the survival probabilities up to $T$, that is,
	$\sum_{t=1}^{T}S^{(\ba)}(t)$. Therefore, the individual optimal treatment $D_{\text{opt}}$ can be derived from $D_{i,\text{opt}}=\arg\max_{\ba\in\bbA}\sum_{t=1}^{T}S_{i}^{(\ba)}(t)$
	for the $i$-th customer.
	
	\subsection{Tensorized hazard model\label{subsec:Bernoulli-model-for}}
	
	We model potential churn statuses of $N$ units over $T$ time points
	and $L$ treatments as a $3$-mode tensor
	\[
	\mathcal{Y}=\left(Y_{i,t}^{(\ba)}\right){}_{i=1,\cdots,N,\quad t=1,\cdots,T,\quad\ba\in\bbA}=\left(\mathcal{Y}_{i,t,l}\right),
	\]
	where $\ba$ is a binary treatment vector converted to decimal form
	$l$, e.g., $l=(00011)_{10}=3$ for $\ba=(00011)$. Likewise, $(l)_{2}=\ba$
	is for binary conversion. Since $\ba$ and $l$ have one-to-one correspondence,
	we will use them interchangeably.
	
	Our objective is to leverage the tensor structures to impute the missing
	potential outcomes in $\mathcal{Y}$ and provide valid estimates for
	the causal parameters. As the entries of the potential outcome tensor,
	representing churn status, are binary ($0$ or $1$) and demonstrate
	a time-monotone pattern, direct low-rank constraints on $\mathcal{Y}$
	are inappropriate. We propose a low-rank hazard model for $\mathcal{Y}\mid\Theta\sim{\rm Bernoulli}\{\pr(\mathcal{Y}=1\mid\Theta)\}$,
	where $\Theta=(\theta_{i,t,l})$ is an unknown parameter tensor of
	the same dimension as $\mathcal{Y}$. In this model, $\theta_{i,t,l}$
	is the parameter in the hazard probability
	\begin{equation}
		\pr\left(\mathcal{Y}_{i,t,l}=1\mid\mathcal{Y}_{i,t-1,l}=1,\bX_{i}\right)=f(\theta_{i,t,l}),\label{eq:binary}
	\end{equation}
	with $f(\cdot)$ being a strictly increasing and log-concave link function. We further assume that the link function $f(\cdot)$ is twice differentiable, satisfying
	$f(\theta)+f(-\theta)=1$, $f'(\theta)=f'(-\theta)$ and $f''(\theta)=-f''(-\theta)$. These conditions are typically employed in the community of one-bit matrix/tensor completion and some common choices of $f(\cdot)$ that satisfy these conditions include the logistic link, the probit link, and the Laplacian link \cite{wang2020learning}. The individual probability of retention at one time point
	$t$ is $S_i^{(\ba)}(t)=\prod_{s=1}^{t}f(\theta_{i,s,l})$
	for treatment $a=(l)_2$, ensuring a decreasing retention probability over
	time.
	
\begin{figure}[ht]
		\vskip -0.1in
		\begin{centering}
			\includegraphics[width=.9\textwidth]{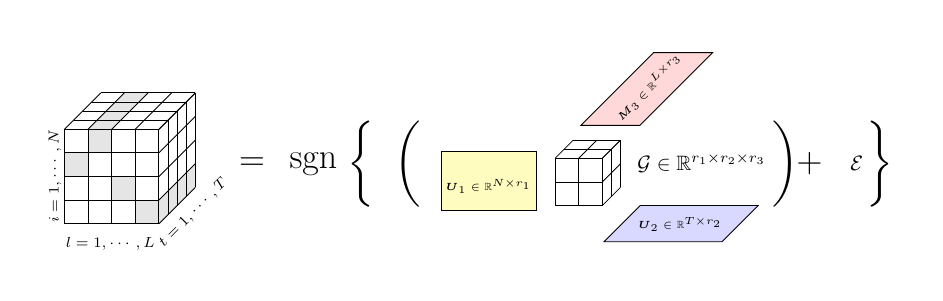}
			\par\end{centering}
		\vskip -0.1in
		\caption{\label{fig:binary-tensor} A new tensor representation of potential
			outcomes with three modes (customer $\times$ time $\times$ intervention).}
		\vskip -0.2in
	\end{figure}

	Although the individual survival probabilities $S^{(\ba)}_i(t)$
	can be approximated, we only observe $\mathcal{Y}_{i,t,l}$, which is a binary and quantized
	version of $\theta_{i,t,l}$. This is similar to the threshold model used in 1-bit matrix/tensor completion \citep{cai2013max,ghadermarzy2018learning}; see Figure \ref{fig:binary-tensor}. As $\Theta$ is latent, structural assumptions are needed for its identification and estimation.
	
	\subsection{Low-rank structure and treatment clustering}
	
	Tensor structures, particularly low-rankness, aid in parameter identification.
	Additionally, in a large treatment space, certain treatments may exhibit
	similar effects. Often, retention strategies involve a mix of various
	incentives, and altering one or two of these incentives might not
	significantly impact the overall effectiveness of the strategy. Additionally,
	some retention strategies, though targeting different aspects, are
	based on similar behavioral models and mechanisms. Identifying and
	clustering these treatments can reduce the treatment space. 
	
	We assume our parameter tensor, $\Theta$, admits the latent factor
	block model with a single discrete structure on the third mode:
	\begin{align}
		\Theta&=\mathcal{S}\times_{1}\boldsymbol{U}_{1}\times_{2}\boldsymbol{U}_{2}\times_{3}\boldsymbol{M}\nonumber\\
		&=\sum_{j_{1}=1}^{r_{1}}\sum_{j_{2}=1}^{r_{2}}\sum_{j_{3}=1}^{r_{3}}\mathcal{S}_{j_{1}j_{2}j_{3}}\bu_{1,j_{1}}\otimes\bu_{2,j_{2}}\otimes\boldsymbol{m}_{j_{3}},\label{eq:tensorM}
	\end{align}
	\vskip -0.05in
	with $\mathcal{S}\in\mathbb{R}^{r_{1}\times r_{2}\times r_{3}}$ as
	the core tensor, $\boldsymbol{U}_{1}=(\cdots\bu_{1,j_{1}}\cdots)\in\mathbb{R}^{N\times r_{1}}$
	and $\boldsymbol{U}_{2}=(\cdots\bu_{2,j_{2}}\cdots)\in\mathbb{R}^{T\times r_{2}}$
	as the\textit{ factor matrices}, and $\boldsymbol{M}=(\cdots\boldsymbol{m}_{j_{3}}\cdots)\in\{0,1\}^{L\times r_{3}}$
	as the \textit{membership matrix} such that $(\boldsymbol{M})_{ij}=1$
	if the $i$-th treatment belongs to the $j$-th cluster. 
	
	Under this representation, the multi-linear ranks $r_{1},r_{2},$
	and $r_{3}$ are constrained to be no greater than the dimensions
	$N$, $T$, and $L$, respectively. In this model, $\boldsymbol{U}_{1}$
	and $\boldsymbol{U}_{2}$ capture latent customer characteristics
	(like age, gender) and temporal patterns, respectively. The \textit{membership
		matrix} corresponds to a \textit{cluster label vector} $\boldsymbol{z}=({z}_{1},\ldots,{z}_{L})$,
	with ${z}_{l}=j$ if and only if $(\boldsymbol{M})_{ij}=1$.
	Thus, we use $\boldsymbol{M}$ and $\boldsymbol{z}$ interchangeably to denote
	the clustering structure for the third mode. The membership matrix
	$\boldsymbol{M}$ clusters treatments with similar effects, reducing
	the number of treatments. Lastly, the core tensor $\mathcal{S}$ indicates
	the interactions among these latent factors within each treatment cluster.

	\subsection{Weighted likelihood estimation for $\Theta$}
	
	For parameter estimation, we propose a weighted maximum likelihood
	estimation. In specific,
	the log-likelihood function of $\Theta$ given the hazard probability model (\ref{eq:binary}) is
	\begin{align*}
		l(\Theta) 
		& =\sum_{i,t,l}\boldsymbol{1}(\mathcal{Y}_{i,t-1,l}=1)\left[\mathcal{Y}_{i,t,l}\log\{f(\theta_{i,t,l})\} \right.\\
		&\left.+(1-\mathcal{Y}_{i,t,l})\delta_{i}\log\{1-f(\theta_{i,t,l})\}\right].
	\end{align*}
	However, the log-likelihood function is infeasible to compute directly
	due to the counterfactuals; specifically, only $N\times T$ realizations
	$\mathcal{Y}^{\text{obs}}$ is observable, leaving other parts of
	$\mathcal{Y}$ missing. This missing data is not completely random, as it is determined by the treatment mechanism, possibly confounded
	by indication, leading to potential biases in estimating $\Theta$
	when relying only on the observed data. To mitigate this issue, we
	use the inverse probability treatment weighting (IPTW) based on the propensity scores
	$\pi(\ba\mid\bX_{i})$ to adjust for confounding biases:
	\begin{equation}
		\begin{split}
			l(\Theta)&=\sum_{i,t,l=(\bA_{i})_{10}}w_{i}\boldsymbol{1}(\mathcal{Y}_{i,t-1,l}=1)\left[\mathcal{Y}_{i,t,l}\log\{f(\theta_{i,t,l})\} \right.\\
			&\left.+(1-\mathcal{Y}_{i,t,l})\delta_{i}\log\{1-f(\theta_{i,t,l})\}\right],
		\end{split}
		\label{eq:loss_1}
	\end{equation}
	where $w_{i}=\pi\{(l)_{2}\mid\bX_{i}\}^{-1}$. This weighting effectively
	generates a pseudo-population where the missingness in the data is uniform. 
	
	Besides, we can exploit additional structural relationships between
	outcomes $\bY_{i}$ and covariates $\bX_{i}$ to improve the estimation
	algorithm. For example, in the loading matrix $\boldsymbol{U}_{1}$,
	rows corresponding to customers with similar covariates might exhibit
	similarity. To utilize this insight, we proposed to decompose $\boldsymbol{U}_{1}$
	into $\boldsymbol{U}_{1}=\bX\boldsymbol{u}_{10}+\boldsymbol{u}_{11}$
	where $\boldsymbol{u}_{10}$ and $\boldsymbol{u}_{11}$ are matrices
	of unknown parameters. This covariate-assisted
	formulation of $\boldsymbol{U}_{1}$ can be easily incorporated by our
	projected gradient descent in Algorithm \ref{alg:gradient_descent}, which not only refines our understanding of the data structure but also potentially improves the precision of our estimation.
	
	\section{Algorithm\label{sec:Algorithms}}
	
	\begin{algorithm*}[tb]
		\caption{\label{alg:gradient_descent} Projected gradient descent and spectral clustering for minimizing (\ref{eq:loss_1})}
		\begin{algorithmic}
			\STATE \textbf{Input}: Observed data tuple $\bV_{i}=(\bX_{i},\bA_{i},\delta_{i},\bY_{i})$, estimated
			weights $\widehat{w}$, stepsize $\eta$, ranks $r_1,r_2$ and $r_3$\\
			\COMMENT{Initialization}
			\STATE Initialize the parameter tensor estimator $\widehat{\Theta}^{(0)}$ via the logistic regression classifier
			\FOR{$k=1, 2$}
			\STATE Initialize the singular space estimator for the first two modes via $\widehat{\boldsymbol{U}}_{k}^{(0)} =\text{SVD}_{r_{k}}\{\mathcal{M}_{(k)}(\widehat{\Theta}^{(0)})\}$
			% where $\text{SVD}_{r}(\cdot)$ returns the orthonormal matrix comprised of the top $r$ left singular vectors of a matrix
			\ENDFOR
			\STATE Compute $\widehat{\boldsymbol{F}}^{(0)}=\mathcal{M}_{(3)}(\widehat{\Theta}^{(0)})(\widehat{\boldsymbol{U}}^{(0)}_{1}\otimes\widehat{\boldsymbol{U}}^{(0)}_{2})$
			\STATE Find $\widehat{\boldsymbol{z}}^{(0)}\in[r_3]^L$ and centroids $\widehat{x}_1, \cdots, \widehat{x}_{r_3}\in \mathbb{R}^{NT}$ such that 
			$
			\sum_{l=1}^{L}\|
			(\widehat{\boldsymbol{F}})_{l:}^{\intercal} -
			\widehat{x}_{{\widehat{\boldsymbol{z}}^{(0)}}_l}
			\|_2^2
			\leq 
			\min_{x, \boldsymbol{z}}
			\sum_{l=1}^{L}
			\|(\widehat{\boldsymbol{F}})_{l:}^{\intercal} -
			{x}_{{\boldsymbol{z}}_l}
			\|_2^2
			$
			\STATE Compute $\widehat{\boldsymbol{W}}^{(0)}=\widehat{\boldsymbol{M}}^{(0)}(\text{diag}(\mathbf{1}_{L}^{\intercal}\widehat{\boldsymbol{M}}^{(0)}))^{-1}$, where $\widehat{\boldsymbol{M}}^{(0)}$ is determined by $\widehat{\boldsymbol{z}}^{(0)}$\\
			\STATE Initialize the core tensor   $\widehat{\mathcal{S}}^{(0)}=\widehat{\Theta}^{(0)}\times_1 (\widehat{\boldsymbol{U}}_{1}^{(0)})^{\intercal}\times_2 (\widehat{\boldsymbol{U}}_{2}^{(0)})^{\intercal}\times_3(\widehat{\boldsymbol{W}}^{(0)})^{\intercal}$\\
			\COMMENT{Updating}
			\FOR{$I=1,\cdots,I_{\max}$}
			\FOR{$k=1,2$} 
			\STATE $\widehat{\boldsymbol{U}}_{k}^{(I)} = \widehat{\boldsymbol{U}}_{k}^{(I-1)}-\eta\frac{\partial L(\mathcal{S}^{(I-1)},\widehat{\boldsymbol{U}}_{1}^{(I-1)},\widehat{\boldsymbol{U}}_{2}^{(I-1)},\widehat{\boldsymbol{M}}^{(I-1)})}{\partial\boldsymbol{U}_{k}}$\\
			\STATE $\widehat{\boldsymbol{U}}_{k}^{(I)} = \boldsymbol{P}_{k}(\widehat{\boldsymbol{U}}_{k}^{(I)})$, where $\boldsymbol{P}_{k}(\cdot)$ is the projection operator for $\boldsymbol{U}_k$\\
			\ENDFOR
			\STATE Update $\widehat{\boldsymbol{F}}^{(I)}$, $\widehat{\boldsymbol{z}}^{(I)}$, $\widehat{\boldsymbol{M}}^{(I)}$, and $\widehat{\boldsymbol{W}}^{(I)}$ via the nearest-neighbor search
			% \STATE Compute $\widehat{\boldsymbol{F}}^{(I)}=\mathcal{M}_{(3)}(\widehat{\Theta}^{(I-1)})(\widehat{\boldsymbol{U}}^{(I-1)}_{1}\otimes\widehat{\boldsymbol{U}}^{(I-1)}_{2})$ and $\widehat{\boldsymbol{S}}^{(I)}=
			% (\widehat{\boldsymbol{W}}^{(I-1)})^{\intercal}
			% \mathcal{M}_{(3)}(\widehat{\Theta}^{(I-1)})(\widehat{\boldsymbol{U}}_{1}^{(I-1)}
			% \otimes \widehat{\boldsymbol{U}}_{2}^{(I-1)})$ 
			% \STATE Update $\widehat{\boldsymbol{z}}^{(I)}$ by $\widehat{\boldsymbol{z}}_{l}^{(I)}=\arg\min_{b\in[r_{3}]}\|\widehat{\boldsymbol{F}}_{l,:}-\widehat{\boldsymbol{S}}^{(I)}_{b,:}\|_{2}^{2},\quad l=1,\cdots,L,$
			% \STATE Update $\widehat{\boldsymbol{W}}^{(I)}=\widehat{\boldsymbol{M}}^{(I)}(\text{diag}(\mathbf{1}_{L}^{\intercal}\widehat{\boldsymbol{M}}^{(I)}))^{-1}$
			\STATE Update $\widehat{\mathcal{S}}^{(I)}=\widehat{\mathcal{S}}^{(I-1)} - \eta \frac{\partial L(\mathcal{S}^{(I-1)},\widehat{\boldsymbol{U}}_{1}^{(I)},\widehat{\boldsymbol{U}}_{2}^{(I)},\widehat{\boldsymbol{M}}^{(I)})}{\partial\widehat{\mathcal{S}}^{(I-1)}}$
			\ENDFOR
			\STATE \COMMENT{Final results}
			\STATE 
			$\widehat{\mathcal{S}} = \widehat{\mathcal{S}}^{(I_{\max})}$, $\widehat{\boldsymbol{U}}_1 = \widehat{\boldsymbol{U}}^{(I_{\max})}_{1}$, $\widehat{\boldsymbol{U}}_2 = \widehat{\boldsymbol{U}}^{(I_{\max})}_{2}$ and $\widehat{\boldsymbol{M}} = \widehat{\boldsymbol{M}}^{(I_{\max})}$
			\STATE \textbf{Output}: Estimated the parameter tensor $\widehat{\Theta}=\widehat{\mathcal{S}}\times_{1}\widehat{\boldsymbol{U}}_{1}\times_{2}\widehat{\boldsymbol{U}}_{2}\times_{3}\widehat{\boldsymbol{M}}$
		\end{algorithmic}
	\end{algorithm*}
	
	\subsection{Propensity score estimation \label{subsec:ps_est}}
	
	Typically, the propensity scores $\pi(\ba\mid\bX_{i})$ are unknown
	in the observational studies and thus require estimation. However,
	for a high-dimensional treatment vector ($L=2^{k}$), likelihood-based
	estimation can yield nearly zero values and weighting by their inverse
	$\{\pi(\ba\mid\bX_{i};\widehat{\alpha})\}^{-1}$ is unstable \citep{yang2016propensity}.
	To improve the stability, we adopt the covariate balance propensity
	score (CBPS) methodology in \citet{imai2014covariate}. In particular,
	the following moment conditions are formulated for $j=1,\cdots,k$ and, $a=0,1$:
	\begin{equation}
		\E\left[\rho_{i,j}\mathbf{1}(A_{i,j}=a)\boldsymbol{b}(\bX_{i})\right]=\E\left[\boldsymbol{b}(\bX_{i})\right],\label{eq:moment}
	\end{equation}
	where $\rho_{i,j}^{-1}=\pr(A_{i,j}\mid\bX_{i})$ and $\boldsymbol{b}(\bX)$
	is a set of arbitrary basis functions, which can be the first, second,
	and higher-order moments of $\bX$. The key insight of (\ref{eq:moment})
	stems from the central role of the propensity scores, which balance
	the covariate distribution within the treatment groups in terms of
	the basis functions. To increase the stability of propensity score
	weighting, we propose to estimate the weights by minimizing the entropy
	balance function: $-\min_{\rho_{i,j}}\sum_{i=1}^{N}\rho_{i,j}\log\rho_{i,j},$subject
	to $\rho_{i,j}\geq0$, $\sum_{i=1}^{N}\rho_{i,j}\mathbf{1}(A_{i,j}=a)=1$,
	and $\sum_{i=1}^{N}\rho_{i,j}\mathbf{1}(A_{i,j}=a)b(\bX_{i})=N^{-1}\sum_{i=1}^{N}b(\bX_{i}).$
	The loss function is the entropy function of the weights that enforces
	the weights to be as close to one as possible, which reduces the variability
	due to heterogeneous weights \citep{hainmueller2012entropy, lee2022doubly,lee2023improving}. The
	final propensity score weights for the treatment vector become $\widehat{w}_{i}=\prod_{j=1}^{k}\widehat{\rho}_{i,j}$.
	Challenges may arise when we are facing a large number of moment constraints,
	which increases the chance of conflicting restrictions and thus do
	not produce a feasible solution space. One remedy is to couple the
	objective function (i.e., entropy balance) with regularization on
	the moment constraints to carefully select the important subset for
	balancing \citep{ning2020robust}.
	
	\subsection{Projected gradient descent and spectral clustering \label{subsec:PGD}}
	
	To maximize the weighted log-likelihood function (\ref{eq:loss_1})
	with $w_{i}$ replaced by $\widehat{w}_{i}$ under model (\ref{eq:tensorM}),
	we use the projected gradient descent method to obtain $(\mathcal{\widehat{S}},\widehat{\boldsymbol{U}}_{1},\widehat{\boldsymbol{U}}_{2})$
	along with the spectral clustering on the third mode to find the optimal
	membership $\widehat{\boldsymbol{M}}$. In particular, we first compute
	the partial gradient of $l(\Theta)$ with respect to $(\mathcal{S},\boldsymbol{U}_{1},\boldsymbol{U}_{2})$:
	\begin{align*}
		& \frac{\partial l(\Theta)}{\partial\mathcal{S}}=\nabla l\times_{1}\boldsymbol{U}_{1}^{\intercal}\times_{2}\boldsymbol{U}_{2}^{\intercal}\times_{3}\boldsymbol{M}^{\intercal},\\
		& \frac{\partial l(\Theta)}{\partial\boldsymbol{U}_{1}}=\mathcal{M}_{(1)}(\nabla l)(\boldsymbol{U}_{2}\otimes\boldsymbol{M})\mathcal{M}_{(1)}(\mathcal{S})^{\intercal},\\
		& \frac{\partial l(\Theta)}{\partial\boldsymbol{U}_{2}}=\mathcal{M}_{(2)}(\nabla l)(\boldsymbol{U}_{1}\otimes\boldsymbol{M})\mathcal{M}_{(2)}(\mathcal{S})^{\intercal},
	\end{align*}
	where $\nabla l=\partial l(\Theta)/\partial\Theta$. Next, we update
	the current solution $(\mathcal{\widehat{S}},\widehat{\boldsymbol{U}}_{1},\widehat{\boldsymbol{U}}_{2})$
	by subtracting $\eta\cdot\partial l(\Theta)/\partial(\mathcal{\widehat{S}},\widehat{\boldsymbol{U}}_{1},\widehat{\boldsymbol{U}}_{2})$,
	which moves it towards the opposite direction of partial gradients
	with step size $\eta$. To find the optimal membership for the third
	mode, we perform the nearest-neighbor search to update our estimate
	for the clustering labels $z$: 
	$$
	\widehat{z}_{l}=\arg\min_{b\in[r_{3}]}\|\mathcal{M}_{(3)}(\widehat{\mathcal{F}})_{l,:}-\mathcal{M}_{(3)}(\widehat{\mathcal{S}})_{b,:}\|_{2}^{2},
	$$
	where $l=1,\cdots,L$, $\widehat{\mathcal{F}}=\widehat{\Theta}\times_{1}(\widehat{\boldsymbol{U}}_{1})^{\intercal}\times_{2}(\widehat{\boldsymbol{U}}_{2})^{\intercal}$
	is the projected mode-$3$ slices, $\widehat{\mathcal{S}}=\widehat{\Theta}\times_{1}(\widehat{\boldsymbol{U}}_{1})^{\intercal}\times_{2}(\widehat{\boldsymbol{U}}_{2})^{\intercal}\times_{3}(\widehat{\boldsymbol{W}})^{\intercal}$
	is the projected mode-$3$ block means, and $\widehat{\boldsymbol{W}}=\widehat{\boldsymbol{M}}(\text{diag}(\mathbf{1}_{L}^{\intercal}\widehat{\boldsymbol{M}}))^{-1}$.  Intuitively, $\mathcal{M}_{(3)}(\widehat{\mathcal{F}})_{l,:}$ contains the information of the $l$-th treatment, and $\mathcal{M}_{(3)}(\widehat{\mathcal{S}})_{b,:}$ contains the information of the $b$-th cluster. Our strategy is to find $b$  for each $l$ such that $\mathcal{M}_{(3)}(\widehat{\mathcal{S}})_{b,:}$ is the closest to $\mathcal{M}_{(3)}(\widehat{\mathcal{F}})_{l,:}$. Besides, $\widehat{\mathcal{F}}$ and $\widehat{\mathcal{S}}$
	utilize the information from the other two modes for projection, which can significantly reduce the noise level within
	the estimated parameter tensor $\widehat{\Theta}$. 
	
	We provide the
	details of our procedure for optimizing (\ref{eq:loss_1}) in Algorithm \ref{alg:gradient_descent}. In practice, we recommend a BIC-type criterion to select the rank parameters $(r_{1},r_{2},r_{3})$, and all the tuning parameters are tuned sequentially \citep{ibriga2022covariate}. 
	% by minimizing:
	% \begin{align*}
		% \mathrm{BIC}&= -\sum_{i,t,l=(\bA_{i})_{10}}w_{i}\mathbf{1}(\mathcal{Y}_{i,t-1,l}=1)\left[\mathcal{Y}_{i,t,l}\log\{f(\theta_{i,t,l})\} \right. \\
		% &\left.+(1-\mathcal{Y}_{i,t,l})\delta_{i}\log\{1-f(\theta_{i,t,l})\}\right]\\
		%  & +\log(NTL)\\
		%  &\times\{r_{1}r_{2}r_{3}+(N-r_{1})r_{1}+(T-r_{2})r_{2}+(L-r_{3})r_{3}\},
		% \end{align*}

	\section{Statistical theory}
	
	In this section, we study the statistical properties of $\widehat{\Theta}$
	under the tensorized latent factor block hazard model (\ref{eq:tensorM}).
	Mainly, we focus on evaluating the performance in two metrics: 1)
	estimation: the estimation accuracy for the parameter tensor $\Theta$
	(Section \ref{subsec:estimation}); 2) clustering: the correct recovery
	rate of the membership $\boldsymbol{M}$ (Section \ref{subsec:clustering}). To begin with, we provide a summary of the notation in Table \ref{tab:notation}.
	
	\begin{table}[ht]
		\centering
		\begin{tabular}{ll}
			\toprule
			Notation & Definition\\
			\midrule
			${\mathcal{F}}$ &${\Theta}\times_{1}({\boldsymbol{U}}_{1})^{\intercal}\times_{2}({\boldsymbol{U}}_{2})^{\intercal}$ \\
			${\mathcal{S}}$ &$\widehat{\Theta}\times_{1}({\boldsymbol{U}}_{1})^{\intercal}\times_{2}({\boldsymbol{U}}_{2})^{\intercal}\times_{3}({\boldsymbol{W}})^{\intercal}$\\
			${\boldsymbol{W}}$ &${\boldsymbol{M}}(\text{diag}(\mathbf{1}_{L}^{\intercal}{\boldsymbol{M}}))^{-1}$\\
			$p_{\min}$  & $p_{\min}\leq \pr(\delta_{i}=1,\mathcal{Y}_{i,t,l}=1)$ $\forall$ $i,t$ and $l$  \\
			$w_{\min}, w_{\max}$ &$w_{\min}\leq\pi(\ba\mid\bX_{i})^{-1}\leq w_{\max}$\\
			$r_1, r_2$ & number of latent unit and temporal factors\\
			$r_3$ &  number of groups for the treatments\\
			\bottomrule
		\end{tabular}
		\caption{\label{tab:notation} Summary of notation}
	\end{table}
	
	\subsection{Upper bound error for estimation \label{subsec:estimation}}
	
	The estimation accuracy of $\widehat{\Theta}$ is evaluated by its
	deviation to $\Theta$ in Frobenius norm. First, we define two quantities
	$L_{\alpha}$ and $\gamma_{\alpha}$ to control the steepness and
	convexity of the link function $f(\cdot)$, where $L_{\alpha}=\sup_{|\theta|\leq\alpha}\left[{f'(\theta)}/{f(\theta)},{f'(\theta)}/\{1-f(\theta)\}\right]$ and $\gamma_{\alpha}=\inf_{|\theta|\leq\alpha}[{\{f'(\theta)\}^{2}}/{f^{2}(\theta)}-{f''(\theta)}/{f(\theta)}, {f''(\theta)}/\{1-f(\theta)\}+{\{f'(\theta)\}^{2}}/{\{1-f(\theta)\}^{2}}]$, where $f'(\theta)=df(\theta)/d\theta$. Theorem \ref{thm:estimation}
	establishes the upper bound of the estimation error of $\widehat{\Theta}$.
	\begin{theorem}
		\label{thm:estimation} Under Assumptions A1) to A4) and some regularity
		conditions, suppose $\mathcal{Y}$ is the binary tensor characterized
		by the parameter tensor $\Theta$ as model (\ref{eq:binary}) with
		the link function $f(\cdot)$. Let $\widehat{\Theta}$ be the local
		maximizer of (\ref{eq:loss_1}), there exist constants $c_{0}$, $C_{0}$,
		$C_{1}$ and $C_{2}$, such that with probability greater than $1-c_{0}(N+T+L)^{-2}$,
		we have
		\begin{align*}
			&\frac{\|\widehat{\Theta}-\Theta\|_{F}^{2}}{NT}  \leq C_{0}\frac{\|\Theta\|_{\max}^{2}}{Np_{\min}}\log(N+T+L)\\
			& \vee C_{1}\frac{L_{\alpha}^{2}}{\gamma_{\alpha}^{2}}\frac{w_{\max}^{2}}{w_{\min}^{2}}\frac{r_{1}r_{2}r_{3}(N\vee T)}{NTp_{\min}^{2}\max(r_{1},r_{2},r_{3})}\log^{2}(N+T+L)\\
			& \vee C_{2}\frac{r_{1}r_{2}r_{3}(N\vee T)\|\Theta\|_{\max}^{2}}{NTp_{\min}^{2}\max(r_{1},r_{2},r_{3})}\log(N+T+L).
		\end{align*}
	\end{theorem}
	
	Theorem \ref{thm:estimation} shows that the upper bound for estimation
	error converges to zero as the sample size $N$ increases; see Appendix for the experimental evidence. The parameter $p_{\min}$ plays an important role in our
	theoretical analysis as the recovery of $\Theta$ will be harder when
	$p_{\min}$ becomes smaller. Intuitively, if $p_{\min}$ is too small,
	it is unlikely to observe the churn statuses for all the customers
	and thus recovering their churn patterns will be impossible. Furthermore,
	the estimation error for the survival probabilities
	is also bounded if $f(\cdot)$ satisfies the local Lipschitz condition,
	that is, $\|f(\widehat{\Theta})-f(\Theta)\|_{F}\leq c_{1}\|\widehat{\Theta}-\Theta\|_{F}$
	for some constant $c_{1}$. Below, we present a special case of Theorem
	\ref{thm:estimation} under the logistic link function.
	\begin{corollary}
		Assume the logistic model for $\mathcal{Y}_{i,t,l}$ with link function
		$f(\theta)=e^{\theta/\sigma}/(1+e^{\theta/\sigma})$, it can be shown
		that 
		\begin{align*}
			% L_{\alpha} & =\frac{1}{\sigma}\sup_{|\theta|\leq\alpha}\max\{1-f(\theta),f(\theta)\}=\frac{1}{\sigma},\\
			% \gamma_{\alpha} & =\frac{1}{\sigma^{2}}\sup_{|\theta|\leq\alpha}f(\theta)\{1-f(\theta)\}=\frac{e^{\alpha/\sigma}}{(1+e^{\alpha/\sigma})^{2}\sigma^{2}},\\
			L_{\alpha}^{2}/\gamma_{\alpha}^{2} & =\frac{(1+e^{\alpha/\sigma})^{4}\sigma^{2}}{e^{2\alpha/\sigma}}=\left(2+e^{\alpha/\sigma}+\frac{1}{e^{\alpha/\sigma}}\right)^{2}\sigma^{2}.
		\end{align*}
		Further, suppose $w_{\min}\asymp w_{\max}$, $p_{\min}$ is bounded
		and $\|\Theta\|_{\max}\asymp\sigma$, the estimation error in Theorem
		\ref{thm:estimation} becomes
		\begin{equation}
			\begin{split}
				\frac{\|\widehat{\Theta}-\Theta\|_{F}^{2}}{NT} & \lesssim\frac{r_{1}r_{2}r_{3}(N\vee T)}{NT\max(r_{1},r_{2},r_{3})}\\
				&\times(\sigma^{2}\vee\|\Theta\|_{\max}^{2})\text{poly}\log(N+T+L),
			\end{split}\label{eq:error_logit}
		\end{equation}
		where $\text{poly}\log(\cdot)$ is a certain polynomial of the logarithmic
		function. This non-asymptotic bound (\ref{eq:error_logit}) is similar
		to the risk bound in Theorem 3, \citet{xia2021statistically}.
	\end{corollary}

	\subsection{Upper bound error for clustering \label{subsec:clustering}}
	
	Next, we examine the clustering accuracy of our proposed model for
	the third mode. The most common metric to assess the clustering performance
	is the \textit{classification error rate}, defined by $h(\boldsymbol{c},\boldsymbol{d})=\min_{\pi\in\Pi_{r_{3}}}\sum_{l=1}^{L}\mathbf{1}\{\boldsymbol{c}_{l}\neq\pi(\boldsymbol{d})_{l}\}/L$, for two label vectors $\boldsymbol{c}=(c_{1},\cdots,c_{L})^{\intercal}$
	and $\boldsymbol{d}=(d_{1},\cdots,d_{L})^{\intercal}$, where $\Pi_{r_{3}}$
	is the collection of all permutations of $\{1,\cdots,r_{3}\}$. Let
	$\widehat{\boldsymbol{z}}$ be the estimated clustering labels, we
	claim that $\widehat{\boldsymbol{z}}$ is a consistent clustering
	for $\boldsymbol{z}$ if $\pr\{h(\widehat{\boldsymbol{z}},\boldsymbol{z})>\varepsilon\}\rightarrow0$
	when the sample size $N$ goes to infinity for any $\varepsilon>0$.
	In order to facilitate our theoretical analysis, we introduce the concept of \textit{misclassification loss}:
	$$
	g(\boldsymbol{c},\boldsymbol{d})=\min_{\pi\in\Pi_{r_{3}}}\frac{1}{L}\sum_{l=1}^{L}\|\mathcal{M}_{(3)}(\mathcal{S})_{(\boldsymbol{c})_{l},:}-\mathcal{M}_{(3)}(\mathcal{S})_{\pi(\boldsymbol{d})_{l},:}\|_{2}^{2},
	$$
	which is a more convenient measure compared to $h(\boldsymbol{c},\boldsymbol{d})$.
	Moreover, a close relationship between $h(\boldsymbol{c},\boldsymbol{d})$
	and $g(\boldsymbol{c},\boldsymbol{d})$ can be formulated as $h(\boldsymbol{c},\boldsymbol{d})\leq g(\boldsymbol{c},\boldsymbol{d})/\Delta_{\min}^{2}$
	(Lemma 1, \citet{han2022exact}), which implies that it suffices to
	bound $g(\boldsymbol{c},\boldsymbol{d})$ for establishing the clustering
	consistency. 
	\begin{theorem}
		\label{thm:clustering} Under the same condition in Theorem \ref{thm:estimation}.
		Assume the signal-to-noise ratio (SNR) satisfies
		\begin{align*}\text{SNR}&=\frac{\Delta_{\min}^{2}}{\|\Theta\|_{\max}^{2}\vee(L_{\alpha}^{2}/\gamma_{\alpha}^{2})}\\
			&\gtrsim\frac{w_{\max}^{2}(N\vee T)r_{1}^{2}r_{2}^{2}r_{3}\mu_{0}^{2}\log^{2}(N+T+L)}{w_{\min}^{2}p_{\min}^{2}NTL\max(r_{1},r_{2},r_{3})},
		\end{align*}
		where $\Delta_{\min}$ measures the minimum separation of the projected block means, and $\mu_{0}$ measures the incoherence of $\Theta$; see details
		in the Appendix. There exists a constant $c_{0}$, such that, with
		probability greater than $1-c_{0}(N+T+L)^{-2}$,
		\begin{align*}
			&h(\widehat{z},z) \leq g(\widehat{z},z)/\Delta_{\min}^2\lesssim\frac{\|\Theta\|_{\max}^{2}\vee(L_{\alpha}^{2}/\gamma_{\alpha}^{2})}{\Delta_{\min}^2(N+T+L)^{2}}.
			% \\
			% &h(\widehat{z},z)\lesssim\frac{\|\Theta\|_{\max}^{2}\vee(L_{\alpha}^{2}/\gamma_{\alpha}^{2})}{\Delta_{\min}^{2}(N+T+L)^{2}}.
		\end{align*}
	\end{theorem}
	
	The notion of SNR under the proposed model is quantified by the minimum
	gaps between the projected means on the third mode, i.e., $\Delta_{\min}$,
	over the level of noises induced by the parameter tensor and the Bernoulli
	model (\ref{eq:binary}). If the SNR is large enough as stated in
	Theorem \ref{thm:clustering}, the consistency of the clustering will
	be established; see Appendix for the experimental evidence. A similar SNR condition also appears in Theorem 2, \citet{han2022exact}. 
	
	\section{Numerical studies \label{sec:experiments}}
	In this section, we examine the performance of our proposed model
	under various settings. For starter, the baseline covariates $\bX\in\mathbb{R}^{N\times d}$
	are generated by $\bX_{i}\overset{i.i.d.}{\sim}\mathcal{N}(0,I_{d})$
	with $d=3$. We generate the true parameter tensor $\Theta$ with
	each entry $\theta_{i,t,l}$ defined by: $\theta_{i,t,l}=(\bX_{i}^{\intercal}\eta_{N})\cdot(t\eta_{T}/T)\cdot\mathrm{cum}\{(l)_{2}\}\eta_{L}$,
	where $\eta_{N}=(1,1,1)$, $\eta_{T}=1$, $\eta_{L}=1$, and $\mathrm{cum}\{(l)_{2}\}$
	indicates the number of active treatments in $(l)_{2}$. The potential
	outcome $\mathcal{Y}_{i,t,l}$ is generated sequentially with the
	conditional probability $P(\mathcal{Y}_{i,t,l}=1\mid\mathcal{Y}_{i,t-1,l}=1,\bX_{i})=\mathrm{expit}(\theta_{i,t,l})$
	when $\mathcal{Y}_{i,t-1,l}=1$ and $\bA_{i}=(l)_{2}$; otherwise, $\mathcal{Y}_{i,t,l}=0$
	by definition. The lifetime for each customer $i$ is computed by
	$\text{Time}_{i}=\sum_{t=1}^{T}\sum_{l=1}^{L}\mathbf{1}\{\bA_{i}=(l)_{2}\}\mathcal{Y}_{i,t,l}$,
	and we randomly select $20\%$ patients to be right-censored with
	random censoring time uniformly drawn from $[0,\text{Time}_{i}]$. Next, each
	entry of the $k$-dimensional binary treatment vector $\bA_{i}$ is
	generated independently for $j=1,\cdots,k$: 
	\[
	A_{i,j}\mid\bX_{i}\sim\text{Bernoulli}\left\{ \frac{\exp(\alpha_{A}^{\intercal}\bX_{i})}{1+\exp(\alpha_{A}^{\intercal}\bX_{i})}\right\},
	\]
	where $\alpha_{A}=(.5,.5,.5)$. The CBPS method in Section \ref{subsec:ps_est}
	is utilized for propensity score estimation with the basis functions
	$b(\bX)$ being the first moment of $\bX$. We consider the setting
	where $T=5,10$, $N=100,300,500,1000,2000$, and $k=2,3,4$. All simulation
	results are reported based on $100$ data replications.

	% \subsection{Comparison with other algorithms \label{subsec:other-algorithms}}
	
	One primary goal for the customer churn analysis is to identify the optimal treatment that leads to the longest customer retention time.
	With the estimated parameter tensor $\widehat{\Theta}$, one can find the individual optimal treatment $\hat{D}_{i,\text{opt}}$
	by $\max_l\sum_{t=1}^{T}\prod_{s=1}^{t}{\rm expit}(\widehat{\theta}_{i,s,l})$,
	which maximize the estimated expected lifetime for each customer $i$. We showcase
	the effectiveness of our proposed model for identifying the optimal
	treatment compared with other methods. In particular, we consider
	the competitive methods within two categories. 1) binary classification:
	logistic regression classifier (logit), random forest classifier (RF), neural network classifier (NN), support vector machine (SVM), gradient-boosted classifier (gradBoost), adaptive boosting classifier (AdaBoost), and a soft voting classifier combining all mentioned binary classifiers (Vote); 2) survival models: Cox proportional
	hazard model (Cox-PH), survival random forest (surv-RF), gradient-boosted
	Cox PH model with regression trees as base learners (Cox-modelBoost),
	gradient boosting with component-wise least squares as base learners
	(Cox-gradBoost). The hyperparameters for these algorithms are chosen
	by default from packages \texttt{sklearn} and \texttt{sksurv}.

\begin{figure}[!ht]
		\begin{center}
			\includegraphics[width=0.8\textwidth]{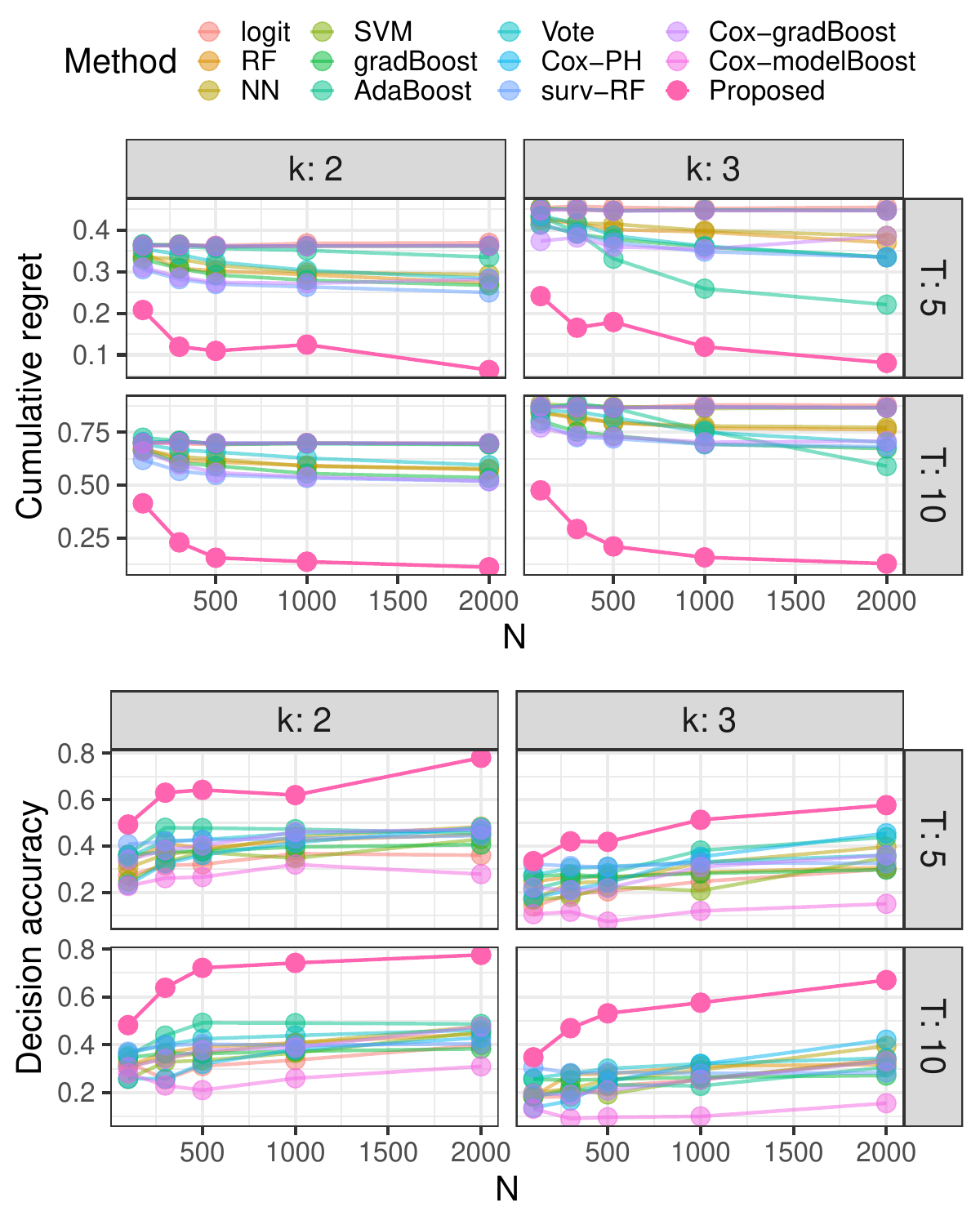}
			\vskip -0.1in
			\caption{\label{fig:cum-regret-decision} Cumulative regret (top) and decision
				accuracy (bottom) of the proposed method and other competitors when
				$N=100,300,500,1000,2000$, $T=5,10$ and $k=2,3$.}
		\end{center}
		\vskip -0.2in
	\end{figure}

	We assess the performance by the cumulative regret and decision accuracy using the true optimal treatment
	$D_{i,\text{opt}}$; more definitions are deferred to the Appendix.
	% as:
	% \begin{align*}
		%     &\text{regret}(\widehat{D}_{\text{opt}},D_{\text{opt}})=N^{-1}\sum_{i=1}^{N}\sum_{t=1}^{T}(P_{i,t,D_{i,\text{opt}}}-P_{i,t,\widehat{D}_{i,\text{opt}}}),\\
		%     &\text{acc}(\widehat{D}_{\text{opt}},D_{\text{opt}})=N^{-1}\sum_{i=1}^{N}\mathbf{1}(D_{i,\text{opt}}=\widehat{D}_{i,\text{opt}}).
		% \end{align*}
	Specifically, cumulative regret is defined as the average difference between
	the survival probabilities evaluated under the estimated and the true optimal treatments summing over all the time points. Therefore, the regret captures how the estimated incorrect decisions
	impair the outcome. The decision accuracy describes the proportion
	of estimated optimal treatment that matches the true optimal treatment.
	We evaluate different methods by their cumulative regrets and decision
	accuracy (Figure \ref{fig:cum-regret-decision}) for varying $N$,
	$T$ and $k$. One can observe the proposed model outperforms the
	other methods by large margins in all cases.

	\section{Real-data application}\label{sec:application}
	
	In this section, we apply our proposed method to one bank customer
	churn data from a Kaggle competition\footnote{https://www.kaggle.com/datasets/radheshyamkollipara/bank-customer-churn.}. Four important indicators of
	customer loyalty are considered: card types (0 for Silver and Gold, and 1 for Platinum and Diamond), the number of bank products (0 for only one product, and 1 for more than one), whether the customer
	has complained or not, and the post-complaint scores (0 for rates less than 3, and 1 otherwise). It
	is known that companies might employ different interventions based
	on different subgroups of customers to avoid customer churn. Therefore,
	we formulate the binary treatment vector $\bA_{i}$ according to these
	indicators. Next, we consider eight customer
	characteristics $\bX_{i}$, including age, gender, geography, estimated
	salary, account balance, earned credit points, and two customer characteristic
	indicators: whether or not the customers have credit cards and whether
	or not the customers are active. The customer retention time $Y_{i}$
	is measured by the number of years that the customers have engaged
	with the bank and the churn status $\delta_{i}$ is an indicator of
	whether or not the customers left the bank. 
	
	For the model training, we divide the dataset into 80\% training data
	and 20\% test data, implementing 5-fold cross-validation. Continuous
	variables are standardized, and categorical variables are one-hot
	encoded.\textbf{ }The goodness-of-fit metrics for assessing different
	methods are the concordance index (C-index) and the average time-dependent
	area under the curve (AUC). C-index, a widely used index for survival
	analysis, examines the performance by the fraction of pairs whose
	predicted retention times have the correct order compared to their
	observed retention times in the test set. The time-dependent ROC curve evaluates the model's ability to distinguish the customers who exit
	by a given time $t$ from those who exit after $t$. Table \ref{tab:res}
	compares the results of our method with other methods in terms of their goodness-of-fit. In conclusion, our proposal performs
	well, exhibiting notably superior performance compared to other methods.
	
	\begin{table}[!ht]
		\begin{centering}
			\begin{tabular}{lccc}
				\toprule 
				& C-index ($\uparrow$) & average AUC ($\uparrow$)
				\tabularnewline
				\midrule 
				logit & 0.19 $\pm$ 0.08 & 0.43 $\pm$ 0.04\tabularnewline
				RF & 0.44 $\pm$ 0.01 & 0.46 $\pm$ 0.01\tabularnewline
				NN & 0.31 $\pm$ 0.02 & 0.36 $\pm$ 0.01\tabularnewline
				SVM & 0.49 $\pm$ 0.04 & 0.47 $\pm$ 0.12\\
				gradBoost & 0.50 $\pm$ 0.01 & 0.47 $\pm$ 0.02\\
				AdaBoost & 0.45 $\pm$ 0.03 & 0.39 $\pm$ 0.04\\
				Vote & 0.44 $\pm$ 0.01 & 0.39 $\pm$ 0.03\\
				Cox-PH & 0.30 $\pm$ 0.01 & 0.30 $\pm$ 0.01\tabularnewline
				surv-RF & 0.46 $\pm$ 0.01 & 0.40 $\pm$ 0.01\tabularnewline
				Cox-gradBoost & 0.43 $\pm$ 0.01 & 0.40 $\pm$ 0.01\tabularnewline
				Cox-modelBoost & 0.46 $\pm$ 0.01 & 0.42 $\pm$ 0.01\tabularnewline
				\textbf{Proposed} & \textbf{0.58} $\pm$ 0.02 & \textbf{0.57} $\pm$ 0.02\tabularnewline
				\bottomrule 
			\end{tabular}
			\par\end{centering}
		\caption{\label{tab:res} Evaluation metrics on the bank customer churn data.}
	\end{table}
	
\begin{figure}[!ht]
		\centering
		\includegraphics[width=.85\textwidth]{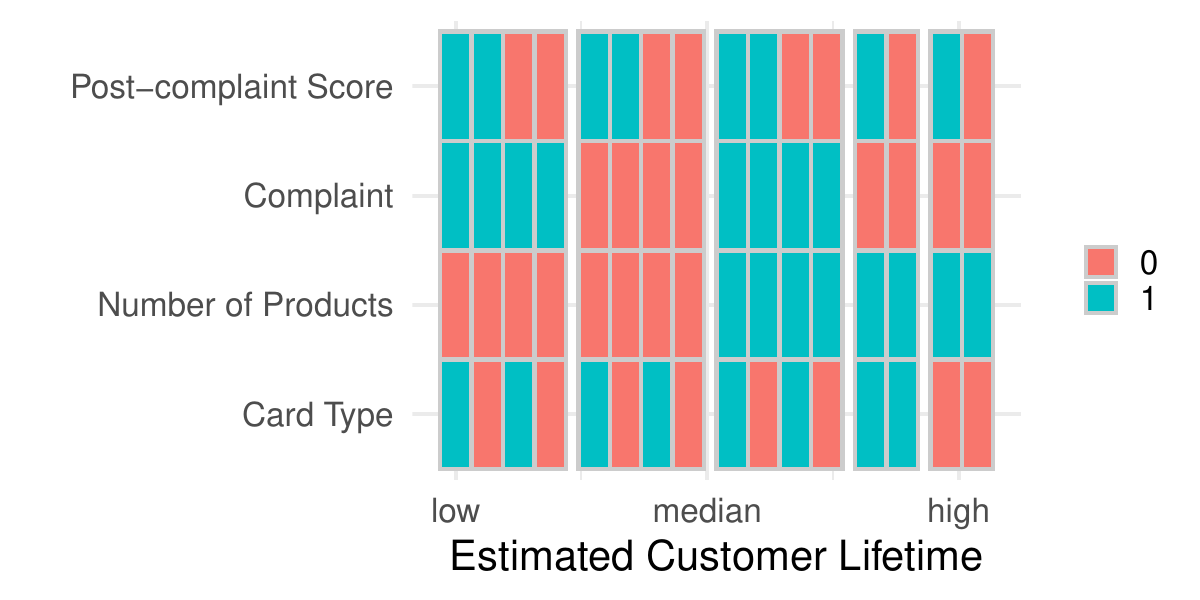}
		\vskip -0.1in
		\caption{\label{fig-treatment-strucutre} Estimated intervention structure by the proposed model for the bank customer churn data. All interventions are clustered by blocks and ordered by their expected customer lifetimes.}
		\vskip -0.1in
	\end{figure}

	To better explore the homogeneity of the intervention effect, we estimate
	the expected customer lifetime within each intervention group under our
	model. Figure \ref{fig-treatment-strucutre} shows that the customers
	with more bank products tend to have longer retention times. This
	finding aligns with our expectations, as customers associated with
	more bank products are more loyal and inclined to remain engaged with
	the company. Conversely, customers who have fewer bank products and
	complained previously exhibit the shortest customer lifetimes regardless
	of their post-complaint scores. Hence, these customers are more
	likely to leave the company. Given the higher cost involved in acquiring
	new customers rather than retaining the existing ones, our churn analysis
	alerts the need for the company to design retention campaigns targeting
	the customers with fewer bank products who have raised complaints
	before. 
	
	\section{Discussion}
	
	In this paper, we focus on the causal analysis of the customer churn
	problem with multiple interventions. In particular, we adopt the idea
	of 1-bit tensor completion to estimate the survival probabilities under all interventions. Moreover, we propose the tensorized
	latent factor block hazard model to cluster the interventions with
	similar impacts. This model enables us to identify the optimal intervention
	group, which improves the practicality of implementing the optimal
	retention strategies in practice. The proposed method can be extended
	into several aspects. First, we only consider the time-invariant treatment
	in this paper. When treatment is time-varying, two classes of models
	namely marginal structural models \citep{yang2018modeling} and structural
	failure time models \citep{yang2020semiparametric} are useful, which,
	however, often posit parametric structural model assumptions.
	The proposed framework can be extended to this setting by expanding
	the treatment mode. Second, our current identification assumption
	requires the treatment ignorability in the sense that all confounders
	are captured and are adjusted. One interesting future direction is
	to extend the current framework under the latent ignorability of treatment
	assignment in the sense that treatment ignorability holds when conditioning
	on the latent factors \citep{lewis2020double,agarwal2022synthetic}.

	\section*{Acknowledgements}
	We would like to thank the anonymous (meta-)reviewers of
	ICML 2024 for helpful comments. This work is partially
	supported by 
	the U.S. National Science Foundation and National Institute of Health.
	
	\section*{Impact Statement}
	This paper presents work aimed at predicting customer churn and developing informed strategies to improve customer retention. The potential societal consequences of this work could be significant, including fostering more sustainable business practices, enhancing customer satisfaction, promoting economic stability by reducing the frequency of business failures, and contributing to higher levels of service quality and consumer trust in various industries.
	
	\bibliography{main}
	\bibliographystyle{icml2024}

	\onecolumn
	\newpage
	% \section*{Appendix}
	\appendix
	\renewcommand{\thefigure}{A\arabic{figure}}
	\renewcommand{\thetable}{A\arabic{table}}
	\setcounter{figure}{0}
	\setcounter{table}{0}
	\section*{Appendix}
	\section{Additional Numerical Experiments}
	\subsection{Convergence analysis for estimation and clustering}
\begin{figure}[ht]
		% \vskip 0.2in
		\begin{center}    \includegraphics[width=0.85\textwidth]{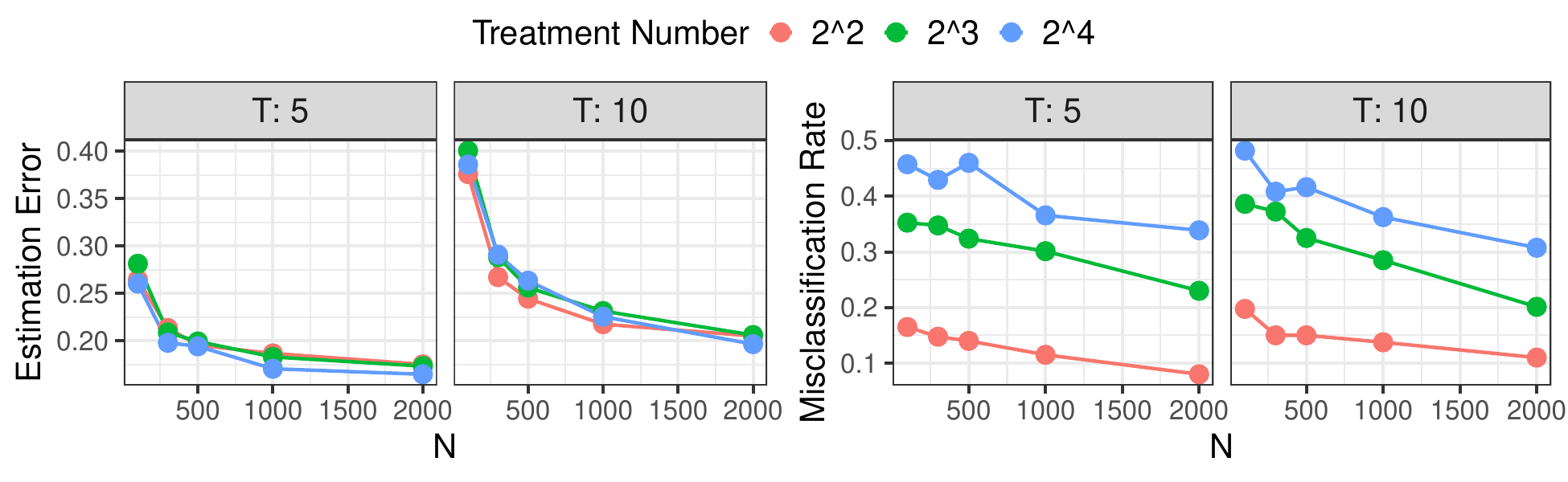}
			
			\caption{\label{fig:error_misclass} Normalized mean squared error (left) and the misclassification error rate (right) when $T=5$ and $N=100,300,500,1000,2000$ over $100$ data replications.}
		\end{center}
		\vskip -0.2in
	\end{figure}

	We first assess the effectiveness of the proposed model in recovering
	the parameter tensor $\Theta$. The performance is evaluated by the
	normalized tensor mean squared error as $\ell_{2}(\widehat{\Theta})=\|\widehat{\Theta}-\Theta\|_{F}^{2}/\|\Theta\|_{F}^{2}$,
	and the classification error rate. Figure \ref{fig:error_misclass}
	shows that the estimation and clustering errors decrease as the sample
	size $N$ increases. This shows that larger sample sizes enhance the
	performance of the proposed model, which corroborates our theoretical
	results in Theorems \ref{thm:estimation} and \ref{thm:clustering}.
	
\begin{figure}[ht]
		\begin{center}    
			% \vskip 0.2in
			\includegraphics[width=0.85\textwidth]{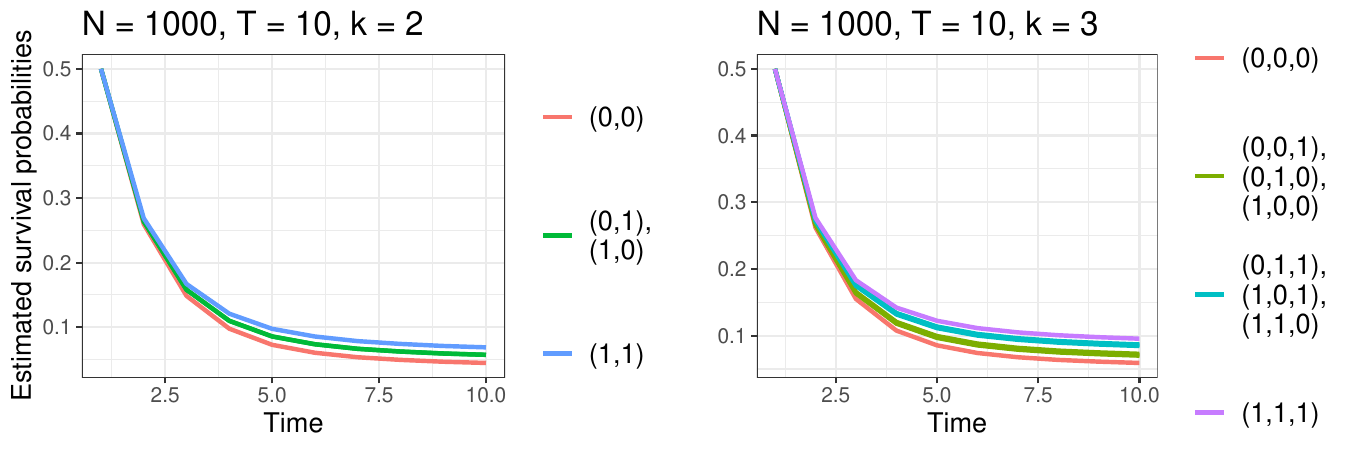}\caption{\label{fig:Plot-estimated-curves} Plot of the average estimated survival
				probabilities for all treatments when $N=1000,T=10$, $k=2$ (left) and $k=3$ (right) over
				$100$ data replications.}
		\end{center}
		\vskip -0.2in
	\end{figure}

	Since the parameter tensor $\Theta$ is estimated, the average survival
	function $S^{(\ba)}(t)$ under any treatment can be obtained by $\widehat{S}^{(\ba)}(t)=N^{-1}\sum_{i=1}^{N}\prod_{s=1}^{t}{\rm expit}(\widehat{\theta}_{i,s,l})$.
	In particular, we plot the average estimated survival functions
	for all treatments over$100$ data replications when $N=1000$ to
	assess the clustering results in Figure \ref{fig:Plot-estimated-curves}.
	The results imply that the more active treatments one receives, the
	higher the survival probabilities will be, which aligns with
	our data generation process of $\theta_{i,t,l}$ as it relies on the
	number of active treatments $\mathrm{cum}\{(l)_{2}\}$.
	
	Next, we provide the complexity of the algorithm by presenting the running time and the normalized tensor mean squared errors $\ell_2$ for our numerical experiments. Empirically, we present the results in Table \ref{tab:running_time} for both the 10000-iteration and 1000-iteration projected gradient descent algorithms. In summary, the running time of the proposed framework is linear with respect to $N$, $T$, $k$, and the number of iterations, which aligns with the theory for the gradient descent algorithm. Additionally, we suggest using the 1000-iteration projected gradient descent if saving time is a priority (the running time is significantly reduced), as the performance, in terms of normalized tensor mean squared error, does not decrease significantly when reducing the number of iterations from 10000 to 1000. All experiments are conducted on a computer with an Intel(R) Xeon(R) Gold 6226R CPU @ 2.90GHz and 32GB RAM.

	\begin{table}[!ht]
		\centering
		\begin{tabular}{lllllll}
			\toprule
			$N$ & $T$ & $k$ & running time$^{(10000)}$ & running time$^{(1000)}$ & $\ell_2^{(10000)}$ & $\ell_2^{(1000)}$ \\ 
			\midrule
			100 & 5 & 2 & 23.95 (0.7) & 1.86 (0.09) & 0.42 (0.07) & 0.57 (0.02) \\ 
			100 & 5 & 3 & 21.49 (1.18) & 2.05 (0.13) & 0.31 (0.07) & 0.59 (0.03) \\ 
			% 100 & 5 & 4 & 24.84 (1.9) & 2.56 (0.06) & 0.27 (0.06) & 0.52 (0.07) \\ 
			100 & 10 & 2 & 26.14 (0.47) & 1.79 (0.05) & 0.44 (0.06) & 0.54 (0.03) \\ 
			100 & 10 & 3 & 19.53 (0.73) & 1.91 (0.21) & 0.43 (0.09) & 0.6 (0.03) \\ 
			% 100 & 10 & 4 & 22.83 (1.79) & 4.08 (0.16) & 0.39 (0.1) & 0.58 (0.07) \\ 
			300 & 5 & 2 & 35.18 (1.4) & 3.54 (0.16) & 0.24 (0.03) & 0.53 (0.02) \\ 
			300 & 5 & 3 & 32.10 (3.66) & 3.28 (0.28) & 0.21 (0.03) & 0.47 (0.05) \\ 
			% 300 & 5 & 4 & 43.39 (7.83) & 4.74 (0.13) & 0.2 (0.03) & 0.29 (0.05) \\ 
			300 & 10 & 2 & 35.46 (1.48) & 3.40 (0.11) & 0.31 (0.05) & 0.52 (0.02) \\ 
			300 & 10 & 3 & 39.49 (3.34) & 5.34 (0.22) & 0.29 (0.05) & 0.51 (0.04) \\ 
			% 300 & 10 & 4 & 61.55 (10.33) & 6.87 (0.2) & 0.29 (0.06) & 0.38 (0.06) \\ 
			500 & 5 & 2 & 43.20 (3.01) & 5.29 (0.19) & 0.20 (0.02) & 0.48 (0.02) \\ 
			500 & 5 & 3 & 46.93 (10.18) & 5.30 (0.31) & 0.20 (0.03) & 0.36 (0.05) \\ 
			% 500 & 5 & 4 & 56.57 (9.85) & 8.04 (0.37) & 0.19 (0.03) & 0.22 (0.04) \\ 
			500 & 10 & 2 & 58.33 (1.96) & 4.78 (0.19) & 0.25 (0.04) & 0.48 (0.03) \\ 
			500 & 10 & 3 & 60.42 (7.21) & 6.96 (0.19) & 0.26 (0.04) & 0.42 (0.06) \\ 
			% 500 & 10 & 4 & 73.19 (21.39) & 11.64 (0.62) & 0.26 (0.05) & 0.3 (0.05) \\ 
			1000 & 5 & 2 & 90.55 (19.42) & 11.48 (0.74) & 0.19 (0.01) & 0.36 (0.03) \\ 
			1000 & 5 & 3 & 125.23 (28.41) & 10.91 (0.41) & 0.19 (0.01) & 0.24 (0.02) \\ 
			% 1000 & 5 & 4 & 161.68 (22.16) & 15.67 (0.72) & 0.17 (0.01) & 0.19 (0.02) \\ 
			1000 & 10 & 2 & 124.75 (12.86) & 13.33 (0.72) & 0.23 (0.02) & 0.41 (0.02) \\ 
			1000 & 10 & 3 & 103.95 (27.14) & 14.29 (0.58) & 0.24 (0.03) & 0.3 (0.03) \\ 
			% 1000 & 10 & 4 & 182.22 (59.41) & 21.62 (1.11) & 0.23 (0.04) & 0.25 (0.04) \\ 
			2000 & 5 & 2 & 234.79 (71.89) & 24.28 (1.42) & 0.18 (0.01) & 0.24 (0.02) \\ 
			2000 & 5 & 3 & 239.01 (48.26) & 28.42 (1.88) & 0.17 (0.01) & 0.19 (0.01) \\ 
			% 2000 & 5 & 4 & 359.53 (87.23) & 37.48 (1.79) & 0.17 (0.01) & 0.18 (0.02) \\ 
			2000 & 10 & 2 & 266.46 (60.08) & 35.39 (2.26) & 0.21 (0.02) & 0.29 (0.03) \\ 
			2000 & 10 & 3 & 266.57 (91.04) & 35.37 (1.38) & 0.22 (0.02) & 0.24 (0.01) \\ 
			% 2000 & 10 & 4 & 437.97 (136.8) & 61.44 (3.51) & 0.22 (0.03) & 0.23 (0.02) \\ 
			\bottomrule
		\end{tabular}
		\caption{\label{tab:running_time} Average running time in second and normalized tensor mean squared error (with standard error in the parenthesis) for the 10000-iteration and 1000-iteration projected gradient descent algorithm over 100 replicated experiments.}
	\end{table}

	Lastly, the cumulative regret and decision accuracy for model comparison are defined by:
	\begin{align*}
		&\text{regret}(\widehat{D}_{\text{opt}},D_{\text{opt}})=N^{-1}\sum_{i=1}^{N}\sum_{t=1}^{T}(P_{i,t,D_{i,\text{opt}}}-P_{i,t,\widehat{D}_{i,\text{opt}}}),\quad \text{acc}(\widehat{D}_{\text{opt}},D_{\text{opt}})=N^{-1}\sum_{i=1}^{N}\mathbf{1}(D_{i,\text{opt}}=\widehat{D}_{i,\text{opt}}).
	\end{align*}
	\subsection{Ablation analysis}
	In this section, we conduct the ablation studies to assess the effectiveness of our proposed framework. In particular, the mean squared error $\ell_2^{\text{group-w}}$ of the proposed framework is compared with other losses yielded by omitting three components individually:
	\begin{enumerate}
		\item [1)] The latent factor structures $\boldsymbol{U}_{1}$ and $\boldsymbol{U}_{2}$: compare with the loss $\ell_2^{\text{GLM-w}}$  yielded by the logistic regression model stratified by the true membership of the treatments. 
		\item [2)] The grouping structure $\boldsymbol{M}$: compare with the loss $\ell_2^{\text{factor-w}}$ yielded by the latent factor model with the membership matrix $\boldsymbol{M}$ replacing by a latent factor matrix $\boldsymbol{U}_3$.
		\item [3)] The inverse probability treatment weighting (IPTW): compare with the loss $\ell_2^{\text{group}}$  yielded by the unweighted minimization under the same latent factor block model.
	\end{enumerate}
	Table \ref{tab:ablation} presents the ablation studies with the smallest normalized tensor mean squared error bolded, leading to the following conclusions: 1) The comparison of the proposed framework versus others highlights the advantages of leveraging latent factors across units and time for estimation; 2) The IPTW noticeably improves the results when the sample size $N$ is small; 3) The hazard model adopting the grouping structure is particularly beneficial when the sample size is small. This is reasonable, as there may not be enough observations for certain treatments due to the limited sample size, and grouping treatments with homogeneous effects can enhance the estimation.
	
	\begin{table}[!ht]
		\centering
		\begin{tabular}{lllllll}
			\toprule
			$N$ & $T$ & $k$ & $\ell_2^{\text{GLM-w}}$ & $\ell_2^{\text{factor-w}}$ & $\ell_2^{\text{group}}$ & $\ell_2^{\text{group-w}}$ \\ 
			\midrule
			$100$ & 5 & 2 & 0.53 (0.02) & 0.58 (0.02) & 0.54 (0.03) & \textbf{0.42 (0.07)} \\ 
			100 & 10 & 2 & 0.54 (0.03) & 0.54 (0.03) & 0.51 (0.03) & \textbf{0.44 (0.06)} \\ 
			100 & 5 & 3 & 0.56 (0.02) & 0.62 (0.02) & 0.57 (0.03) & \textbf{0.31 (0.07)} \\ 
			100 & 10 & 3 & 0.58 (0.03) & 0.61 (0.03) & 0.57 (0.03) & \textbf{0.43 (0.09)} \\ 
			300 & 5 & 2 & 0.51 (0.01) & 0.54 (0.02) & 0.41 (0.03) & \textbf{0.24 (0.03)} \\ 
			300 & 10 & 2 & 0.53 (0.02) & 0.52 (0.02) & 0.43 (0.03) & \textbf{0.31 (0.05)} \\ 
			300 & 5 & 3 & 0.54 (0.01) & 0.51 (0.05) & 0.39 (0.03) & \textbf{0.21 (0.03)} \\ 
			300 & 10 & 3 & 0.58 (0.01) & 0.55 (0.04) & 0.42 (0.04) & \textbf{0.29 (0.05)} \\ 
			500 & 5 & 2 & 0.51 (0.01) & 0.48 (0.03) & 0.3 (0.03) & \textbf{0.20 (0.02)} \\ 
			500 & 10 & 2 & 0.52 (0.01) & 0.49 (0.03) & 0.33 (0.04) & \textbf{0.25 (0.04)} \\ 
			500 & 5 & 3 & 0.54 (0.01) & 0.37 (0.06) & 0.27 (0.03) & \textbf{0.20 (0.03)} \\ 
			500 & 10 & 3 & 0.57 (0.01) & 0.44 (0.07) & 0.31 (0.03) & \textbf{0.26 (0.04)} \\ 
			1000 & 5 & 2 & 0.51 (0.01) & 0.31 (0.03) & 0.21 (0.01) & \textbf{0.19 (0.01)} \\ 
			1000 & 10 & 2 & 0.52 (0.01) & 0.38 (0.04) & 0.24 (0.01) & \textbf{0.23 (0.02)} \\ 
			1000 & 5 & 3 & 0.54 (0.01) & \textbf{0.18 (0.02)} & 0.2 (0.01) & 0.19 (0.01) \\ 
			1000 & 10 & 3 & 0.57 (0.01) & \textbf{0.23 (0.04)} & 0.24 (0.01) & 0.24 (0.03) \\ 
			2000 & 5 & 2 & 0.5 (0.01) & \textbf{0.15 (0.02)} & 0.18 (0.01) & 0.18 (0.01) \\ 
			2000 & 10 & 2 & 0.52 (0.01) & \textbf{0.19 (0.03)} & 0.22 (0.01) & 0.21 (0.02) \\ 
			2000 & 5 & 3 & 0.53 (0.01) & \textbf{0.16 (0.02)} & 0.18 (0.01) & 0.17 (0.01) \\ 
			2000 & 10 & 3 & 0.57 (0.01) & \textbf{0.20 (0.03)} & 0.22 (0.01) & 0.22 (0.02) \\ 
			\bottomrule
		\end{tabular}
		\caption{\label{tab:ablation} Ablation analyses in terms of the normalized tensor mean squared error (with standard error in the parenthesis) over 100 replicated experiments}
	\end{table}

	\subsection{Additional real-data application}
	We provide an additional real-data application involving customer churn analysis of online retail to provide more empirical evidence. The data\footnote{https://www.kaggle.com/datasets/ankitverma2010/ecommerce-customer-churn-analysis-and-prediction} are collected by an E-commerce company. Five important indicators for customer churn are considered: total number of registered devices (0 if less than $3$, and 1 otherwise), the total number of used coupons in the last month (0 if less than the $50\%$ quantile and 1 otherwise), the distance between warehouse to the customer (0 if less than the $50\%$ quantile and 1 otherwise), whether the customer has complained or not, and post-complaint scores (0 for rates less than $3$, and 1 otherwise). Next, we consider ten customer characteristics $\boldsymbol{X}_i$, including gender, marital status, city tier, preferred login device, preferred payment method, preferred order category, the total number of added addresses, percentage increases in orders from the last year, days since last order and the average cashback in the last month. We evaluate the models using the same cross-validation scheme as in Section \ref{sec:application}, using the C-index and the average AUC to assess model performance. The best is \textbf{bolded}, and the second best is \underline{underlined}. Our proposed framework continues to perform well on this E-commerce dataset based on these performance metrics.  
	
	\begin{table}[!ht]
		\begin{centering}
			\begin{tabular}{lcc}
				\hline 
				& C-index ($\uparrow$) & average AUC ($\uparrow$) \tabularnewline
				\hline 
				logit & 0.33 $\pm$ 0.04 & 0.36 $\pm$ 0.03 \tabularnewline
				RF & \textbf{0.42 $\pm$ 0.01} & 0.37 $\pm$ 0.02 \tabularnewline
				NN & 0.40 $\pm$ 0.00 & 0.41 $\pm$ 0.01 \tabularnewline
				SVM & 0.27 $\pm$ 0.03 & \underline{0.38 $\pm$ 0.04}\\
				gradBoost & 0.37 $\pm$ 0.01 & 0.25 $\pm$ 0.01\\
				AdaBoost & 0.27 $\pm$ 0.02 & 0.37 $\pm$ 0.01 \\
				Vote & 0.38 $\pm$ 0.01& 0.22 $\pm$ 0.01 \\
				Cox-PH & 0.37 $\pm$ 0.01 & 0.37 $\pm$ 0.01 \tabularnewline
				surv-RF & 0.35 $\pm$ 0.01 & 0.33 $\pm$ 0.01 \tabularnewline
				Cox-gradBoost & 0.23 $\pm$ 0.01 & 0.18 $\pm$ 0.01 \tabularnewline
				Cox-modelBoost & 0.23 $\pm$ 0.01 & 0.26 $\pm$ 0.01 \tabularnewline
				\textbf{Proposed} & \underline{0.41 $\pm$ 0.03} & \textbf{0.44 $\pm$ 0.03} \tabularnewline
				\hline 
			\end{tabular}
			\par\end{centering}
		\caption{\label{tab:res2} Evaluation metrics on the E-commerce customer churn data.}
	\end{table}
	
	\section{Proofs}
	
	\subsection{Assumptions}
	
	We first assume some regularity conditions to proceed with our illustrations
	of the theoretical results. 
	\begin{itemize}
		\item[R1)]  (Positive Retention Probability) Let $p_{i,t,l}=P(\delta_{i}=1,\mathcal{Y}_{i,t,l}=1\mid\mathcal{Y}_{i,t-1,l}=1)$,
		we have $p_{\min}\leq p_{i,t,l}$ for any $i,t$ and $l$;
		\item[R2)]  (Incoherent Tensor Parameter) Suppose $\boldsymbol{U}_{1}$ and
		$\boldsymbol{U}_{2}$ have orthonormal columns, there exists some
		constant $\mu_{0}$ such that 
		\begin{align*}
			& \max\left\{ \frac{N}{r_{1}}\|\boldsymbol{U}_{1}\|_{2,\infty}^{2},\frac{T}{r_{2}}\|\boldsymbol{U}_{2}\|_{2,\infty}^{2}\right\} \leq\mu_{0},\quad\max_{k\in\{1,2,3\}}\|\mathcal{M}_{(k)}(\mathcal{S})\|\leq\|\Theta\|_{\max}\sqrt{\frac{NTL}{\mu_{0}^{3/2}(r_{1}r_{2}r_{3})^{1/2}}},
		\end{align*}
		where $\|\boldsymbol{U}\|_{2,\infty}^{2}=\max_{i}\|\boldsymbol{U}_{i,:}\|^{2}$
		and $\boldsymbol{U}_{i,:}$ is the $i$-th row vector of $\boldsymbol{U}$.
		\item[R3)]  (Non-degenerate Separation) The projected block means of $\Theta$
		is defined as $\mathcal{S}=\Theta\times_{1}\boldsymbol{U}_{1}^{\intercal}\times_{2}\boldsymbol{U}_{2}^{\intercal}\times_{3}\boldsymbol{W}^{\intercal}$,
		where 
		\[
		\Delta_{\min}^{2}=\Delta_{\min}(\mathcal{S})^{2}=\min_{i_{1}\neq i_{2}}\|\mathcal{M}_{(3)}(\mathcal{S})_{i_{1},:}-\mathcal{M}_{(3)}(\mathcal{S})_{i_{2},:}\|_{2}^{2}>0.
		\]
		\item[R4)]  (Balanced Clustering) There exists generic positive constants $c$
		and $C$ such that 
		\[
		cL/r_{3}\leq|\{l\in[L]:z_{l}=a\}|\leq CL/r_{3},\quad\forall a=1,\cdots,r_{3},
		\]
		where $|\cdot|$ represents the cardinality of a set.
	\end{itemize}
	Assumption R1) requires each unit to have a non-zero probability of maintaining the subscription at each time point $t$. This condition
	is necessary to establish the restricted strong convexity of the objective
	function $l(\Theta)$ in Section \ref{subsec:estimation}. Assumption
	R2) entails that the loading matrices $\boldsymbol{U}_{1}$ and $\boldsymbol{U}_{2}$
	should satisfy the incoherence condition, which is commonly imposed
	in the matrix/tensor completion literature \citep{candes2009exact,ma2017exploration,cao2020multisample,cai2021nonconvex}.
	In particular, this condition indicates that each tensor entry contains
	a similar amount of information so that missing any of them will not
	prevent us from being able to recover the entire tensor. In addition,
	we also require an upper bound on the spectral norm of each matricization
	of the core tensor $\mathcal{S}$, which leads to an entry-wise upper
	bound on the absolute value of $\Theta$ together with the incoherence
	conditions. Assumption R3) requires that the projected mode-$3$ slices
	$\mathcal{M}_{(3)}(\mathcal{S})$ should have distinct rows; otherwise
	the number of the clustering size should be reduced to smaller. Assumption
	R4) is imposed to control the spectral norm of $\boldsymbol{M}$,
	and is widely used in mixture model clustering literature \citep{gao2022iterative,han2022exact}.
	
	\subsection{Proof of Theorem \ref{thm:estimation}}
	
	The objective function for maximization is 
	\begin{align*}
		l(\Theta) & =\ensuremath{\sum_{i,t,l}w_{i}\mathbf{1}(\mathcal{Y}_{i,t-1,l}=1)\mathcal{Y}_{i,t,l}\log\{f(\theta_{i,t,l})}\}\\
		& +\ensuremath{\sum_{i,t,l}\delta_{i}w_{i}\mathbf{1}(\mathcal{Y}_{i,t-1,l}=1)(1-\mathcal{Y}_{i,t,l})\log\{1-f(\theta_{i,t,l})}\},
	\end{align*}
	where the link function $f(\theta)$ is monotonically increasing.
	We further assume that $f(\theta)+f(-\theta)=1$, $f'(\theta)=f'(-\theta)$
	and $f''(\theta)=-f''(-\theta)$. It follows from the expression of
	$l(\Theta)$ that 
	\begin{align*}
		\frac{\partial l(\Theta)}{\partial\theta_{i,t,l}} & =w_{i}\mathbf{1}(\mathcal{Y}_{i,t-1,l}=1)\mathcal{Y}_{i,t,l}\frac{f'(\theta_{i,t,l})}{f(\theta_{i,t,l})}-\delta_{i}w_{i}\mathbf{1}(\mathcal{Y}_{i,t-1,l}=1)(1-\mathcal{Y}_{i,t,l})\frac{f'(\theta_{i,t,l})}{1-f(\theta_{i,t,l})},\\
		\frac{\partial l^{2}(\Theta)}{\partial\theta_{i,t,l}^{2}} & =-w_{i}\mathbf{1}(\mathcal{Y}_{i,t-1,l}=1)\mathcal{Y}_{i,t,l}\left[\frac{\{f'(\theta_{i,t,l})\}^{2}}{f^{2}(\theta_{i,t,l})}-\frac{f''(\theta_{i,t,l})}{f(\theta_{i,t,l})}\right]\\
		& -\delta_{i}w_{i}\mathbf{1}(\mathcal{Y}_{i,t-1,l}=1)(1-\mathcal{Y}_{i,t,l})\left[\frac{f''(\theta_{i,t,l})}{1-f(\theta_{i,t,l})}+\frac{\{f'(\theta_{i,t,l})\}^{2}}{\{1-f(\theta_{i,t,l})\}^{2}}\right].
	\end{align*}
	Define 
	\[
	S(\Theta)=\left\llbracket \frac{\partial l(\Theta)}{\partial\theta_{i,t,l}}\right\rrbracket ,\quad H(\Theta)=\left\llbracket \frac{\partial l(\Theta)}{\partial\theta_{i,t,l}\partial\theta_{i',t',l'}}\right\rrbracket ,
	\]
	where $S(\Theta)$ and $H(\Theta)$ are the collection of the first
	and second derivatives of $l(\Theta)$. By the second-order Taylor's
	Theorem, we expand $l(\widehat{\Theta})$ around the true parameter
	$\Theta$ and obtain 
	\begin{equation}
		l(\widehat{\Theta})=l(\Theta)+\langle S(\Theta),\widehat{\Theta}-\Theta^{*}\rangle+\frac{1}{2}\text{vec}(\widehat{\Theta}-\Theta^{*})^{\intercal}H(\tilde{\Theta})\text{vec}(\widehat{\Theta}-\Theta^{*}),\label{eq:taylor-exp}
	\end{equation}
	where $\tilde{\Theta}=\gamma\Theta+(1-\gamma)\widehat{\Theta}$ for
	some $\gamma\in[0,1]$, and 
	\[
	\text{vec}(\widehat{\Theta}-\Theta^{*})^{\intercal}H(\tilde{\Theta})\text{vec}(\widehat{\Theta}-\Theta^{*})=\sum_{i,t,l}\left\{ \frac{\partial l^{2}(\Theta)}{\partial\theta_{i,t,l}^{2}}\bigg|_{\Theta=\tilde{\Theta}}\right\} (\widehat{\theta}_{i,t,l}-\theta_{i,t,l}^{*})^{2}.
	\]
	Let 
	\[
	L_{\alpha}=\sup_{|\theta|\leq\alpha}\left[\frac{f'(\theta_{i,t,l})}{f(\theta_{i,t,l})},\frac{f'(\theta_{i,t,l})}{1-f(\theta_{i,t,l})}\right],\quad\gamma_{\alpha}=\inf_{|\theta|\leq\alpha}\left[\frac{\{f'(\theta_{i,t,l})\}^{2}}{f^{2}(\theta_{i,t,l})}-\frac{f''(\theta_{i,t,l})}{f(\theta_{i,t,l})},\frac{f''(\theta_{i,t,l})}{1-f(\theta_{i,t,l})}+\frac{\{f'(\theta_{i,t,l})\}^{2}}{\{1-f(\theta_{i,t,l})\}^{2}}\right],
	\]
	where $\alpha=\|\Theta\|_{\max}$ is the bound on the entry-wise magnitude
	of $\Theta$. First, we bound the quadratic term in (\ref{eq:taylor-exp})
	as:
	
	\[
	\sum_{i,t,l}\left\{ \frac{\partial l^{2}(\Theta)}{\partial\theta_{i,t,l}^{2}}\bigg|_{\Theta=\tilde{\Theta}}\right\} (\widehat{\theta}_{i,t,l}-\theta_{i,t,l}^{*})^{2}\leq-\gamma_{\alpha}w_{\min}\sum_{i,t,l}\delta_{i}\mathbf{1}(\mathcal{Y}_{i,t-1,l}=1)(\widehat{\theta}_{i,t,l}-\theta_{i,t,l}^{*})^{2}.
	\]
	
	\begin{lemma}
		\label{lem:restricted_convex} Under the same conditions in Theorem
		\ref{thm:estimation} and $\|\widehat{\Theta}-\Theta\|_{F}^{2}\geq C_{0}\|\Theta\|_{\max}^{2}T\log(N+T+K)/p_{\min}$,
		there exists an constant $c_{0}$, such that, with probability greater
		than $1-c_{0}(N+T+K)^{-2}$,
		\[
		\sum_{i,t,l}\delta_{i}\mathbf{1}(\mathcal{Y}_{i,t-1,l}=1)(\widehat{\theta}_{i,t,l}-\theta_{i,t,l}^{*})^{2}\geq\frac{p_{\min}}{2}\|\Delta_{\Theta}\|_{F}^{2}-2\vartheta,
		\]
		where 
		\[
		\vartheta=\frac{C''r_{1}r_{2}r_{3}(N\vee T)}{\max(r_{1},r_{2},r_{3})p_{\min}}\|\Theta\|_{\max}^{2}\log(N+T+K).
		\]
	\end{lemma}
	
	By Lemma \ref{lem:restricted_convex}, we have
	\[
	l(\widehat{\Theta})\leq l(\Theta)+\langle S(\Theta),\widehat{\Theta}-\Theta\rangle-\frac{\gamma_{\alpha}w_{\min}p_{\min}}{2}\|\widehat{\Theta}-\Theta\|_{F}^{2}+2\gamma_{\alpha}w_{\min}\vartheta,
	\]
	where $\gamma_{\alpha}w_{\min}p_{\min}/2$ is curvature of $l(\Theta)$,
	and $\gamma_{\alpha}w_{\min}\vartheta$ is the tolerance term induced
	by the Rademacher complexity of $l(\Theta)$. Next, we bound the linear
	term in (\ref{eq:taylor-exp}) as:
	\begin{align*}
		|\langle S(\Theta),\widehat{\Theta}-\Theta\rangle| & \leq\|S(\Theta)\|\cdot\|\widehat{\Theta}-\Theta\|_{*}\\
		& \leq\|S(\Theta)\|\cdot2\sqrt{\frac{r_{1}r_{2}r_{3}}{\max(r_{1},r_{2},r_{3})}}\|\widehat{\Theta}-\Theta\|_{F},
	\end{align*}
	where the last inequality is justified by Lemma 1, \citet{wang2020learning}.
	\begin{lemma}
		\label{lem:tensor_bernstein} (Tensor Bernstein Inequality, Theorem
		4.3, \citet{luo2020tensor}) Let $Z_{1},\cdots,Z_{N}$ be independent
		tensor in $\mathbb{R}^{d\times d\times d}$, such that $\mathbb{E}\left(Z_{i}\right)=0$
		and $\|Z_{i}\|\leq D_{Z}$ for all $i=1,\cdots,N$. Let $\sigma_{Z}^{2}$
		be such that
		\[
		\sigma_{Z}^{2}\geq\max\left\{ \left\Vert \mathbb{E}\left(\sum_{i=1}^{N}Z_{i}\overline{\square}\sum_{i=1}^{N}Z_{i}\right)\right\Vert ,\left\Vert \mathbb{E}\left(\sum_{i=1}^{N}Z_{i}\underline{\square}\sum_{i=1}^{N}Z_{i}\right)\right\Vert \right\} .
		\]
		Then for any $\alpha\geq0$
		\[
		\mathbb{P}\left(\left\Vert \sum_{i=1}^{N}Z_{i}\right\Vert ^{\overline{\square}}\geq\alpha\right)\leq d^{2}\exp\left\{ \frac{-\alpha^{2}}{2\sigma_{Z}^{2}+\left(2D_{Z}\alpha\right)/3}\right\} ,
		\]
		where $\overline{\square}$ and $\underline{\square}$ are two generalized
		Einstein products of tensors.
	\end{lemma}
	
	From Lemma \ref{lem:tensor_bernstein}, we know
	\[
	P(\|S(\Theta)\|\geq\alpha)\leq(N+T+L)^{2}\exp\left\{ -\frac{\alpha^{2}}{2\sigma_{S}^{2}+(2D_{S}\alpha)/3}\right\} ,
	\]
	where 
	\[
	D_{S}=w_{\max}L_{\alpha}T^{1/2}\log(N+T+L),\quad\sigma_{S}^{2}=w_{\max}^{2}L_{\alpha}(N\vee T)\log(N+T+L).
	\]
	Hence, implies that 
	\[
	\|S(\Theta)\|\leq w_{\max}\sqrt{L_{\alpha}(N\vee T)}\log(N+T+L)
	\]
	holds with probability greater than $1-c_{0}(N+T+L)^{-2}$. Since
	$\widehat{\Theta}$ is the local maximizer, i.e., $\widehat{\Theta}=\arg\max_{\Theta}l(\Theta)$,
	we have 
	\[
	0\leq l(\widehat{\Theta})-l(\Theta)\leq\langle S(\Theta),\widehat{\Theta}-\Theta\rangle-\frac{\gamma_{\alpha}w_{\min}p_{\min}}{2}\|\widehat{\Theta}-\Theta\|_{F}^{2}+2\gamma_{\alpha}w_{\min}\vartheta,
	\]
	which it gives us 
	\begin{align*}
		\|\widehat{\Theta}-\Theta\|_{F}^{2} & \leq\frac{C_{1}}{\gamma_{\alpha}w_{\min}p_{\min}}\|S(\Theta)\|\cdot\|\widehat{\Theta}-\Theta\|_{*}\\
		& +\frac{C_{2}r_{1}r_{2}r_{3}(N\vee T)}{\max(r_{1},r_{2},r_{3})p_{\min}^{2}}\|\Theta\|_{\max}^{2}\log(N+T+L)\\
		& \leq\frac{C_{1}}{\gamma_{\alpha}w_{\min}p_{\min}}\|S(\Theta)\|\cdot\sqrt{\frac{r_{1}r_{2}r_{3}}{\max(r_{1},r_{2},r_{3})}}\|\widehat{\Theta}-\Theta\|_{F}\\
		& +\frac{C_{2}r_{1}r_{2}r_{3}(N\vee T)}{\max(r_{1},r_{2},r_{3})p_{\min}^{2}}\|\Theta\|_{\max}^{2}\log(N+T+L).
	\end{align*}
	Therefore, with at least $1-c_{0}(N+T+K)^{-2}$, we have $ab\leq(a^{2}+b^{2})/2$
	and
	\begin{align*}
		\|\widehat{\Theta}-\Theta\|_{F}^{2} & \leq\frac{C_{1}w_{\max}L_{\alpha}}{\gamma_{\alpha}w_{\min}p_{\min}}\sqrt{\frac{r_{1}r_{2}r_{3}(N\vee T)}{\max(r_{1},r_{2},r_{3})}}\log(N+T+L)\cdot\|\widehat{\Theta}-\Theta\|_{F}\\
		& +\frac{C_{2}r_{1}r_{2}r_{3}(N\vee T)}{p_{\min}^{2}\max(r_{1},r_{2},r_{3})}\|\Theta\|_{\max}^{2}\log(N+T+L)\\
		& \leq\frac{C_{1}^{'}w_{\max}^{2}L_{\alpha}^{2}}{\gamma_{\alpha}^{2}w_{\min}^{2}p_{\min}^{2}}\frac{r_{1}r_{2}r_{3}(N\vee T)}{\max(r_{1},r_{2},r_{3})}\log^{2}(N+T+L)+\frac{\|\widehat{\Theta}-\Theta\|_{F}^{2}}{2}\\
		& +\frac{C_{2}r_{1}r_{2}r_{3}(N\vee T)}{p_{\min}^{2}\max(r_{1},r_{2},r_{3})}\|\Theta\|_{\max}^{2}\log(N+T+L).
	\end{align*}
	To sum up, we conclude, with probability $1-c_{0}(N+T+L)^{-2}$, we
	have
	\begin{align*}
		\|\widehat{\Theta}-\Theta\|_{F}^{2} & \leq C_{0}\|\Theta\|_{\max}^{2}\frac{T}{p_{\min}}\log(N+T+L)\\
		& \vee C_{1}\frac{L_{\alpha}^{2}}{\gamma_{\alpha}^{2}}\frac{w_{\max}^{2}}{w_{\min}^{2}}\frac{r_{1}r_{2}r_{3}(N\vee T)}{p_{\min}^{2}\max(r_{1},r_{2},r_{3})}\log^{2}(N+T+L)\\
		& \vee C_{2}\frac{r_{1}r_{2}r_{3}(N\vee T)}{p_{\min}^{2}\max(r_{1},r_{2},r_{3})}\|\Theta\|_{\max}^{2}\log(N+T+L).
	\end{align*}

	\subsection{Proof of Theorem \ref{thm:clustering}}
	
	The proof of Theorem \ref{thm:clustering} is divided into several
	steps to ease the understanding. 
	
	\subsubsection{Step 1}
	
	We lay out some notations and technical Lemmas that will be useful
	for our main proof. First, we define the normalized membership matrices
	$\boldsymbol{W}=\boldsymbol{M}(\text{diag}(\mathbf{1}_{L}^{\intercal}\boldsymbol{M}))^{-1}$.
	Next, we define the estimators of the projected block mean $\mathcal{S}$
	and the projected mode-$3$ slices $\mathcal{F}$ by
	\begin{align*}
		& \mathcal{F}=\Theta\times_{1}\boldsymbol{U}_{1}^{\intercal}\times_{2}\boldsymbol{U}_{2}^{\intercal},\quad\mathcal{S}=\Theta\times_{1}\boldsymbol{U}_{1}^{\intercal}\times_{2}\boldsymbol{U}_{2}^{\intercal}\times_{3}\boldsymbol{W}^{\intercal},\\
		& \widehat{\mathcal{F}}=\widehat{\Theta}\times_{1}(\widehat{\boldsymbol{U}}_{1})^{\intercal}\times_{2}(\widehat{\boldsymbol{U}}_{2})^{\intercal},\quad\mathcal{\widehat{S}}=\widehat{\Theta}\times_{1}(\widehat{\boldsymbol{U}}_{1})^{\intercal}\times_{2}(\widehat{\boldsymbol{U}}_{2})^{\intercal}\times_{3}(\widehat{\boldsymbol{W}})^{\intercal}\\
		& \tilde{\mathcal{F}}=\widehat{\Theta}\times_{1}\boldsymbol{U}_{1}^{\intercal}\times_{2}\boldsymbol{U}_{2}^{\intercal},\quad\mathcal{\tilde{S}}=\widehat{\Theta}\times_{1}\boldsymbol{U}_{1}^{\intercal}\times_{2}\boldsymbol{U}_{2}^{\intercal}\times_{3}\boldsymbol{W}^{\intercal}.
	\end{align*}
	Next, we define the matricizations of each tensor $(\mathcal{S},\mathcal{F})$
	for the mode $3$ by 
	\begin{align*}
		& \boldsymbol{S}=\mathcal{M}_{(3)}(\mathcal{S}),\quad\boldsymbol{\widehat{S}}=\mathcal{M}_{(3)}(\mathcal{\widehat{S}}),\quad\tilde{\boldsymbol{S}}=\mathcal{M}_{(3)}(\mathcal{\tilde{S}}),\\
		& \boldsymbol{F}=\mathcal{M}_{(3)}(\mathcal{F})=\mathcal{M}_{(3)}(\Theta)(\boldsymbol{U}_{1}\otimes\boldsymbol{U}_{2})=\mathcal{M}_{(3)}(\Theta)\boldsymbol{V},\\
		& \widehat{\boldsymbol{F}}=\mathcal{M}_{(3)}(\mathcal{\widehat{F}})=\mathcal{M}_{(3)}(\widehat{\Theta})(\widehat{\boldsymbol{U}}_{1}\otimes\widehat{\boldsymbol{U}}_{2})=\mathcal{M}_{(3)}(\widehat{\Theta})\widehat{\boldsymbol{V}},\\
		& \tilde{\boldsymbol{F}}=\mathcal{M}_{(3)}(\mathcal{\tilde{F}})=\mathcal{M}_{(3)}(\widehat{\Theta})(\boldsymbol{U}_{1}\otimes\boldsymbol{U}_{2})=\mathcal{M}_{(3)}(\widehat{\Theta})\boldsymbol{V},
	\end{align*}
	where $\boldsymbol{V}=\boldsymbol{U}_{1}\otimes\boldsymbol{U}_{2}$
	and $\widehat{\boldsymbol{V}}=\widehat{\boldsymbol{U}}_{1}\otimes\widehat{\boldsymbol{U}}_{2}$. 
	\begin{lemma}
		\label{lem:theta_hat_V} Under the same assumptions in Theorem \ref{thm:estimation},
		we have 
		\begin{align}
			& \|(\boldsymbol{W}_{:,b}-\widehat{\boldsymbol{W}}_{:,b})^{\intercal}\mathcal{M}_{(3)}(\widehat{\Theta})\boldsymbol{V}\|\lesssim\frac{r_{3}g(\widehat{z},z)}{\Delta_{\min}}+\frac{\mu_{0}r_{1}^{1/2}r_{2}^{1/2}r_{3}^{3/2}g(\widehat{z},z)\|\widehat{\Theta}-\Theta\|_{F}}{\Delta_{\min}^{2}\sqrt{NTL}},\label{eq:error_W_V}\\
			& \|(\boldsymbol{W}_{:,b}-\widehat{\boldsymbol{W}}_{:,b})^{\intercal}\mathcal{M}_{(3)}(\widehat{\Theta})\widehat{\boldsymbol{V}}\|\lesssim\frac{r_{3}g(\widehat{z},z)}{\Delta_{\min}}+\frac{\mu_{0}r_{1}^{1/2}r_{2}^{1/2}r_{3}^{3/2}g(\widehat{z},z)\|\widehat{\Theta}-\Theta\|_{F}}{\Delta_{\min}^{2}\sqrt{NTL}}+\frac{\kappa^{2}r_{3}^{3/2}g(\widehat{z},z)\|\widehat{\Theta}-\Theta\|_{F}}{\Delta_{\min}^{2}\sqrt{L}},\label{eq:error_W_Vhat}\\
			& \|\widehat{\boldsymbol{W}}_{:,b}^{\intercal}\mathcal{M}_{(3)}(\widehat{\Theta})(\boldsymbol{V}-\widehat{\boldsymbol{V}})\|\lesssim\kappa^{2}\sqrt{r_{3}/L}\|\widehat{\Theta}-\Theta\|_{F},\label{eq:W_error_V}\\
			& g(\widehat{z},z)\lesssim\Delta_{\min}^{2}/r_{3}\label{eq:loss_condition}\\
			& \sqrt{r_{3}/L}\lesssim\lambda_{r_{3}}(\boldsymbol{W})\lesssim\|\boldsymbol{W}\|\lesssim\sqrt{r_{3}/L},\label{eq:eigen_W}\\
			& \sqrt{L/r_{3}}\lesssim\lambda_{r_{3}}(\boldsymbol{M})\lesssim\|\boldsymbol{M}\|\lesssim\sqrt{L/r_{3}},\label{eq:eigen_M}\\
			& \|\widehat{\boldsymbol{V}}-\boldsymbol{V}\|\lesssim\frac{\kappa\|\widehat{\Theta}-\Theta\|_{F}}{\lambda_{\min}},\label{eq:Vhat-V}\\
			& \|\widehat{\boldsymbol{V}}-\boldsymbol{V}\|_{2,\max}\lesssim\frac{\kappa\|\widehat{\Theta}-\Theta\|_{F}}{\min(\delta_{r_{1}},\delta_{r_{2}})}.\label{eq:Vhat-V_2max}
		\end{align}
	\end{lemma}

	\subsubsection{Step 2}
	
	In our nearest neighbor search algorithm, the estimated clustering
	label vector $\widehat{z}$ should satisfy: 
	\[
	\widehat{z}_{l}=\arg\min_{a\in[r_{3}]}\|\boldsymbol{\widehat{F}}_{l,:}-\boldsymbol{\widehat{S}}_{a,:}\|_{2}^{2},
	\]
	for $l=1,\cdots,L$. Therefore, it is important to analyze the probably
	of the event:
	\begin{equation}
		\mathbf{1}(\widehat{z}_{l}=b)=\mathbf{1}\left(\widehat{z}_{l}=b,\|\boldsymbol{\widehat{F}}_{l,:}-\boldsymbol{\widehat{S}}_{b,:}\|_{2}^{2}\leq\|\boldsymbol{\widehat{F}}_{l,:}-\boldsymbol{\widehat{S}}_{(z)_{l},:}\|_{2}^{2}\right).\label{eq:event}
	\end{equation}
	Assume $z_{l}=a$, we can check $\|\boldsymbol{\widehat{F}}_{l,:}-\boldsymbol{\widehat{S}}_{b,:}\|_{2}^{2}\leq\|\boldsymbol{\widehat{F}}_{l,:}-\boldsymbol{\widehat{S}}_{a,:}\|_{2}^{2}$
	is equivalent to 
	\begin{align*}
		2\langle\mathcal{M}_{(3)}(\widehat{\Theta}-\Theta)_{l,:}\boldsymbol{V},\tilde{\boldsymbol{S}}_{a,:}-\tilde{\boldsymbol{S}}_{b,:}\rangle & \leq-\|\boldsymbol{S}_{a,:}-\boldsymbol{S}_{b,:}\|^{2}+F_{l}(a,b;\widehat{z})+G_{l}(a,b;\widehat{z})+H_{l}(a,b),
	\end{align*}
	where 
	\begin{align*}
		\widehat{F}_{l}=F_{l}(a,b;\widehat{z}) & =2\langle\mathcal{M}_{(3)}(\widehat{\Theta}-\Theta)_{l,:}\widehat{\boldsymbol{V}},(\tilde{\boldsymbol{S}}_{a,:}-\widehat{\boldsymbol{S}}_{a,:})-(\tilde{\boldsymbol{S}}_{b,:}-\widehat{\boldsymbol{S}}_{b,:})\rangle\\
		& +2\langle\mathcal{M}_{(3)}(\widehat{\Theta}-\Theta)_{l,:}(\boldsymbol{V}-\widehat{\boldsymbol{V}}),\tilde{\boldsymbol{S}}_{a,:}-\tilde{\boldsymbol{S}}_{b,:}\rangle,\\
		\widehat{G}_{l}=G_{l}(a,b;\widehat{z}) & =\left(\|\Theta_{l:}\widehat{\boldsymbol{V}}-\widehat{\boldsymbol{S}}_{a,:}\|_{F}^{2}-\|\Theta_{l:}\widehat{\boldsymbol{V}}-\boldsymbol{W}_{:,a}^{\intercal}\widehat{\Theta}\widehat{\boldsymbol{V}}\|_{F}^{2}\right)\\
		& -\left(\|\Theta_{l:}\widehat{\boldsymbol{V}}-\widehat{\boldsymbol{S}}_{b,:}\|_{F}^{2}-\|\Theta_{l:}\widehat{\boldsymbol{V}}-\boldsymbol{W}_{:,b}^{\intercal}\widehat{\Theta}\widehat{\boldsymbol{V}}\|_{F}^{2}\right),\\
		\widehat{H}_{l}=H_{l}(a,b) & =\|\Theta_{l:}\widehat{\boldsymbol{V}}-\boldsymbol{W}_{:,a}^{\intercal}\widehat{\Theta}\widehat{\boldsymbol{V}}\|_{F}^{2}-\|\Theta_{l:}\widehat{\boldsymbol{V}}-\boldsymbol{W}_{:,b}^{\intercal}\widehat{\Theta}\widehat{\boldsymbol{V}}\|_{F}^{2}\\
		& +\|\boldsymbol{S}_{a,:}-\boldsymbol{S}_{b,:}\|^{2}.
	\end{align*}
	From these three error terms, the first two $F_{l}(a,b;\widehat{z})$
	and $G_{l}(a,b;\widehat{z})$ are controlled by the differences between
	$(\widehat{\mathcal{S}},\widehat{\mathcal{F}})$ and $(\tilde{\mathcal{S}},\mathcal{\tilde{F}})$,
	the last one $H_{l}(a,b)$ is controlled by the differences between
	$(\tilde{\mathcal{S}},\mathcal{\tilde{F}})$ and $(\mathcal{S},\mathcal{F})$.
	If we ignore all these error terms, the event $2\langle\mathcal{M}_{(3)}(\widehat{\Theta}-\Theta)_{l,:}\boldsymbol{V},\tilde{\boldsymbol{S}}_{a,:}-\tilde{\boldsymbol{S}}_{b,:}\rangle\leq-\|\boldsymbol{S}_{a,:}-\boldsymbol{S}_{b,:}\|^{2}$
	will constitute the oracle statistical loss after applying our algorithm
	if we are given the true label vector $z$. Then, we can show that
	\begin{align*}
		(\widehat{z}_{l}=b) & \subset\left\{ \langle\mathcal{M}_{(3)}(\widehat{\Theta}-\Theta)_{l,:}\boldsymbol{V},\tilde{\boldsymbol{S}}_{a,:}-\tilde{\boldsymbol{S}}_{b,:}\rangle\leq-\frac{1}{4}\|\boldsymbol{S}_{a,:}-\boldsymbol{S}_{b,:}\|^{2}\right\} \\
		& \bigcup\left\{ \widehat{z}_{l}=b,\frac{1}{2}\|\boldsymbol{S}_{a,:}-\boldsymbol{S}_{b,:}\|^{2}\leq\widehat{F}_{l}+\widehat{G}_{l}+\widehat{H}_{l}\right\} .
	\end{align*}
	Recall the definition of $g(a,b)$, the misclassification loss for
	the estimated clustering label vector $\widehat{z}$ versus the truth
	will be:
	\begin{align*}
		g(\widehat{z},z) & =\min_{\pi\in\Pi_{r_{3}}}\frac{1}{L}\sum_{l=1}^{L}\|\boldsymbol{S}_{a,:}-\boldsymbol{S}_{\pi(\widehat{z})_{l,:}}\|_{2}^{2}\\
		& =\min_{\pi\in\Pi_{r_{3}}}\frac{1}{L}\sum_{l=1}^{L}\sum_{b=1}^{r_{3}}\mathbf{1}\{\pi(\widehat{z})_{l}=b\}\|\boldsymbol{S}_{a,:}-\boldsymbol{S}_{b,:}\|_{2}^{2}\\
		& \leq\min_{\pi\in\Pi_{r_{3}}}\frac{1}{L}\sum_{l=1}^{L}\sum_{b=1}^{r_{3}}\left\{ \mathbf{1}(\mathcal{E}_{1})+\mathbf{1}(\mathcal{E}_{2})\right\} \times\|\boldsymbol{S}_{a,:}-\boldsymbol{S}_{b,:}\|_{2}^{2}\\
		& =\widehat{g}_{1}+\widehat{g}_{2},
	\end{align*}
	where 
	\begin{align*}
		\mathcal{E}_{1} & =\left\{ \langle\mathcal{M}_{(3)}(\widehat{\Theta}-\Theta)_{l,:}\boldsymbol{V},\tilde{\boldsymbol{S}}_{a,:}-\tilde{\boldsymbol{S}}_{b,:}\rangle\leq-\frac{1}{4}\|\boldsymbol{S}_{a,:}-\boldsymbol{S}_{b,:}\|^{2}\right\} ,\\
		\mathcal{E}_{2} & =\left\{ \widehat{z}_{l}=b,\frac{1}{2}\|\boldsymbol{S}_{a,:}-\boldsymbol{S}_{b,:}\|_{2}^{2}\leq\widehat{F}_{l}+\widehat{G}_{l}+\widehat{H}_{l}\right\} .
	\end{align*}
	The following steps (Step 3 and Step 4) aim to bound $\widehat{g}_{1}$
	and $\widehat{g}_{2}$, respectively.
	
	\subsubsection{Step 3}
	
	In this step, we focus on proving the upper bound for $\widehat{g}_{1}=\min_{\pi\in\Pi_{r_{3}}}\sum_{l=1}^{L}\sum_{b=1}^{r_{3}}\mathbf{1}(\mathcal{E}_{1})\|\boldsymbol{S}_{a,:}-\boldsymbol{S}_{b,:}\|_{2}^{2}/L$.
	Recall that 
	\[
	\mathcal{E}_{1}=\left\{ \langle\mathcal{M}_{(3)}(\widehat{\Theta}-\Theta)_{l,:}\boldsymbol{V},\tilde{\boldsymbol{S}}_{a,:}-\tilde{\boldsymbol{S}}_{b,:}\rangle\leq-\frac{1}{4}\|\boldsymbol{S}_{a,:}-\boldsymbol{S}_{b,:}\|^{2}\right\} ,
	\]
	we can decompose the probability of event $\mathcal{E}_{1}$ into
	three parts:
	\begin{align*}
		& P\left(\langle\mathcal{M}_{(3)}(\widehat{\Theta}-\Theta)_{l,:}\boldsymbol{V},\tilde{\boldsymbol{S}}_{a,:}-\tilde{\boldsymbol{S}}_{b,:}\rangle\leq-\frac{1}{4}\|\boldsymbol{S}_{a,:}-\boldsymbol{S}_{b,:}\|^{2}\right)\\
		\leq & P\left(\langle\mathcal{M}_{(3)}(\widehat{\Theta}-\Theta)_{l,:}\boldsymbol{V},\boldsymbol{S}_{a,:}-\boldsymbol{S}_{b,:}\rangle\leq-\frac{1}{8}\|\boldsymbol{S}_{a,:}-\boldsymbol{S}_{b,:}\|^{2}\right)\\
		& +P\left(\langle\mathcal{M}_{(3)}(\widehat{\Theta}-\Theta)_{l,:}\boldsymbol{V},\tilde{\boldsymbol{S}}_{a,:}-\boldsymbol{S}_{a,:}\rangle\leq-\frac{1}{16}\|\boldsymbol{S}_{a,:}-\boldsymbol{S}_{b,:}\|^{2}\right)\\
		& +P\left(\langle\mathcal{M}_{(3)}(\widehat{\Theta}-\Theta)_{l,:}\boldsymbol{V},\boldsymbol{S}_{b,:}-\tilde{\boldsymbol{S}}_{b,:}\rangle\leq-\frac{1}{16}\|\boldsymbol{S}_{a,:}-\boldsymbol{S}_{b,:}\|^{2}\right).
	\end{align*}
	
	\begin{lemma}
		\label{lem:E1_decomposed} Under the same assumptions in Theorem \ref{thm:estimation},
		if the signal-to-noise ratio satisfies the condition in Theorem \ref{thm:clustering},
		there exist generic constants $c_{1}$ and $c_{2}$ such that
		\begin{align*}
			& P\left(\langle\mathcal{M}_{(3)}(\widehat{\Theta}-\Theta)_{l,:}\boldsymbol{V},\boldsymbol{S}_{a,:}-\boldsymbol{S}_{b,:}\rangle\leq-\frac{1}{8}\|\boldsymbol{S}_{a,:}-\boldsymbol{S}_{b,:}\|^{2}\right)\leq\frac{c_{1}}{(N+T+L)^{2}},\\
			& P\left(\langle\mathcal{M}_{(3)}(\widehat{\Theta}-\Theta)_{l,:}\boldsymbol{V},\tilde{\boldsymbol{S}}_{a,:}-\boldsymbol{S}_{a,:}\rangle\leq-\frac{1}{8}\|\boldsymbol{S}_{a,:}-\boldsymbol{S}_{b,:}\|^{2}\right)\leq\frac{c_{1}}{(N+T+L)^{2}}.
		\end{align*}
	\end{lemma}
	
	With Lemma \ref{lem:E1_decomposed}, we can bound the expectation
	of $\widehat{g}_{1}$ if the signal-to-noise ratio satisfies the condition
	in Theorem \ref{thm:clustering}: 
	\begin{align*}
		\E\widehat{g}_{1} & =\min_{\pi\in\Pi_{r_{3}}}\frac{1}{L}\sum_{l=1}^{L}\sum_{b\in[r_{3}]/a}P\left(\langle\mathcal{M}_{(3)}(\widehat{\Theta}-\Theta)_{l,:}\boldsymbol{V},\tilde{\boldsymbol{S}}_{a,:}-\tilde{\boldsymbol{S}}_{b,:}\rangle\leq-\frac{1}{4}\|\boldsymbol{S}_{a,:}-\boldsymbol{S}_{b,:}\|^{2}\right)\cdot\|\boldsymbol{S}_{a,:}-\boldsymbol{S}_{b,:}\|_{2}^{2}\\
		& \lesssim\min_{\pi\in\Pi_{r_{3}}}\frac{1}{L}\sum_{l=1}^{L}\sum_{b\in[r_{3}]/a}\exp\left\{ -\frac{w_{\min}^{2}p_{\min}^{2}NTL\max(r_{1},r_{2},r_{3})\Delta_{\min}^{2}}{w_{\max}^{2}(N\vee T)r_{1}^{2}r_{2}^{2}r_{3}\mu_{0}^{2}\{\|\Theta\|_{\max}^{2}\vee(L_{\alpha}^{2}/\gamma_{\alpha}^{2})\}}\right\} \cdot\|\boldsymbol{S}_{a,:}-\boldsymbol{S}_{b,:}\|_{2}^{2}.
	\end{align*}
	Since $\Delta_{\min}^{2}\leq\|\boldsymbol{S}_{a,:}-\boldsymbol{S}_{b,:}\|^{2}$
	for any $a,b\in[r_{3}]$, it implies that
	\begin{align*}
		& \sum_{l=1}^{L}\sum_{b\in[r_{3}]/a}\|\boldsymbol{S}_{a,:}-\boldsymbol{S}_{b,:}\|_{2}^{2}\cdot\exp\left\{ -\frac{w_{\min}^{2}p_{\min}^{2}NTL\max(r_{1},r_{2},r_{3})\Delta_{\min}^{2}}{w_{\max}^{2}(N\vee T)r_{1}^{2}r_{2}^{2}r_{3}\mu_{0}^{2}\{\|\Theta\|_{\max}^{2}\vee(L_{\alpha}^{2}/\gamma_{\alpha}^{2})\}}\right\} \\
		& \lesssim\{\|\Theta\|_{\max}^{2}\vee(L_{\alpha}^{2}/\gamma_{\alpha}^{2})\}\sum_{l=1}^{L}\sum_{b\in[r_{3}]/a}\frac{\|\boldsymbol{S}_{a,:}-\boldsymbol{S}_{b,:}\|_{2}^{2}}{\|\Theta\|_{\max}^{2}\vee(L_{\alpha}^{2}/\gamma_{\alpha}^{2})}\cdot\exp\left\{ -\frac{w_{\min}^{2}p_{\min}^{2}NTL\max(r_{1},r_{2},r_{3})\Delta_{\min}^{2}}{w_{\max}^{2}(N\vee T)r_{1}^{2}r_{2}^{2}r_{3}\mu_{0}^{2}\{\|\Theta\|_{\max}^{2}\vee(L_{\alpha}^{2}/\gamma_{\alpha}^{2})\}}\right\} \\
		& \lesssim\{\|\Theta\|_{\max}^{2}\vee(L_{\alpha}^{2}/\gamma_{\alpha}^{2})\}\sum_{l=1}^{L}\sum_{b\in[r_{3}]/a}\exp\left\{ -\frac{w_{\min}^{2}p_{\min}^{2}NTL\max(r_{1},r_{2},r_{3})\Delta_{\min}^{2}}{w_{\max}^{2}(N\vee T)r_{1}^{2}r_{2}^{2}r_{3}\mu_{0}^{2}\{\|\Theta\|_{\max}^{2}\vee(L_{\alpha}^{2}/\gamma_{\alpha}^{2})\}}\right\} \\
		& \lesssim\{\|\Theta\|_{\max}^{2}\vee(L_{\alpha}^{2}/\gamma_{\alpha}^{2})\}L\exp\left\{ -\frac{w_{\min}^{2}p_{\min}^{2}NTL\max(r_{1},r_{2},r_{3})\Delta_{\min}^{2}}{w_{\max}^{2}(N\vee T)r_{1}^{2}r_{2}^{2}r_{3}\mu_{0}^{2}\{\|\Theta\|_{\max}^{2}\vee(L_{\alpha}^{2}/\gamma_{\alpha}^{2})\}}\right\} .
	\end{align*}
	By Markov inequality, it yields that
	\begin{align*}
		& P\left[\widehat{g}_{1}\leq\E\widehat{g}_{1}\cdot\exp\left\{ \frac{c_{0}}{2}\frac{w_{\min}^{2}p_{\min}^{2}NTL\max(r_{1},r_{2},r_{3})\Delta_{\min}^{2}}{w_{\max}^{2}(N\vee T)r_{1}^{2}r_{2}^{2}r_{3}\mu_{0}^{2}\{\|\Theta\|_{\max}^{2}\vee(L_{\alpha}^{2}/\gamma_{\alpha}^{2})\}}\right\} \right]\\
		& \geq1-\exp\left\{ -\frac{c_{0}}{2}\frac{w_{\min}^{2}p_{\min}^{2}NTL\max(r_{1},r_{2},r_{3})\Delta_{\min}^{2}}{w_{\max}^{2}(N\vee T)r_{1}^{2}r_{2}^{2}r_{3}\mu_{0}^{2}\{\|\Theta\|_{\max}^{2}\vee(L_{\alpha}^{2}/\gamma_{\alpha}^{2})\}}\right\} .
	\end{align*}
	Therefore, with probability at least $1-c(N+T+L)^{-2}$, we have 
	\[
	\widehat{g}_{1}\lesssim\{\|\Theta\|_{\max}^{2}\vee(L_{\alpha}^{2}/\gamma_{\alpha}^{2})\}\exp\left\{ -\frac{w_{\min}^{2}p_{\min}^{2}NTL\max(r_{1},r_{2},r_{3})\Delta_{\min}^{2}}{w_{\max}^{2}(N\vee T)r_{1}^{2}r_{2}^{2}r_{3}\mu_{0}^{2}\{\|\Theta\|_{\max}^{2}\vee(L_{\alpha}^{2}/\gamma_{\alpha}^{2})\}}\right\} ,
	\]
	for some constant $c$.
	
	\subsubsection{Step 4}
	
	This step aims to derive the upper bound for $\widehat{g}_{2}=\min_{\pi\in\Pi_{r_{3}}}\sum_{l=1}^{L}\sum_{b=1}^{r_{3}}\mathbf{1}(\mathcal{E}_{2})\|\boldsymbol{S}_{a,:}-\boldsymbol{S}_{b,:}\|_{2}^{2}/L$,
	i.e., the deterministic bounds for the error terms $\widehat{F}_{l},\widehat{G}_{l}$
	and $\widehat{H}_{l}$: 
	\begin{align*}
		\widehat{g}_{2} & =\min_{\pi\in\Pi_{r_{3}}}\frac{1}{L}\sum_{l=1}^{L}\sum_{b=1}^{r_{3}}\|\boldsymbol{S}_{a,:}-\boldsymbol{S}_{b,:}\|_{2}^{2}\cdot\mathbf{1}\left\{ \widehat{z}_{l}=b,\frac{1}{2}\|\boldsymbol{S}_{a,:}-\boldsymbol{S}_{b,:}\|_{2}^{2}\leq\widehat{F}_{l}+\widehat{G}_{l}+\widehat{H}_{l}\right\} .
	\end{align*}
	Hereafter, we provide the deterministic upper bounds for $\widehat{F}_{l}$,
	$\widehat{G}_{l}$ and $\widehat{H}_{l}$, respectively.
	
	(a) Upper bound for $F_{l}(a,b;\widehat{z})$. 
	\begin{align}
		F_{l}(a,b;\widehat{z})^{2} & \leq8|\langle\mathcal{M}_{(3)}(\widehat{\Theta}-\Theta)_{l,:}\widehat{\boldsymbol{V}},(\tilde{\boldsymbol{S}}_{a,:}-\widehat{\boldsymbol{S}}_{a,:})-(\tilde{\boldsymbol{S}}_{b,:}-\widehat{\boldsymbol{S}}_{b,:})\rangle|^{2}\nonumber \\
		& +8|\langle\mathcal{M}_{(3)}(\widehat{\Theta}-\Theta)_{l,:}(\boldsymbol{V}-\widehat{\boldsymbol{V}}),\tilde{\boldsymbol{S}}_{a,:}-\tilde{\boldsymbol{S}}_{b,:}\rangle|^{2}\nonumber \\
		& \leq32\|\mathcal{M}_{(3)}(\widehat{\Theta}-\Theta)_{l,:}\widehat{\boldsymbol{V}}\|_{\max}^{2}\cdot\max_{b\in[K]}\|\tilde{\boldsymbol{S}}_{b,:}-\widehat{\boldsymbol{S}}_{b,:}\|_{F}^{2}\nonumber \\
		& +8\|\mathcal{M}_{(3)}(\widehat{\Theta}-\Theta)_{l,:}(\boldsymbol{V}-\widehat{\boldsymbol{V}})\|_{\max}^{2}\cdot\|\tilde{\boldsymbol{S}}_{a,:}-\tilde{\boldsymbol{S}}_{b,:}\|^{2}\nonumber \\
		& \leq64\left\{ \|\mathcal{M}_{(3)}(\widehat{\Theta}-\Theta)_{l,:}\boldsymbol{V}\|_{\max}^{2}+\|\mathcal{M}_{(3)}(\widehat{\Theta}-\Theta)_{l,:}(\boldsymbol{V}-\widehat{\boldsymbol{V}})\|_{\max}^{2}\right\} \cdot\max_{b\in[K]}\|\tilde{\boldsymbol{S}}_{b,:}-\widehat{\boldsymbol{S}}_{b,:}\|_{2}^{2}\nonumber \\
		& +8\|\mathcal{M}_{(3)}(\widehat{\Theta}-\Theta)_{l,:}(\boldsymbol{V}-\widehat{\boldsymbol{V}})\|_{\max}^{2}\cdot\|\tilde{\boldsymbol{S}}_{a,:}-\tilde{\boldsymbol{S}}_{b,:}\|^{2},\label{eq:F_1}
	\end{align}
	where 
	\begin{align}
		\|\tilde{\boldsymbol{S}}_{b,:}-\widehat{\boldsymbol{S}}_{b,:}\|_{F}^{2} & =\|(\boldsymbol{W}_{:,b}-\widehat{\boldsymbol{W}}_{:,b})^{\intercal}\mathcal{M}_{(3)}(\widehat{\Theta})\boldsymbol{V}+\widehat{\boldsymbol{W}}_{:,b}^{\intercal}\mathcal{M}_{(3)}(\widehat{\Theta})(\boldsymbol{V}-\widehat{\boldsymbol{V}})\|^{2}\nonumber \\
		& \leq2\|(\boldsymbol{W}_{:,b}-\widehat{\boldsymbol{W}}_{:,b})^{\intercal}\mathcal{M}_{(3)}(\widehat{\Theta})\boldsymbol{V}\|^{2}+2\|\widehat{\boldsymbol{W}}_{:,b}^{\intercal}\mathcal{M}_{(3)}(\widehat{\Theta}-\Theta)(\boldsymbol{V}-\widehat{\boldsymbol{V}})\|^{2}\nonumber \\
		& \overset{(\ref{eq:error_W_V}),(\ref{eq:W_error_V})}{\lesssim}\left\{ \frac{r_{3}g(\widehat{z},z)}{\Delta_{\min}}\right\} ^{2}+\left\{ \frac{\mu_{0}r_{1}^{1/2}r_{2}^{1/2}r_{3}^{3/2}g(\widehat{z},z)\|\widehat{\Theta}-\Theta\|_{F}}{\Delta_{\min}^{2}\sqrt{NTL}}\right\} ^{2}+\frac{\mu_{0}^{2}r_{1}r_{2}r_{3}\|\widehat{\Theta}-\Theta\|_{F}^{2}}{NTL}\nonumber \\
		& \lesssim r_{3}g(\widehat{z},z)+\frac{\mu_{0}^{2}r_{1}r_{2}r_{3}\|\widehat{\Theta}-\Theta\|_{F}^{2}}{NTL},\label{eq:F_2}
	\end{align}
	and
	\begin{align}
		\|\tilde{\boldsymbol{S}}_{a,:}-\tilde{\boldsymbol{S}}_{b,:}\|^{2} & =\|\tilde{\boldsymbol{S}}_{a,:}-\boldsymbol{S}_{a,:}+\boldsymbol{S}_{a,:}-\boldsymbol{S}_{b,:}+\boldsymbol{S}_{b,:}-\tilde{\boldsymbol{S}}_{b,:}\|^{2}\nonumber \\
		& \leq3\|\boldsymbol{S}_{a,:}-\boldsymbol{S}_{b,:}\|^{2}+6\max_{a\in[r_{3}]}\|\tilde{\boldsymbol{S}}_{a,:}-\boldsymbol{S}_{a,:}\|^{2}\nonumber \\
		& =3\|\boldsymbol{S}_{a,:}-\boldsymbol{S}_{b,:}\|^{2}+6\max_{a\in[r_{3}]}\|\boldsymbol{W}_{:,a}^{\intercal}\mathcal{M}_{(3)}(\widehat{\Theta}-\Theta)\boldsymbol{V}\|^{2}\nonumber \\
		& \lesssim\|\boldsymbol{S}_{a,:}-\boldsymbol{S}_{b,:}\|^{2}+\|\boldsymbol{W}_{:,a}\|^{2}\cdot\|\mathcal{M}_{(3)}(\widehat{\Theta}-\Theta)\boldsymbol{V}\|_{\max}^{2}\nonumber \\
		& \overset{(\ref{eq:eigen_W})}{\lesssim}\|\boldsymbol{S}_{a,:}-\boldsymbol{S}_{b,:}\|^{2}+r_{3}/L\cdot\|\mathcal{M}_{(3)}(\widehat{\Theta}-\Theta)\|_{F}^{2}\|\boldsymbol{V}\|_{2,\max}^{2}\nonumber \\
		& \overset{(\ref{eq:eigen_W})}{\lesssim}\|\boldsymbol{S}_{a,:}-\boldsymbol{S}_{b,:}\|^{2}+\frac{\mu_{0}^{2}r_{1}r_{2}r_{3}\|\widehat{\Theta}-\Theta\|_{F}^{2}}{NTL}.\label{eq:F_3}
	\end{align}
	Combing (\ref{eq:F_1}), (\ref{eq:F_2}) and (\ref{eq:F_3}), we obtain
	\begin{align*}
		\frac{F_{l}(a,b;\widehat{z})^{2}}{\|\boldsymbol{S}_{a,:}-\boldsymbol{S}_{b,:}\|^{2}} & \lesssim\|\mathcal{M}_{(3)}(\widehat{\Theta}-\Theta)_{l,:}\boldsymbol{V}\|_{\max}^{2}\cdot\left\{ \frac{r_{3}g(\widehat{z},z)+\frac{\mu_{0}^{2}r_{1}r_{2}r_{3}\|\widehat{\Theta}-\Theta\|_{F}^{2}}{NTL}}{\Delta_{\min}^{2}}\right\} \\
		& +\|\mathcal{M}_{(3)}(\widehat{\Theta}-\Theta)_{l,:}(\boldsymbol{V}-\widehat{\boldsymbol{V}})\|_{\max}^{2}\cdot\left(1+\frac{r_{3}g(\widehat{z},z)+\frac{\mu_{0}^{2}r_{1}r_{2}r_{3}\|\widehat{\Theta}-\Theta\|_{F}^{2}}{NTL}}{\Delta_{\min}^{2}}\right),
	\end{align*}
	which is bounded by
	\begin{align}
		& \frac{F_{l}(a,b;\widehat{z})^{2}}{\|\boldsymbol{S}_{a,:}-\boldsymbol{S}_{b,:}\|_{2}^{2}}\nonumber \\
		\lesssim & \frac{1}{L}\sum_{l=1}^{L}\|\mathcal{M}_{(3)}(\widehat{\Theta}-\Theta)_{l,:}\boldsymbol{V}\|_{\max}^{2}\cdot\left\{ \frac{r_{3}g(\widehat{z},z)+\frac{\mu_{0}^{2}r_{1}r_{2}r_{3}\|\widehat{\Theta}-\Theta\|_{F}^{2}}{NTL}}{\Delta_{\min}^{2}}\right\} \nonumber \\
		& +\frac{1}{L}\sum_{l=1}^{L}\|\mathcal{M}_{(3)}(\widehat{\Theta}-\Theta)_{l,:}(\boldsymbol{V}-\widehat{\boldsymbol{V}})\|_{\max}^{2}\cdot\left(1+\frac{r_{3}g(\widehat{z},z)+\frac{\mu_{0}^{2}r_{1}r_{2}r_{3}\|\widehat{\Theta}-\Theta\|_{F}^{2}}{NTL}}{\Delta_{\min}^{2}}\right)\nonumber \\
		\lesssim & \frac{\|\widehat{\Theta}-\Theta\|_{F}^{2}}{L}\cdot\|\boldsymbol{V}\|_{2,\max}^{2}\cdot\frac{r_{3}g(\widehat{z},z)}{\Delta_{\min}^{2}}+\frac{\|\widehat{\Theta}-\Theta\|_{F}^{2}}{L}\cdot\|\boldsymbol{V}-\widehat{\boldsymbol{V}}\|_{2,\max}^{2}\cdot\frac{r_{3}g(\widehat{z},z)}{\Delta_{\min}^{2}}\nonumber \\
		\lesssim & \frac{\mu_{0}^{2}r_{1}r_{2}r_{3}g(\widehat{z},z)\|\widehat{\Theta}-\Theta\|_{F}^{2}}{\Delta_{\min}^{2}NTL}.\label{eq:F_part}
	\end{align}
	(b) Upper bound for $G_{l}(a,b;\widehat{z})$. 
	
	\begin{align}
		G_{l}(a,b;\widehat{z}) & =\left(\|\Theta_{l:}\widehat{\boldsymbol{V}}-\widehat{\boldsymbol{S}}_{a,:}\|_{F}^{2}-\|\Theta_{l:}\widehat{\boldsymbol{V}}-\boldsymbol{W}_{:,a}^{\intercal}\widehat{\Theta}\widehat{\boldsymbol{V}}\|_{F}^{2}\right)\nonumber \\
		& -\left(\|\Theta_{l:}\widehat{\boldsymbol{V}}-\widehat{\boldsymbol{S}}_{b,:}\|_{F}^{2}-\|\Theta_{l:}\widehat{\boldsymbol{V}}-\boldsymbol{W}_{:,b}^{\intercal}\widehat{\Theta}\widehat{\boldsymbol{V}}\|_{F}^{2}\right)\nonumber \\
		& =\left(\|\Theta_{l:}\widehat{\boldsymbol{V}}-\boldsymbol{W}_{:,a}^{\intercal}\widehat{\Theta}\widehat{\boldsymbol{V}}+\boldsymbol{W}_{:,a}^{\intercal}\widehat{\Theta}\widehat{\boldsymbol{V}}-\widehat{\boldsymbol{S}}_{a,:}\|_{F}^{2}-\|\Theta_{l:}\widehat{\boldsymbol{V}}-\boldsymbol{W}_{:,a}^{\intercal}\widehat{\Theta}\widehat{\boldsymbol{V}}\|_{F}^{2}\right)\nonumber \\
		& -\left(\|\Theta_{l:}\widehat{\boldsymbol{V}}-\boldsymbol{W}_{:,b}^{\intercal}\widehat{\Theta}\widehat{\boldsymbol{V}}+\boldsymbol{W}_{:,b}^{\intercal}\widehat{\Theta}\widehat{\boldsymbol{V}}-\widehat{\boldsymbol{S}}_{b,:}\|_{F}^{2}-\|\Theta_{l:}\widehat{\boldsymbol{V}}-\boldsymbol{W}_{:,b}^{\intercal}\widehat{\Theta}\widehat{\boldsymbol{V}}\|_{F}^{2}\right)\nonumber \\
		& =\|\boldsymbol{W}_{:,a}^{\intercal}\widehat{\Theta}\widehat{\boldsymbol{V}}-\widehat{\boldsymbol{S}}_{a,:}\|_{F}^{2}-\|\boldsymbol{W}_{:,b}^{\intercal}\widehat{\Theta}\widehat{\boldsymbol{V}}-\widehat{\boldsymbol{S}}_{b,:}\|_{F}^{2}\label{eq:G_part1}\\
		& +2\langle\Theta_{l:}\widehat{\boldsymbol{V}}-\boldsymbol{W}_{:,a}^{\intercal}\widehat{\Theta}\widehat{\boldsymbol{V}},\boldsymbol{W}_{:,a}^{\intercal}\widehat{\Theta}\widehat{\boldsymbol{V}}-\widehat{\boldsymbol{S}}_{a,:}\rangle\label{eq:G_part2}\\
		& -2\langle\Theta_{l:}\widehat{\boldsymbol{V}}-\boldsymbol{W}_{:,b}^{\intercal}\widehat{\Theta}\widehat{\boldsymbol{V}},\boldsymbol{W}_{:,b}^{\intercal}\widehat{\Theta}\widehat{\boldsymbol{V}}-\widehat{\boldsymbol{S}}_{b,:}\rangle,\label{eq:G_part3}
	\end{align}
	where $\Theta_{l:}=\boldsymbol{W}_{:,(z)_{l}}^{\intercal}\Theta=\boldsymbol{W}_{:,a}^{\intercal}\{\widehat{\Theta}-(\widehat{\Theta}-\Theta)\}$.
	For the second part (\ref{eq:G_part2}), we have
	\begin{align*}
		& \langle\Theta_{l:}\widehat{\boldsymbol{V}}-\boldsymbol{W}_{:,a}^{\intercal}\widehat{\Theta}\widehat{\boldsymbol{V}},\boldsymbol{W}_{:,a}^{\intercal}\widehat{\Theta}\widehat{\boldsymbol{V}}-\widehat{\boldsymbol{S}}_{a,:}\rangle\\
		= & \langle\boldsymbol{W}_{:,a}^{\intercal}\{\widehat{\Theta}-(\widehat{\Theta}-\Theta)\}\widehat{\boldsymbol{V}}-\boldsymbol{W}_{:,a}^{\intercal}\widehat{\Theta}\widehat{\boldsymbol{V}},\boldsymbol{W}_{:,a}^{\intercal}\widehat{\Theta}\widehat{\boldsymbol{V}}-\widehat{\boldsymbol{S}}_{a,:}\rangle\\
		= & -\langle\boldsymbol{W}_{:,a}^{\intercal}(\widehat{\Theta}-\Theta)\widehat{\boldsymbol{V}},\boldsymbol{W}_{:,a}^{\intercal}\widehat{\Theta}\widehat{\boldsymbol{V}}\rangle+\langle\boldsymbol{W}_{:,a}^{\intercal}(\widehat{\Theta}-\Theta)\widehat{\boldsymbol{V}},\widehat{\boldsymbol{S}}_{a,:}\rangle\\
		= & -\langle\boldsymbol{W}_{:,a}^{\intercal}(\widehat{\Theta}-\Theta)\widehat{\boldsymbol{V}},(\boldsymbol{W}_{:,a}-\widehat{\boldsymbol{W}}_{:,a})^{\intercal}\widehat{\Theta}\widehat{\boldsymbol{V}}\rangle\rangle\\
		& +\langle\boldsymbol{W}_{:,a}^{\intercal}(\widehat{\Theta}-\Theta)\widehat{\boldsymbol{V}},\widehat{\boldsymbol{S}}_{a,:}-\widehat{\boldsymbol{W}}_{:,a}^{\intercal}\widehat{\Theta}\widehat{\boldsymbol{V}}\rangle\\
		= & -\langle\boldsymbol{W}_{:,a}^{\intercal}(\widehat{\Theta}-\Theta)\widehat{\boldsymbol{V}},(\boldsymbol{W}_{:,a}-\widehat{\boldsymbol{W}}_{:,a})^{\intercal}\widehat{\Theta}\widehat{\boldsymbol{V}}\rangle\rangle,
	\end{align*}
	where $\widehat{\boldsymbol{S}}_{a,:}=\widehat{\boldsymbol{W}}_{:,a}^{\intercal}\widehat{\Theta}\widehat{\boldsymbol{V}}$.
	For the third part (\ref{eq:G_part3}), we have
	\begin{align*}
		& \langle\Theta_{l:}\widehat{\boldsymbol{V}}-\boldsymbol{W}_{:,b}^{\intercal}\widehat{\Theta}\widehat{\boldsymbol{V}},\boldsymbol{W}_{:,b}^{\intercal}\widehat{\Theta}\widehat{\boldsymbol{V}}-\widehat{\boldsymbol{S}}_{b,:}\rangle\\
		= & \langle\boldsymbol{W}_{:,a}^{\intercal}\{\widehat{\Theta}-(\widehat{\Theta}-\Theta)\}\widehat{\boldsymbol{V}}-\boldsymbol{W}_{:,b}^{\intercal}\widehat{\Theta}\widehat{\boldsymbol{V}},\boldsymbol{W}_{:,b}^{\intercal}\widehat{\Theta}\widehat{\boldsymbol{V}}-\widehat{\boldsymbol{S}}_{b,:}\rangle\\
		= & \langle\boldsymbol{W}_{:,a}^{\intercal}\Theta\widehat{\boldsymbol{V}}-\boldsymbol{W}_{:,b}^{\intercal}\Theta\widehat{\boldsymbol{V}}+\boldsymbol{W}_{:,b}^{\intercal}\Theta\widehat{\boldsymbol{V}}-\boldsymbol{W}_{:,b}^{\intercal}\widehat{\Theta}\widehat{\boldsymbol{V}},(\boldsymbol{W}_{:,b}-\widehat{\boldsymbol{W}}_{:,b})^{\intercal}\widehat{\Theta}\widehat{\boldsymbol{V}}\rangle\\
		= & \langle(\boldsymbol{W}_{:,a}-\boldsymbol{W}_{:,b})^{\intercal}\Theta\widehat{\boldsymbol{V}}+\boldsymbol{W}_{:,b}^{\intercal}(\Theta-\widehat{\Theta})\widehat{\boldsymbol{V}},(\boldsymbol{W}_{:,b}-\widehat{\boldsymbol{W}}_{:,b})^{\intercal}\widehat{\Theta}\widehat{\boldsymbol{V}}\rangle\\
		= & -\langle\boldsymbol{W}_{:,b}^{\intercal}(\widehat{\Theta}-\Theta)\widehat{\boldsymbol{V}},(\boldsymbol{W}_{:,b}-\widehat{\boldsymbol{W}}_{:,b})^{\intercal}\widehat{\Theta}\widehat{\boldsymbol{V}}\rangle\\
		& +\langle(\boldsymbol{W}_{:,a}-\boldsymbol{W}_{:,b})^{\intercal}\Theta\widehat{\boldsymbol{V}},(\boldsymbol{W}_{:,b}-\widehat{\boldsymbol{W}}_{:,b})^{\intercal}\widehat{\Theta}\widehat{\boldsymbol{V}}\rangle.
	\end{align*}
	Combined with (\ref{eq:G_part1}), we further have 
	\begin{align*}
		|G_{l}(a,b;\widehat{z})| & \leq\left|\|\boldsymbol{W}_{:,a}^{\intercal}\widehat{\Theta}\widehat{\boldsymbol{V}}-\widehat{\boldsymbol{S}}_{a,:}\|_{F}^{2}-\|\boldsymbol{W}_{:,b}^{\intercal}\widehat{\Theta}\widehat{\boldsymbol{V}}-\widehat{\boldsymbol{S}}_{b,:}\|_{F}^{2}\right|\\
		& +4\max_{a\in[r_{3}]}\langle\boldsymbol{W}_{:,a}^{\intercal}(\widehat{\Theta}-\Theta)\widehat{\boldsymbol{V}},(\boldsymbol{W}_{:,a}-\widehat{\boldsymbol{W}}_{:,a})^{\intercal}\widehat{\Theta}\widehat{\boldsymbol{V}}\rangle\\
		& +2\left|\langle(\boldsymbol{W}_{:,a}-\boldsymbol{W}_{:,b})^{\intercal}\widehat{\Theta}\widehat{\boldsymbol{V}},(\boldsymbol{W}_{:,b}-\widehat{\boldsymbol{W}}_{:,b})^{\intercal}\widehat{\Theta}\widehat{\boldsymbol{V}}\rangle\right|.
	\end{align*}
	Then, we analyze these three terms separately. First, the first term
	is
	\begin{align}
		& \left|\|\boldsymbol{W}_{:,a}^{\intercal}\widehat{\Theta}\widehat{\boldsymbol{V}}-\widehat{\boldsymbol{S}}_{a,:}\|_{F}^{2}-\|\boldsymbol{W}_{:,b}^{\intercal}\widehat{\Theta}\widehat{\boldsymbol{V}}-\widehat{\boldsymbol{S}}_{b,:}\|_{F}^{2}\right|^{2}\nonumber \\
		\leq & \max_{a\in[r_{3}]}\|\boldsymbol{W}_{:,a}^{\intercal}\widehat{\Theta}\widehat{\boldsymbol{V}}-\widehat{\boldsymbol{S}}_{a,:}\|_{F}^{4}\nonumber \\
		= & \max_{a\in[r_{3}]}\|(\boldsymbol{W}_{:,a}-\widehat{\boldsymbol{W}}_{:,a})^{\intercal}\widehat{\Theta}\widehat{\boldsymbol{V}}\|_{F}^{4}\nonumber \\
		\overset{(\ref{eq:error_W_V})}{\lesssim} & \frac{r_{3}^{4}g^{4}(\widehat{z},z)}{\Delta_{\min}^{4}}+\frac{\mu_{0}^{4}r_{1}^{2}r_{2}^{2}r_{3}^{6}g^{4}(\widehat{z},z)\|\widehat{\Theta}-\Theta\|_{F}^{4}}{\Delta_{\min}^{8}N^{2}T^{2}L^{2}}+\frac{\kappa^{8}r_{3}^{6}g^{4}(\widehat{z},z)\|\widehat{\Theta}-\Theta\|_{F}^{4}}{\Delta_{\min}^{8}L^{2}}\nonumber \\
		\overset{(\ref{eq:loss_condition})}{\lesssim} & \Delta_{\min}^{4}+\frac{\mu_{0}^{4}r_{1}^{2}r_{2}^{2}r_{3}^{4}g^{2}(\widehat{z},z)\|\widehat{\Theta}-\Theta\|_{F}^{4}}{\Delta_{\min}^{4}N^{2}T^{2}L^{2}}+\frac{\kappa^{8}r_{3}^{4}g^{2}(\widehat{z},z)\|\widehat{\Theta}-\Theta\|_{F}^{4}}{\Delta_{\min}^{4}L^{2}}.\label{eq:G_part1_bound}
	\end{align}
	The second term is:
	\begin{align}
		& \max_{a}\left|\langle\boldsymbol{W}_{:,a}^{\intercal}\mathcal{M}_{(3)}(\widehat{\Theta}-\Theta)\widehat{\boldsymbol{V}},(\boldsymbol{W}_{:,a}-\widehat{\boldsymbol{W}}_{:,a})^{\intercal}\widehat{\Theta}\widehat{\boldsymbol{V}}\rangle\right|^{2}\nonumber \\
		\leq & \max_{a}\|\boldsymbol{W}_{:,a}^{\intercal}\mathcal{M}_{(3)}(\widehat{\Theta}-\Theta)\widehat{\boldsymbol{V}}\|^{2}\cdot\max_{a}\|(\boldsymbol{W}_{:,a}-\widehat{\boldsymbol{W}}_{:,a})^{\intercal}\widehat{\Theta}\widehat{\boldsymbol{V}}\|^{2}\nonumber \\
		\leq & \max_{a}\|\boldsymbol{W}_{:,a}\|^{2}\cdot\|\mathcal{M}_{(3)}(\widehat{\Theta}-\Theta)\|_{F}^{2}\cdot\|\boldsymbol{V}\|_{2,\max}^{2}\nonumber \\
		& \times\left\{ \frac{r_{3}^{2}g^{2}(\widehat{z},z)}{\Delta_{\min}^{2}}+\frac{\mu_{0}^{2}r_{1}r_{2}r_{3}^{3}g^{2}(\widehat{z},z)\|\widehat{\Theta}-\Theta\|_{F}^{2}}{\Delta_{\min}^{4}NTL}+\frac{\kappa^{4}r_{3}^{3}g^{2}(\widehat{z},z)\|\widehat{\Theta}-\Theta\|_{F}^{2}}{\Delta_{\min}^{4}L}\right\} \nonumber \\
		& \lesssim\frac{r_{3}}{L}\cdot\|\widehat{\Theta}-\Theta\|_{F}^{2}\cdot\frac{\mu_{0}^{2}r_{1}r_{2}}{NT}\nonumber \\
		& \times\left\{ \Delta_{\min}^{2}+\frac{\mu_{0}^{2}r_{1}r_{2}r_{3}^{3}g^{2}(\widehat{z},z)\|\widehat{\Theta}-\Theta\|_{F}^{2}}{\Delta_{\min}^{4}NTL}+\frac{\kappa^{4}r_{3}^{3}g^{2}(\widehat{z},z)\|\widehat{\Theta}-\Theta\|_{F}^{2}}{\Delta_{\min}^{4}L}\right\} \nonumber \\
		& \lesssim\frac{\mu_{0}^{2}r_{1}r_{2}r_{3}\Delta_{\min}^{2}\|\widehat{\Theta}-\Theta\|_{F}^{2}}{NTL}.\label{eq:G_part2_bound}
	\end{align}
	The last term is: 
	\begin{align}
		& \left|\langle(\boldsymbol{W}_{:,a}-\boldsymbol{W}_{:,b})^{\intercal}\Theta\widehat{\boldsymbol{V}},(\boldsymbol{W}_{:,b}-\widehat{\boldsymbol{W}}_{:,b})^{\intercal}\widehat{\Theta}\widehat{\boldsymbol{V}}\rangle\right|^{2}\nonumber \\
		\leq & \|(\boldsymbol{W}_{:,a}-\boldsymbol{W}_{:,b})^{\intercal}\Theta\widehat{\boldsymbol{V}}\|^{2}\cdot\|(\boldsymbol{W}_{:,b}-\widehat{\boldsymbol{W}}_{:,b})^{\intercal}\widehat{\Theta}\widehat{\boldsymbol{V}}\|^{2}\nonumber \\
		\leq & \|(\boldsymbol{S}_{:,a}-\boldsymbol{S}_{:,b})\boldsymbol{V}^{\intercal}\widehat{\boldsymbol{V}}\|^{2}\cdot\|(\boldsymbol{W}_{:,b}-\widehat{\boldsymbol{W}}_{:,b})^{\intercal}\widehat{\Theta}\widehat{\boldsymbol{V}}\|^{2}\nonumber \\
		\lesssim & \|\boldsymbol{S}_{:,a}-\boldsymbol{S}_{:,b}\|^{2}\times\left\{ \Delta_{\min}^{2}+\frac{\mu_{0}^{2}r_{1}r_{2}r_{3}^{3}g^{2}(\widehat{z},z)\|\widehat{\Theta}-\Theta\|_{F}^{2}}{\Delta_{\min}^{4}NTL}\right\} ,\label{eq:G_part3_bound}
	\end{align}
	where $\boldsymbol{W}_{:,a}^{\intercal}\Theta=\boldsymbol{S}_{a,:}\boldsymbol{V}^{\intercal}$.
	Combining (\ref{eq:G_part1_bound}), (\ref{eq:G_part2_bound}) and
	(\ref{eq:G_part3_bound}), we obtain 
	\begin{align}
		\frac{G_{l}(a,b;\widehat{z})^{2}}{\|\boldsymbol{S}_{a,:}-\boldsymbol{S}_{b,:}\|^{2}} & \lesssim\Delta_{\min}^{2}+\frac{\mu_{0}^{2}r_{1}r_{2}r_{3}\|\widehat{\Theta}-\Theta\|_{F}^{2}}{NTL}+\frac{\mu_{0}^{2}r_{1}r_{2}r_{3}\|\widehat{\Theta}-\Theta\|_{F}^{2}}{\Delta_{\min}^{2}NTL}\cdot\frac{\mu_{0}^{2}r_{1}r_{2}r_{3}^{3}g^{2}(\widehat{z},z)\|\widehat{\Theta}-\Theta\|_{F}^{2}}{\Delta_{\min}^{4}NTL}\nonumber \\
		& \lesssim\Delta_{\min}^{2}+\frac{\mu_{0}^{2}r_{1}r_{2}r_{3}\|\widehat{\Theta}-\Theta\|_{F}^{2}}{NTL}\lesssim\Delta_{\min}^{2},\label{eq:G_part}
	\end{align}
	where $\|\widehat{\Theta}-\Theta\|_{F}^{2}\lesssim NTL\Delta_{\min}^{2}/(\mu_{0}^{2}r_{1}r_{2}r_{3})$
	holds with high probability by Theorem \ref{thm:estimation}.
	
	(c) Upper bound for $H_{l}(a,b)$. 
	\begin{align*}
		\widehat{H}_{l} & =\|\Theta_{l:}\widehat{\boldsymbol{V}}-\boldsymbol{W}_{:,a}^{\intercal}\widehat{\Theta}\widehat{\boldsymbol{V}}\|_{F}^{2}-\|\Theta_{l:}\widehat{\boldsymbol{V}}-\boldsymbol{W}_{:,b}^{\intercal}\widehat{\Theta}\widehat{\boldsymbol{V}}\|_{F}^{2}+\|\boldsymbol{S}_{a,:}-\boldsymbol{S}_{b,:}\|^{2}\\
		& =\|\boldsymbol{W}_{:,a}^{\intercal}(\widehat{\Theta}-\Theta)\widehat{\boldsymbol{V}}\|_{F}^{2}+\left(\|\boldsymbol{S}_{a,:}-\boldsymbol{S}_{b,:}\|^{2}-\|\Theta_{l:}\widehat{\boldsymbol{V}}-\boldsymbol{W}_{:,b}^{\intercal}\Theta\widehat{\boldsymbol{V}}\|_{F}^{2}\right)\\
		& -\left(\|\Theta_{l:}\widehat{\boldsymbol{V}}-\boldsymbol{W}_{:,b}^{\intercal}\widehat{\Theta}\widehat{\boldsymbol{V}}\|_{F}^{2}-\|\Theta_{l:}\widehat{\boldsymbol{V}}-\boldsymbol{W}_{:,b}^{\intercal}\Theta\widehat{\boldsymbol{V}}\|_{F}^{2}\right)\\
		& =\|\boldsymbol{W}_{:,a}^{\intercal}(\widehat{\Theta}-\Theta)\widehat{\boldsymbol{V}}\|_{F}^{2}+\left(\|\boldsymbol{S}_{a,:}-\boldsymbol{S}_{b,:}\|^{2}-\|\Theta_{l:}\widehat{\boldsymbol{V}}-\boldsymbol{W}_{:,b}^{\intercal}\Theta\widehat{\boldsymbol{V}}\|_{F}^{2}\right)\\
		& -\left(\|\Theta_{l:}\widehat{\boldsymbol{V}}-\boldsymbol{W}_{:,b}^{\intercal}\widehat{\Theta}\widehat{\boldsymbol{V}}\|_{F}^{2}-\|\Theta_{l:}\widehat{\boldsymbol{V}}-\boldsymbol{W}_{:,b}^{\intercal}\Theta\widehat{\boldsymbol{V}}\|_{F}^{2}\right)\\
		& =\left(\|\boldsymbol{S}_{a,:}-\boldsymbol{S}_{b,:}\|^{2}-\|\Theta_{l:}\widehat{\boldsymbol{V}}-\boldsymbol{W}_{:,b}^{\intercal}\Theta\widehat{\boldsymbol{V}}\|_{F}^{2}\right)\\
		& +\|\boldsymbol{W}_{:,a}^{\intercal}(\widehat{\Theta}-\Theta)\widehat{\boldsymbol{V}}\|_{F}^{2}-\|\boldsymbol{W}_{:,b}^{\intercal}(\widehat{\Theta}-\Theta)\widehat{\boldsymbol{V}}\|_{F}^{2}\\
		& +2\langle(\boldsymbol{S}_{a,:}-\boldsymbol{S}_{b,:})\boldsymbol{V}^{\intercal}\widehat{\boldsymbol{V}}-\boldsymbol{W}_{:,b}^{\intercal}(\widehat{\Theta}-\Theta)\widehat{\boldsymbol{V}}\rangle,
	\end{align*}
	where $\Theta_{l:}=\boldsymbol{W}_{:,(z)_{l}}^{\intercal}\Theta=\boldsymbol{W}_{:,a}^{\intercal}\{\widehat{\Theta}-(\widehat{\Theta}-\Theta)\}$
	and $\boldsymbol{W}_{:,a}^{\intercal}\Theta=\boldsymbol{S}_{a,:}\boldsymbol{V}^{\intercal}$.
	For the first term, we have
	\begin{align*}
		& \left|\|\boldsymbol{S}_{a,:}-\boldsymbol{S}_{b,:}\|^{2}-\|\Theta_{l:}\widehat{\boldsymbol{V}}-\boldsymbol{W}_{:,b}^{\intercal}\Theta\widehat{\boldsymbol{V}}\|_{F}^{2}\right|\\
		= & \left|\|\boldsymbol{S}_{a,:}-\boldsymbol{S}_{b,:}\|^{2}-\|(\boldsymbol{S}_{a,:}-\boldsymbol{S}_{b,:})\boldsymbol{V}^{\intercal}\widehat{\boldsymbol{V}}\|_{F}^{2}\right|\\
		\lesssim & \|\boldsymbol{S}_{a,:}-\boldsymbol{S}_{b,:}\|^{2}.
	\end{align*}
	And the second term is bounded by:
	\begin{align*}
		& \left|\|\boldsymbol{W}_{:,a}^{\intercal}(\widehat{\Theta}-\Theta)\widehat{\boldsymbol{V}}\|_{F}^{2}-\|\boldsymbol{W}_{:,b}^{\intercal}(\widehat{\Theta}-\Theta)\widehat{\boldsymbol{V}}\|_{F}^{2}\right|\\
		\leq & \max_{a}\|\boldsymbol{W}_{:,a}^{\intercal}(\widehat{\Theta}-\Theta)\widehat{\boldsymbol{V}}\|_{F}^{2}\\
		\leq & \max_{a}\|\boldsymbol{W}_{:,a}\|^{2}\cdot\|\mathcal{M}_{(3)}(\widehat{\Theta}-\Theta)\|_{F}^{2}\cdot\|\boldsymbol{V}\|_{2,\max}^{2}\\
		\lesssim & \frac{\mu_{0}^{2}r_{1}r_{2}r_{3}\|\widehat{\Theta}-\Theta\|_{F}^{2}}{NTL}.
	\end{align*}
	The third term is
	\begin{align*}
		& \langle(\boldsymbol{S}_{a,:}-\boldsymbol{S}_{b,:})\boldsymbol{V}^{\intercal}\widehat{\boldsymbol{V}}-\boldsymbol{W}_{:,b}^{\intercal}(\widehat{\Theta}-\Theta)\widehat{\boldsymbol{V}}\rangle\\
		\leq & \|(\boldsymbol{S}_{a,:}-\boldsymbol{S}_{b,:})\boldsymbol{V}^{\intercal}\widehat{\boldsymbol{V}}\|\cdot\|\boldsymbol{W}_{:,b}^{\intercal}(\widehat{\Theta}-\Theta)\widehat{\boldsymbol{V}}\|\\
		\lesssim & \|\boldsymbol{S}_{a,:}-\boldsymbol{S}_{b,:}\|\cdot\frac{\mu_{0}^{2}r_{1}r_{2}r_{3}\|\widehat{\Theta}-\Theta\|_{F}^{2}}{NTL}.
	\end{align*}
	Thus, we have 
	\begin{equation}
		\frac{|H_{l}(a,b)|}{\|\boldsymbol{S}_{a,:}-\boldsymbol{S}_{b,:}\|^{2}}\leq c+\frac{\mu_{0}^{2}r_{1}r_{2}r_{3}\|\widehat{\Theta}-\Theta\|_{F}^{2}}{\Delta_{\min}^{2}NTL}\leq\frac{1}{4}.\label{eq:H_part}
	\end{equation}
	Combining (\ref{eq:F_part}), (\ref{eq:G_part}) and (\ref{eq:H_part}),
	we obtain
	\begin{align*}
		\widehat{g}_{2} & =\min_{\pi\in\Pi_{r_{3}}}\frac{1}{L}\sum_{l=1}^{L}\sum_{b=1}^{r_{3}}\|\boldsymbol{S}_{a,:}-\boldsymbol{S}_{b,:}\|_{2}^{2}\cdot\mathbf{1}\left\{ \widehat{z}_{l}=b,\frac{1}{2}\|\boldsymbol{S}_{a,:}-\boldsymbol{S}_{b,:}\|_{2}^{2}\leq\widehat{F}_{l}+\widehat{G}_{l}+\widehat{H}_{l}\right\} \\
		& \leq=\min_{\pi\in\Pi_{r_{3}}}\frac{1}{L}\sum_{l=1}^{L}\sum_{b=1}^{r_{3}}\|\boldsymbol{S}_{a,:}-\boldsymbol{S}_{b,:}\|_{2}^{2}\cdot\mathbf{1}\left\{ \widehat{z}_{l}=b,\frac{1}{4}\|\boldsymbol{S}_{a,:}-\boldsymbol{S}_{b,:}\|_{2}^{2}\leq\widehat{F}_{l}+\widehat{G}_{l}\right\} \\
		& \leq\min_{\pi\in\Pi_{r_{3}}}\frac{1}{L}\sum_{l=1}^{L}\sum_{b=1}^{r_{3}}\|\boldsymbol{S}_{a,:}-\boldsymbol{S}_{b,:}\|_{2}^{2}\cdot\mathbf{1}\left\{ \widehat{z}_{l}=b,\|\boldsymbol{S}_{a,:}-\boldsymbol{S}_{b,:}\|_{2}^{4}\leq64(\widehat{F}_{l}^{2}+\widehat{G}_{l}^{2})\right\} \\
		& \leq\min_{\pi\in\Pi_{r_{3}}}\frac{1}{L}\sum_{l=1}^{L}\sum_{b\in[r_{3}]/a}\mathbf{1}\left\{ \widehat{z}_{l}=b\right\} \cdot64\left(\frac{\widehat{F}_{l}^{2}}{\|\boldsymbol{S}_{a,:}-\boldsymbol{S}_{b,:}\|_{2}^{2}}+\frac{\widehat{G}_{l}^{2}}{\|\boldsymbol{S}_{a,:}-\boldsymbol{S}_{b,:}\|_{2}^{2}}\right)\\
		& \leq\min_{\pi\in\Pi_{r_{3}}}\frac{64}{L}\sum_{l=1}^{L}\max_{b\in[r_{3}]/a}\frac{\widehat{F}_{l}^{2}}{\|\boldsymbol{S}_{a,:}-\boldsymbol{S}_{b,:}\|_{2}^{2}}+\min_{\pi\in\Pi_{r_{3}}}\frac{64}{L}\sum_{l=1}^{L}\mathbf{1}\left\{ \widehat{z}_{l}\neq z_{l}\right\} \max_{b\in[r_{3}]/a}\frac{\widehat{G}_{l}^{2}}{\|\boldsymbol{S}_{a,:}-\boldsymbol{S}_{b,:}\|_{2}^{2}}\\
		& \leq\frac{g(\widehat{z},z)}{8}\frac{\mu_{0}^{2}r_{1}r_{2}r_{3}\|\widehat{\Theta}-\Theta\|_{F}^{2}}{\Delta_{\min}^{2}NTL}+\min_{\pi\in\Pi_{r_{3}}}\frac{1}{L}\sum_{l=1}^{L}\mathbf{1}\left\{ \widehat{z}_{l}\neq z_{l}\right\} \frac{\Delta_{\min}^{2}}{16}\\
		& \leq\frac{g(\widehat{z},z)}{8}\frac{\mu_{0}^{2}r_{1}r_{2}r_{3}\|\widehat{\Theta}-\Theta\|_{F}^{2}}{\Delta_{\min}^{2}NTL}+\frac{g(\widehat{z},z)}{16}\\
		& \leq\frac{g(\widehat{z},z)}{16}\left(1+\frac{2\mu_{0}^{2}r_{1}r_{2}r_{3}\|\widehat{\Theta}-\Theta\|_{F}^{2}}{\Delta_{\min}^{2}NTL}\right)\leq\frac{g(\widehat{z},z)}{16}\left(1+\frac{1}{32}\right)\leq\frac{g(\widehat{z},z)}{10}.
	\end{align*}

	\paragraph{Step 5}
	
	By combining Step 4 and Step3, we obtain 
	\begin{align*}
		g(\widehat{z},z) & \leq\widehat{g}_{1}+\widehat{g}_{2}\\
		& \leq\left\{ \|\Theta\|_{\max}^{2}\vee(L_{\alpha}^{2}/\gamma_{\alpha}^{2})\right\} \exp\left\{ -\frac{w_{\min}^{2}p_{\min}^{2}NTL\max(r_{1},r_{2},r_{3})\Delta_{\min}^{2}}{w_{\max}^{2}(N\vee T)r_{1}^{2}r_{2}^{2}r_{3}\mu_{0}^{2}\left\{ \|\Theta\|_{\max}^{2}\vee(L_{\alpha}^{2}/\gamma_{\alpha}^{2})\right\} }\right\} +\frac{g(\widehat{z},z)}{10},
	\end{align*}
	which implies that 
	\[
	g(\widehat{z},z)\lesssim\left\{ \|\Theta\|_{\max}^{2}\vee(L_{\alpha}^{2}/\gamma_{\alpha}^{2})\right\} \exp\left\{ -\frac{w_{\min}^{2}p_{\min}^{2}NTL\max(r_{1},r_{2},r_{3})\Delta_{\min}^{2}}{w_{\max}^{2}(N\vee T)r_{1}^{2}r_{2}^{2}r_{3}\mu_{0}^{2}\left\{ \|\Theta\|_{\max}^{2}\vee(L_{\alpha}^{2}/\gamma_{\alpha}^{2})\right\} }\right\} ,
	\]
	with probability at least $1-c_{0}(N+T+L)^{-2}.$
	
	\subsection{Proof of Lemmas}
	
	\subsubsection{Proof of Lemma \ref{lem:restricted_convex}}
	
	We aim to prove that $\sum_{i,t,l}\mathbf{1}(Y_{i,t-1,l}=1)(\widehat{\theta}_{i,t,l}-\theta_{i,t,l})^{2}$
	is larger than $\|\Delta_{\Theta}\|_{F}^{2}=\|\widehat{\Theta}-\Theta\|_{F}^{2}$
	up to an additive term. Denote 
	\begin{align}
		& \|\Delta_{\Theta}\|_{\E}^{2}=\E\left\{ \sum_{i,t,l}\mathbf{1}(Y_{i,t-1,l}=1)(\widehat{\theta}_{i,t,l}-\theta_{i,t,l})^{2}\right\} \geq p_{\min}\|\Delta_{\Theta}\|_{F}^{2},\label{eq:rsc_1}\\
		& \|\Delta_{\Theta}\|_{\max}^{2}=\max_{i,t,l}|(\Delta_{\Theta})_{i,t,l}|\leq2|\Theta_{\max}|,\nonumber 
	\end{align}
	where the expectation is taken with respect to $\mathbf{1}(Y_{i,t-1,l}=1)$.
	Given (\ref{eq:rsc_1}), we can show that 
	\begin{align}
		& P\left\{ \frac{p_{\min}}{2}\|\Delta_{\Theta}\|_{F}^{2}\geq\sum_{i,t,l}\mathbf{1}(Y_{i,t-1,l}=1)(\widehat{\theta}_{i,t,l}-\theta_{i,t,l})^{2}+2\|\Delta_{\Theta}\|_{\max}^{2}\vartheta\right\} \nonumber \\
		\leq & P\left\{ \frac{\|\Delta_{\Theta}\|_{\E}^{2}}{2}\geq\sum_{i,t,l}\mathbf{1}(Y_{i,t-1,l}=1)(\widehat{\theta}_{i,t,l}-\theta_{i,t,l})^{2}+2\|\Delta_{\Theta}\|_{\max}^{2}\vartheta\right\} .\label{eq:rsc_2}
	\end{align}
	Therefore, our goal reduces to prove (\ref{eq:rsc_2}) is negligible. 
	
	Our proof for the restricted strong convexity proceeds by using the
	standard peeling argument. Let $\xi$ be a constant larger than $1$
	(i.e., $\xi=2$) and define for every $\rho\geq0$, 
	\[
	\mathcal{B}_{\rho\xi^{l-1}}=\left\{ \Delta_{\Theta}\in\mathcal{B}:\rho\xi^{l-1}\leq\frac{\|\Delta_{\Theta}\|_{\E}^{2}}{\|\Theta\|_{\max}^{2}}\leq\rho\xi^{l}\right\} ,\quad l=1,2,\cdots,
	\]
	and therefore $\mathcal{B}=\cup_{l=1}^{\infty}\mathcal{B}_{\rho\xi^{l-1}}$.
	For some $l\geq1$ and $\Delta_{\Theta}\in\mathcal{B}_{\theta\xi^{l-1}}$,
	denote the event in (\ref{eq:rsc_2}) by 
	\[
	\mathcal{E}_{l}=\left\{ \exists\Delta_{\Theta}\in\mathcal{B}_{\rho\xi^{l-1}}:\left|\|\Delta_{\Theta}\|_{\E}^{2}-\sum_{i,t,l}\mathbf{1}(Y_{i,t-1,l}=1)(\widehat{\theta}_{i,t,l}-\theta_{i,t,l})^{2}\right|\geq\frac{1}{2}\|\Delta_{\Theta}\|_{\E}^{2}+2\vartheta\geq\frac{1}{2\xi}\|\Theta\|_{\max}^{2}\rho\xi^{l}+2\vartheta\right\} ,
	\]
	and $\mathcal{E}\subset\cup_{l=1}^{\infty}\mathcal{E}_{l}$. Let 
	\[
	\tilde{Z}_{\rho}=\sup_{\Delta_{\Theta}\in\mathcal{B}_{\rho\xi^{l-1}}}\left|\|\Delta_{\Theta}\|_{\E}^{2}-\sum_{i=1}^{N}\left\{ \sum_{t,l}\mathbf{1}(Y_{i,t-1,l}=1)(\widehat{\theta}_{i,t,l}-\theta_{i,t,l})^{2}\right\} \right|,
	\]
	where $\Delta_{\Theta}\in\mathcal{B}_{\rho\xi^{l-1}}$. Since each unit is at most being observed for $T$ times for only one treatment,
	we know 
	\begin{align*}
		\sigma_{\tilde{Z}_{\rho}}^{2} & =\sup_{\Delta_{\Theta}\in\mathcal{B}_{\rho\xi^{l-1}}}\sum_{i=1}^{N}\text{var}\left\{ \sum_{t,l}\mathbf{1}(Y_{i,t-1,l}=1)(\widehat{\theta}_{i,t,l}-\theta_{i,t,l})^{2}\right\} \\
		& \leq\sup_{\Delta_{\Theta}\in\mathcal{B}_{\rho\xi^{l-1}}}\sum_{i=1}^{N}\E\left\{ \sum_{t,l}\mathbf{1}(Y_{i,t-1,l}=1)(\widehat{\theta}_{i,t,l}-\theta_{i,t,l})^{2}\right\} ^{2}\\
		& \leq T\sup_{\Delta_{\Theta}\in\mathcal{B}_{\rho\xi^{l-1}}}\sum_{i=1}^{N}\E\left\{ \sum_{t,l}\mathbf{1}(Y_{i,t-1,l}=1)(\widehat{\theta}_{i,t,l}-\theta_{i,t,l})^{2}\right\} \\
		& \leq T\sup_{\Delta_{\Theta}\in\mathcal{B}_{\rho\xi^{l-1}}}\|\Delta_{\Theta}\|_{\E}^{2}\leq T\|\Theta\|_{\max}^{2}\rho\xi^{l},
	\end{align*}
	by the definition of $\mathcal{B}_{\rho\xi^{l-1}}$ which provides
	upper bounds for $\|\Delta_{\Theta}\|_{\E}^{2}$ and $\|\Delta_{\Theta}\|_{\max}^{2}$.
	By the symmetrization argument (Lemma 6.3, \citet{ledoux1991probability}),
	\begin{align*}
		\E(\tilde{Z}_{\rho}) & \leq2\E\left\{ \sup_{\Delta_{\Theta}\in\mathcal{B}_{\rho\xi^{l-1}}}\left|\sum_{i=1}^{N}\zeta_{i}\sum_{t,l}\mathbf{1}(Y_{i,t-1,l}=1)(\widehat{\theta}_{i,t,l}-\theta_{i,t,l})^{2}\right|\right\} \\
		& \leq4\sqrt{\frac{r_{1}r_{2}r_{3}\|\Theta\|_{\max}^{2}\rho\xi^{l}}{\max(r_{1},r_{2},r_{3})p_{\min}}}\E\left\{ \sup_{\Delta_{\Theta}\in\mathcal{B}_{\rho\xi^{l-1}}}\|\Delta_{\Theta}^{\text{Rad}}\|\right\} \\
		& \leq C_{1}\|\Theta\|_{\max}^{2}\rho\xi^{l}+\frac{r_{1}r_{2}r_{3}}{\max(r_{1},r_{2},r_{3})p_{\min}}\left[\E\left\{ \sup_{\Delta_{\Theta}\in\mathcal{B}_{\rho\xi^{l-1}}}\|\Delta_{\Theta}^{\text{Rad}}\|\right\} \right]^{2},
	\end{align*}
	where $\E\left\{ \sup_{\Delta_{\Theta}\in\mathcal{B}_{\rho\xi^{l-1}}}\|\Delta_{\Theta}^{\text{Rad}}\|\right\} $
	is the Rademacher complexity. Each entry of $\Delta_{\Theta}^{\text{Rad}}$
	is defined by
	\[
	(\Delta_{\Theta}^{\text{Rad}})_{i,t,l}=\zeta_{i}\mathbf{1}(Y_{i,t-1,l}=1)(\widehat{\theta}_{i,t,l}-\theta_{i,t,l})\cdot e_{i}(N)\otimes e_{t}(T)\otimes e_{l}(L),
	\]
	where $e_{i}(N)\otimes e_{t}(T)\otimes e_{l}(K)$ is a zero tensor
	except its $(i,t,l)$-entry and $\{\zeta_{i}\}_{i=1}^{N}$ are i.i.d
	Rademacher random variables. Our next step is to obtain a close form
	of $\E\left\{ \sup_{\Delta_{\Theta}\in\mathcal{B}_{\rho\xi^{l-1}}}\|\Delta_{\Theta}^{\text{Rad}}\|\right\} $
	by tensor inequality. By Lemma \ref{lem:tensor_bernstein}, we can
	show that
	\begin{align*}
		P\left(\sup_{\Delta_{\Theta}\in\mathcal{B}_{\rho\xi^{l-1}}}\|\Delta_{\Theta}^{\text{Rad}}\|\geq\alpha\right) & \leq(N+T+L)\exp\left[\frac{-\alpha^{2}}{2\sigma_{\Delta_{\Theta}}^{2}+\{T^{1/2}\|\Theta\|_{\max}\log(N+T+L)\alpha\}/3}\right]\\
		& \leq(N+T+L)\exp\left\{ -\frac{3}{4}\frac{\alpha^{2}}{(N\vee T)\|\Theta\|_{\max}^{2}\log(N+T+L)}\right\} ,
	\end{align*}
	where $\sigma_{\Delta_{\Theta}}^{2}\lesssim(N\vee T)\|\Theta\|_{\max}^{2}\log(N+T+L)$.
	By H\"older's inequality, we have
	\begin{align*}
		& \E\left\{ \sup_{\Delta_{\Theta}\in\mathcal{B}_{\rho\xi^{l-1}}}\|\Delta_{\Theta}^{\text{Rad}}\|\right\} \lesssim\sqrt{(N\vee T)}\|\Theta\|_{\max}\{\log(N+T+L)\}^{1/2}.
	\end{align*}
	Then, we invoke the Theorem 3 in \citet{massart2000constants} with
	$\varepsilon=1$: 
	\begin{align*}
		\pr(\mathcal{E}_{l})\leq & P\left\{ \tilde{Z}_{\rho}\geq\frac{1}{2\xi}\|\Theta\|_{\max}^{2}\rho\xi^{l}+\frac{2r_{1}r_{2}r_{3}(N\vee T)}{\max(r_{1},r_{2},r_{3})p_{\min}}\|\Theta\|_{\max}^{2}\log(N+T+L)\right\} \\
		\leq & P\left\{ \tilde{Z}_{\rho}\geq2\sigma_{\tilde{Z}_{\rho}}\sqrt{x}+34.5Tx+2\E(\tilde{Z}_{\rho})\right\} \\
		\leq & \exp\left(-C_{3}x\right),
	\end{align*}
	where $x=C_{3}\rho\xi^{l}/T$. Therefore, we have 
	\begin{align*}
		\pr(\mathcal{E}_{l}) & =\pr\left\{ \sup_{\Delta_{\Theta}\in\mathcal{B}_{\rho\xi^{l-1}}}\left|\|\Delta_{\Theta}\|_{\E}^{2}-\sum_{i,t,l}\mathbf{1}(Y_{i,t-1,l}=1)(\widehat{\theta}_{i,t,l}-\theta_{i,t,l})^{2}\right|\geq\frac{1}{2\xi}\rho\xi^{l}+2\vartheta\right\} \\
		& \leq\exp\left(-\frac{C_{3}\rho\xi^{l}}{T}\right)\leq\exp\left(-\frac{C_{3}\rho l\log(\xi)}{T}\right)\leq\exp\left(-\frac{C_{3}'\rho l}{T}\right),
	\end{align*}
	since $\xi^{l}\geq l\log(\xi)$ for $\xi=2$ and 
	\[
	\vartheta=\frac{r_{1}r_{2}r_{3}(N\vee T)}{\max(r_{1},r_{2},r_{3})p_{\min}}\|\Theta\|_{\max}^{2}\log(N+T+L).
	\]
	Thus, 
	\[
	\pr(\mathcal{E})\leq\sum_{l=1}^{\infty}\left\{ \exp\left(-\frac{C_{3}'\rho}{T}\right)\right\} ^{l}=\frac{\exp\left(-\frac{C_{3}'\rho}{T}\right)}{1-\exp\left(-\frac{C_{3}'\rho}{T}\right)}\leq2\exp\left(-\frac{C_{3}'\rho}{T}\right),
	\]
	and 
	\[
	\pr\left\{ \frac{1}{2}\|\Delta_{\Theta}\|_{\E}^{2}\geq\sum_{i,t,l}\mathbf{1}(Y_{i,t-1,l}=1)(\widehat{\theta}_{i,t,l}-\theta_{i,t,l})^{2}+2\vartheta\right\} \leq2\exp\left(-\frac{C_{3}'\rho}{T}\right)\lesssim\frac{1}{(N+T+L)^{2}},
	\]
	where $\rho=C'T\log(N+T+L)$ is determined to match the tail probability.
	To sum up, when $\|\widehat{\Theta}-\Theta\|_{F}^{2}\geq C_{0}\|\Theta\|_{\max}^{2}T\log(N+T+L)/p_{\min}$,
	we have 
	\begin{align*}
		& \pr\left\{ \frac{p_{\min}}{2}\sum_{i,t,l}(\widehat{\theta}_{i,t,l}-\theta_{i,t,l})^{2}\geq\sum_{i,t,l}\mathbf{1}(Y_{i,t-1,l}=1)(\widehat{\theta}_{i,t,l}-\theta_{i,t,l})^{2}+2\vartheta\right\} \\
		\leq & \pr\left\{ \frac{1}{2}\|\Delta_{\Theta}\|_{\E}^{2}\geq\sum_{i,t,l}\mathbf{1}(Y_{i,t-1,l}=1)(\widehat{\theta}_{i,t,l}-\theta_{i,t,l})^{2}+2\vartheta\right\} \lesssim\frac{1}{(N+T+L)^{2}},
	\end{align*}
	where 
	\[
	\vartheta=\frac{C''r_{1}r_{2}r_{3}(N\vee T)}{\max(r_{1},r_{2},r_{3})p_{\min}}\|\Theta\|_{\max}^{2}\log(N+T+L).
	\]
	Thus, the proof of Lemma \ref{lem:restricted_convex} is completed.
	
	\subsubsection{Proof of Lemma \ref{lem:theta_hat_V}}
	\begin{proof}
		To prove (\ref{eq:error_W_V}), we notice that $\mathcal{M}_{(3)}(\Theta)\boldsymbol{V}=\boldsymbol{M}\mathcal{M}_{(3)}(\mathcal{S})$
		and therefore 
		\begin{align*}
			& \|(\boldsymbol{W}_{:,b}-\widehat{\boldsymbol{W}}_{:,b})^{\intercal}\mathcal{M}_{(3)}(\Theta)\boldsymbol{V}\|\\
			= & \|(\boldsymbol{W}_{:,b}-\widehat{\boldsymbol{W}}_{:,b})^{\intercal}\boldsymbol{M}\boldsymbol{S}\|\\
			= & \left\Vert \mathcal{M}_{(3)}(\mathcal{S})_{b,:}-\frac{\sum_{l=1}^{L}\mathcal{M}_{(3)}(\mathcal{S})_{(z)_{l},:}\mathbf{1}(\widehat{z}_{l}=b)}{\sum_{l=1}^{L}\mathbf{1}(\widehat{z}_{l}=b)}\right\Vert \\
			= & \left\Vert \frac{1}{\sum_{l=1}^{L}\mathbf{1}(\widehat{z}_{l}=b)}\sum_{l=1}^{L}\sum_{b'\in[r_{3}]/b}\mathbf{1}(z_{l}=b,\widehat{z}_{l}=b')\{\mathcal{M}_{(3)}(\mathcal{S})_{b,:}-\mathcal{M}_{(3)}(\mathcal{S})_{b',:}\}\right\Vert \\
			\leq & C_{0}\frac{r_{3}}{L\Delta_{\min}}\sum_{l=1}^{L}\sum_{b'\in[r_{3}]/b}\mathbf{1}(z_{l}=b,\widehat{z}_{l}=b')\|\mathcal{M}_{(3)}(\mathcal{S})_{b,:}-\mathcal{M}_{(3)}(\mathcal{S})_{b',:}\|^{2}\\
			\leq & C_{0}\frac{r_{3}g(\widehat{z},z)}{\Delta_{\min}}.
		\end{align*}
		Also, we know that 
		\begin{align}
			\|\boldsymbol{W}-\widehat{\boldsymbol{W}}\| & \leq\{\lambda_{r_{3}}(\boldsymbol{M})\}^{-1}\cdot\|\boldsymbol{M}^{\intercal}\boldsymbol{W}-\boldsymbol{M}^{\intercal}\widehat{\boldsymbol{W}}\|\nonumber \\
			& \leq\{\lambda_{r_{3}}(\boldsymbol{M})\}^{-1}\cdot\|\boldsymbol{M}^{\intercal}\boldsymbol{W}-\boldsymbol{M}^{\intercal}\widehat{\boldsymbol{W}}\|\nonumber \\
			& \leq\{\lambda_{r_{3}}(\boldsymbol{M})\}^{-1}\cdot\|\boldsymbol{I}-\boldsymbol{M}^{\intercal}\widehat{\boldsymbol{W}}\|_{F}\nonumber \\
			& \overset{(\ref{eq:eigen_M})}{\leq}\sqrt{\frac{r_{3}}{L}}\|\boldsymbol{I}-\boldsymbol{M}^{\intercal}\widehat{\boldsymbol{W}}\|_{F},\label{eq:W-What}
		\end{align}
		where $\boldsymbol{M}^{\intercal}\boldsymbol{W}=\boldsymbol{I}$ by
		definition. For any $b\in[r_{3}]$, denote $\widehat{n}_{b}=\sum_{l=1}^{L}\boldsymbol{1}\{\widehat{z}_{l}=b\}$.
		For any $b\neq b'$, we have 
		\[
		(\boldsymbol{M}^{\intercal}\widehat{\boldsymbol{W}})_{b'b}=\frac{\sum_{l=1}^{L}\boldsymbol{1}\{z_{l}=b',\widehat{z}_{l}=b\}}{\widehat{n}_{b}},\quad\delta_{b}=\sum_{b'\in[r_{3}]/b}(\boldsymbol{M}^{\intercal}\widehat{\boldsymbol{W}})_{b'b}=1-(\boldsymbol{M}^{\intercal}\widehat{\boldsymbol{W}})_{bb}.
		\]
		Therefore, 
		\begin{align*}
			\|\boldsymbol{I}-\boldsymbol{M}^{\intercal}\widehat{\boldsymbol{W}}\|_{F} & =\sqrt{\sum_{b\in[r_{3}]}\left\{ \delta_{b}^{2}+\sum_{b'\in[r_{3}]/b}(\boldsymbol{M}^{\intercal}\widehat{\boldsymbol{W}})_{b'b}^{2}\right\} }\\
			& \leq\sqrt{\sum_{b\in[r_{3}]}\left\{ \delta_{b}^{2}+\left(\sum_{b'\in[r_{3}]/b}(\boldsymbol{M}^{\intercal}\widehat{\boldsymbol{W}})_{b'b}\right)^{2}\right\} }\\
			& \leq\sqrt{2\sum_{b\in[r_{3}]}\delta_{b}^{2}}\\
			& \leq\sqrt{2}\sum_{b\in[r_{3}]}\delta_{b}\\
			& \leq\sqrt{2}\sum_{b\in[r_{3}]}\frac{\sum_{l=1}^{L}\boldsymbol{1}\{z_{l}\neq b',\widehat{z}_{l}=b\}}{\widehat{n}_{b}}\\
			& \leq\sqrt{2}\max_{b\in[r_{3}]}(\widehat{n}_{b})^{-1}\sum_{l=1}^{L}\mathbf{1}(z_{l}\neq\widehat{z}_{l})\\
			& \lesssim\frac{r_{3}}{L}\cdot Lh(\widehat{z},z)\lesssim\frac{r_{3}g(\widehat{z},z)}{\Delta_{\min}^{2}},
		\end{align*}
		where the last two inequalities are justified by (\ref{eq:eigen_W}).
		Then, we can show that 
		\begin{align*}
			\|(\boldsymbol{W}_{:,b}-\widehat{\boldsymbol{W}}_{:,b})^{\intercal}\mathcal{M}_{(3)}(\widehat{\Theta}-\Theta)\boldsymbol{V}\| & \leq\|\boldsymbol{W}_{:,b}-\widehat{\boldsymbol{W}}_{:,b}\|\cdot\|\mathcal{M}_{(3)}(\widehat{\Theta}-\Theta)\boldsymbol{V}\|_{\max}\\
			& \overset{(\ref{eq:W-What})}{\leq}C_{1}\frac{r_{3}^{3/2}g(\widehat{z},z)}{\Delta_{\min}^{2}\sqrt{L}}\cdot\|\mathcal{M}_{(3)}(\widehat{\Theta}-\Theta)\|_{F}\cdot\|\boldsymbol{V}\|_{2,\max}\\
			& \leq C_{1}\frac{\mu_{0}r_{1}^{1/2}r_{2}^{1/2}r_{3}^{3/2}g(\widehat{z},z)\|\widehat{\Theta}-\Theta\|_{F}}{\Delta_{\min}^{2}\sqrt{NTL}}.
		\end{align*}
		By triangle inequality, it completes the proof for (\ref{eq:error_W_V})
		\begin{align*}
			\|(\boldsymbol{W}_{:,b}-\widehat{\boldsymbol{W}}_{:,b})^{\intercal}\mathcal{M}_{(3)}(\widehat{\Theta})\boldsymbol{V}\| & \leq\|(\boldsymbol{W}_{:,b}-\widehat{\boldsymbol{W}}_{:,b})^{\intercal}\mathcal{M}_{(3)}(\Theta)\boldsymbol{V}\|\\
			& +\|(\boldsymbol{W}_{:,b}-\widehat{\boldsymbol{W}}_{:,b})^{\intercal}\mathcal{M}_{(3)}(\widehat{\Theta}-\Theta)\boldsymbol{V}\|\\
			& \lesssim\frac{r_{3}g(\widehat{z},z)}{\Delta_{\min}}+\frac{\mu_{0}r_{1}^{1/2}r_{2}^{1/2}r_{3}^{3/2}g(\widehat{z},z)\|\widehat{\Theta}-\Theta\|_{F}}{\Delta_{\min}^{2}\sqrt{NTL}}.
		\end{align*}
		To prove (\ref{eq:error_W_Vhat}), 
		\begin{align*}
			& \|(\boldsymbol{W}_{:,b}-\widehat{\boldsymbol{W}}_{:,b})^{\intercal}\mathcal{M}_{(3)}(\widehat{\Theta})\widehat{\boldsymbol{V}}\|\\
			& \leq\|(\boldsymbol{W}_{:,b}-\widehat{\boldsymbol{W}}_{:,b})^{\intercal}\mathcal{M}_{(3)}(\widehat{\Theta})\boldsymbol{V}\|\\
			& +\|(\boldsymbol{W}_{:,b}-\widehat{\boldsymbol{W}}_{:,b})^{\intercal}\mathcal{M}_{(3)}(\widehat{\Theta})(\boldsymbol{V}-\widehat{\boldsymbol{V}})\|\\
			& \overset{(\ref{eq:error_W_V})}{\lesssim}\frac{r_{3}g(\widehat{z},z)}{\Delta_{\min}}+\frac{\mu_{0}r_{1}^{1/2}r_{2}^{1/2}r_{3}^{3/2}g(\widehat{z},z)\|\widehat{\Theta}-\Theta\|_{F}}{\Delta_{\min}^{2}\sqrt{NTL}}\\
			& +\|(\boldsymbol{W}_{:,b}-\widehat{\boldsymbol{W}}_{:,b})^{\intercal}\mathcal{M}_{(3)}(\Theta)(\boldsymbol{V}-\widehat{\boldsymbol{V}})\|\\
			& +\|(\boldsymbol{W}_{:,b}-\widehat{\boldsymbol{W}}_{:,b})^{\intercal}\mathcal{M}_{(3)}(\widehat{\Theta}-\Theta)(\boldsymbol{V}-\widehat{\boldsymbol{V}})\|\\
			& \lesssim\frac{r_{3}g(\widehat{z},z)}{\Delta_{\min}}+\frac{\mu_{0}r_{1}^{1/2}r_{2}^{1/2}r_{3}^{3/2}g(\widehat{z},z)\|\widehat{\Theta}-\Theta\|_{F}}{\Delta_{\min}^{2}\sqrt{NTL}}\\
			& +\|\boldsymbol{W}_{:,b}-\widehat{\boldsymbol{W}}_{:,b}\|\cdot\|\mathcal{M}_{(3)}(\Theta)\|\cdot\|\boldsymbol{V}-\widehat{\boldsymbol{V}}\|\\
			& +\|\boldsymbol{W}_{:,b}-\widehat{\boldsymbol{W}}_{:,b}\|\cdot\|\mathcal{M}_{(3)}(\widehat{\Theta}-\Theta)\|_{F}\cdot\|\boldsymbol{V}-\widehat{\boldsymbol{V}}\|_{2,\max}\\
			& \overset{(\ref{eq:W-What}),(\ref{eq:Vhat-V})}{\lesssim}\frac{r_{3}g(\widehat{z},z)}{\Delta_{\min}}+\frac{\mu_{0}r_{1}^{1/2}r_{2}^{1/2}r_{3}^{3/2}g(\widehat{z},z)\|\widehat{\Theta}-\Theta\|_{F}}{\Delta_{\min}^{2}\sqrt{NTL}}+\frac{\kappa^{2}r_{3}^{3/2}g(\widehat{z},z)\|\widehat{\Theta}-\Theta\|_{F}}{\Delta_{\min}^{2}\sqrt{L}}.
		\end{align*}
		To prove (\ref{eq:W_error_V}), 
		\begin{align*}
			\|\widehat{\boldsymbol{W}}_{:,b}^{\intercal}\mathcal{M}_{(3)}(\widehat{\Theta})(\boldsymbol{V}-\widehat{\boldsymbol{V}})\| & \leq\|\widehat{\boldsymbol{W}}_{:,b}^{\intercal}\mathcal{M}_{(3)}(\Theta)(\boldsymbol{V}-\widehat{\boldsymbol{V}})\|\\
			& +\|\widehat{\boldsymbol{W}}_{:,b}^{\intercal}\mathcal{M}_{(3)}(\widehat{\Theta}-\Theta)(\boldsymbol{V}-\widehat{\boldsymbol{V}})\|\\
			& \leq\|\boldsymbol{W}_{:,b}^{\intercal}\mathcal{M}_{(3)}(\Theta)(\boldsymbol{V}-\widehat{\boldsymbol{V}})\|+\|(\widehat{\boldsymbol{W}}_{:,b}^{\intercal}-\boldsymbol{W}_{:,b}^{\intercal})\mathcal{M}_{(3)}(\Theta)(\boldsymbol{V}-\widehat{\boldsymbol{V}})\|\\
			& +\|\boldsymbol{W}_{:,b}^{\intercal}\mathcal{M}_{(3)}(\widehat{\Theta}-\Theta)(\boldsymbol{V}-\widehat{\boldsymbol{V}})\|+\|(\widehat{\boldsymbol{W}}_{:,b}^{\intercal}-\boldsymbol{W}_{:,b}^{\intercal})\mathcal{M}_{(3)}(\widehat{\Theta}-\Theta)(\boldsymbol{V}-\widehat{\boldsymbol{V}})\|\\
			& \lesssim\sqrt{r_{3}/L}\cdot\|\mathcal{M}_{(3)}(\Theta)\|\cdot\|\boldsymbol{V}-\widehat{\boldsymbol{V}}\|+\frac{r_{3}^{3/2}g(\widehat{z},z)}{\Delta_{\min}^{2}\sqrt{L}}\cdot\|\mathcal{M}_{(3)}(\Theta)\|\cdot\|\boldsymbol{V}-\widehat{\boldsymbol{V}}\|\\
			& +\sqrt{r_{3}/L}\cdot\|\mathcal{M}_{(3)}(\widehat{\Theta}-\Theta)\|_{F}\cdot\|\boldsymbol{V}-\widehat{\boldsymbol{V}}\|_{2,\max}\\
			& +\frac{r_{3}^{3/2}g(\widehat{z},z)}{\Delta_{\min}^{2}\sqrt{L}}\cdot\|\mathcal{M}_{(3)}(\widehat{\Theta}-\Theta)\|_{F}\cdot\|\boldsymbol{V}-\widehat{\boldsymbol{V}}\|_{2,\max}\\
			& \overset{(\ref{eq:Vhat-V})}{\lesssim}\sqrt{r_{3}/L}\cdot\kappa^{2}\|\widehat{\Theta}-\Theta\|_{F}+\frac{r_{3}^{3/2}g(\widehat{z},z)}{\Delta_{\min}^{2}\sqrt{L}}\cdot\kappa^{2}\|\widehat{\Theta}-\Theta\|_{F}\\
			& +\frac{\mu_{0}r_{1}^{1/2}r_{2}^{1/2}r_{3}^{1/2}}{\sqrt{NTL}}\|\widehat{\Theta}-\Theta\|_{F}+\frac{\mu_{0}r_{1}^{1/2}r_{2}^{1/2}r_{3}^{3/2}g(\widehat{z},z)}{\Delta_{\min}^{2}\sqrt{NTL}}\|\widehat{\Theta}-\Theta\|_{F}\\
			& \lesssim\kappa^{2}\sqrt{r_{3}/L}\|\widehat{\Theta}-\Theta\|_{F}.
		\end{align*}
		To prove (\ref{eq:Vhat-V}), we have 
		\begin{align*}
			\|\widehat{\boldsymbol{V}}-\boldsymbol{V}\| & =\|\boldsymbol{\widehat{U}}_{1}\otimes\widehat{\boldsymbol{U}}_{2}-\boldsymbol{U}_{1}\otimes\boldsymbol{U}_{2}\|\\
			& =\|\boldsymbol{\widehat{U}}_{1}\otimes(\widehat{\boldsymbol{U}}_{2}-\boldsymbol{U}_{2})+(\boldsymbol{\widehat{U}}_{1}-\boldsymbol{U}_{1})\otimes\boldsymbol{U}_{2}\|\\
			& \leq\|\boldsymbol{\widehat{U}}_{1}\otimes(\widehat{\boldsymbol{U}}_{2}-\boldsymbol{U}_{2})\|+\|(\boldsymbol{\widehat{U}}_{1}-\boldsymbol{U}_{1})\otimes\boldsymbol{U}_{2}\|\\
			& \lesssim\left\{ \|\widehat{\boldsymbol{U}}_{1}-\boldsymbol{U}_{1}\|+\|\widehat{\boldsymbol{U}}_{2}-\boldsymbol{U}_{2}\|\right\} \overset{\text{Lemma}\ref{lem:Davis-Kahan}}{\lesssim}\frac{\kappa\|\widehat{\Theta}-\Theta\|_{F}}{\lambda_{\min}}.
		\end{align*}
		Since $\boldsymbol{U}_{1}$ and $\boldsymbol{U}_{2}$ are only identifiable
		up to rotation and permutation, we instead focus on the upper bound
		of $\min_{\boldsymbol{R}_{1}\in\mathbb{O}_{r_{1}}}\|\widehat{\boldsymbol{U}}_{1}-\boldsymbol{U}_{1}\boldsymbol{R}_{1}\|$
		and $\min_{\boldsymbol{R}_{2}\in\mathbb{O}_{r_{2}}}\|\widehat{\boldsymbol{U}}_{2}-\boldsymbol{U}_{2}\boldsymbol{R}_{2}\|$,
		where $\mathbb{O}_{r}$ is the collection of all $r$-by-$r$ matrices
		with orthonormal columns. Lemma \ref{lem:Davis-Kahan} is a variant
		of the Davis-Kahan $\sin(\Theta)$ Theorem from \citet{yu2015useful},
		which bound the distance between subspaces spanned by the population
		eigenvectors and their sample versions. 
		
		For (\ref{eq:Vhat-V}) and (\ref{eq:Vhat-V_2max}), the proof is similar
		to Theorem 4, \citet{yu2015useful}:
		\begin{align*}
			\|\boldsymbol{V}\|_{2,\max}^{2} & =\max_{j}\|e_{j}^{\intercal}(\boldsymbol{U}_{1}\otimes\boldsymbol{U}_{2})\|_{2}^{2}\\
			& \leq\max_{i,t}\|e_{i}^{\intercal}\boldsymbol{U}_{1}\|^{2}\cdot\|e_{t}^{\intercal}\boldsymbol{U}_{2}\|^{2}\text{ (Cauchy-Schwatz inequality)}\\
			& \leq\|\boldsymbol{U}_{1}\|_{2,\max}^{2}\|\boldsymbol{U}_{2}\|_{2,\max}^{2}\\
			& \leq\frac{\mu_{0}^{2}r_{1}r_{2}}{NT},
		\end{align*}
		and 
		\begin{align*}
			\|\boldsymbol{V}-\widehat{\boldsymbol{V}}\|_{2,\max} & =\max_{j}\|e_{j}^{\intercal}(\boldsymbol{U}_{1}\otimes\boldsymbol{U}_{2}-\widehat{\boldsymbol{U}}_{1}\otimes\widehat{\boldsymbol{U}}_{2})\|\\
			& =\max_{j}\|e_{j}^{\intercal}(\boldsymbol{U}_{1}\otimes\boldsymbol{U}_{2}-\boldsymbol{U}_{1}\otimes\widehat{\boldsymbol{U}}_{2}+\boldsymbol{U}_{1}\otimes\widehat{\boldsymbol{U}}_{2}-\widehat{\boldsymbol{U}}_{1}\otimes\widehat{\boldsymbol{U}}_{2})\|\\
			& \leq\max_{j}\|e_{j}^{\intercal}(\boldsymbol{U}_{1}\otimes\boldsymbol{U}_{2}-\boldsymbol{U}_{1}\otimes\widehat{\boldsymbol{U}}_{2})\|+\max_{j'}\|e_{j'}^{\intercal}(\boldsymbol{U}_{1}\otimes\widehat{\boldsymbol{U}}_{2}-\widehat{\boldsymbol{U}}_{1}\otimes\widehat{\boldsymbol{U}}_{2})\|\\
			& \leq\|\boldsymbol{U}_{1}\|_{2,\max}\|\widehat{\boldsymbol{U}}_{2}-\boldsymbol{U}_{2}\|_{2,\max}+\|\widehat{\boldsymbol{U}}_{2}\|_{2,\max}\|\widehat{\boldsymbol{U}}_{1}-\boldsymbol{U}_{1}\|_{2,\max}\\
			& \leq\sqrt{\frac{\mu_{0}r_{1}}{N}}\|\widehat{\boldsymbol{U}}_{2}-\boldsymbol{U}_{2}\|_{2,\max}+\sqrt{\frac{\mu_{0}r_{2}}{T}}\|\widehat{\boldsymbol{U}}_{1}-\boldsymbol{U}_{1}\|_{2,\max}.
		\end{align*}
		Under the assumption of eigengap, we are able to bound the perturbation
		of individual eigenvectors: $$\|\widehat{u}_{2,i}-u_{2,i}\|_{2}\lesssim\frac{\lambda_{\max}\|\widehat{\Theta}-\Theta\|_{F}}{\min(\lambda_{i-1}^{2}-\lambda_{i}^{2},\lambda_{i}^{2}-\lambda_{i+1}^{2})}.$$
		Assume for any $i\in[r_{2}]$, the interval $[\lambda_{i}-\delta_{r_{2}},\lambda_{i}+\delta_{r_{2}}]$
		does not contain any eigenvalues of $\mathcal{M}_{(2)}(\Theta)$ other
		than $\lambda_{i}$:
		$$\|\widehat{u}_{2,i}-u_{2,i}\|_{2}\lesssim\frac{\lambda_{\max}\|\widehat{\Theta}-\Theta\|_{F}}{\min(\lambda_{i-1}^{2}-\lambda_{i}^{2},\lambda_{i}^{2}-\lambda_{i+1}^{2})}\lesssim\frac{\kappa\|\widehat{\Theta}-\Theta\|_{F}}{\delta_{r_{2}}},$$
		which implies $\|\widehat{\boldsymbol{U}}_{2}-\boldsymbol{U}_{2}\|_{2,\max}\leq\kappa\|\widehat{\Theta}-\Theta\|_{F}/\delta_{r_{2}}$.
		Identical bounds also hold if $\widehat{u}_{2,i}$ and $u_{2,i}$
		are replaced with $\widehat{u}_{1,i}$ and $u_{1,i}$. Suppose 
		$$
		\delta_{r_{1}}\geq\sqrt{\frac{N}{\mu_{0}r_{1}}}\cdot\kappa\|\widehat{\Theta}-\Theta\|_{F},\quad\delta_{r_{2}}\geq\sqrt{\frac{T}{\mu_{0}r_{2}}}\cdot\kappa\|\widehat{\Theta}-\Theta\|_{F},
		$$
		we claim that $\|\boldsymbol{V}-\widehat{\boldsymbol{V}}\|_{2,\max}^{2}\leq\|\boldsymbol{V}\|_{2,\max}^{2}$.
	\end{proof}
	
	\subsubsection{Proof of Lemma \ref{lem:E1_decomposed}}
	
	First, we want to prove the probability $P\left(\langle\mathcal{M}_{(3)}(\widehat{\Theta}-\Theta)_{l,:}\boldsymbol{V},\boldsymbol{S}_{a,:}-\boldsymbol{S}_{b,:}\rangle\leq-\frac{1}{8}\|\boldsymbol{S}_{a,:}-\boldsymbol{S}_{b,:}\|^{2}\right)$
	is negligible. We can show that
	\begin{align*}
		& P\left(\langle\mathcal{M}_{(3)}(\widehat{\Theta}-\Theta)_{l,:}\boldsymbol{V},\boldsymbol{S}_{a,:}-\boldsymbol{S}_{b,:}\rangle\leq-\frac{1}{8}\|\boldsymbol{S}_{a,:}-\boldsymbol{S}_{b,:}\|^{2}\right)\\
		& \leq P(-\|\mathcal{M}_{(3)}(\widehat{\Theta}-\Theta)_{l,:}\|\cdot\|(\boldsymbol{S}_{a,:}-\boldsymbol{S}_{b,:})\boldsymbol{V}^{\intercal}\|_{\max}\leq-\frac{1}{8}\|\boldsymbol{S}_{a,:}-\boldsymbol{S}_{b,:}\|^{2})\\
		& \leq P(\|\mathcal{M}_{(3)}(\widehat{\Theta}-\Theta)_{l,:}\|\cdot\|\boldsymbol{S}_{a,:}-\boldsymbol{S}_{b,:}\|\cdot\|\boldsymbol{V}\|_{2,\max}\geq\frac{1}{8}\|\boldsymbol{S}_{a,:}-\boldsymbol{S}_{b,:}\|^{2})\\
		& \leq P(\|\mathcal{M}_{(3)}(\widehat{\Theta}-\Theta)_{l,:}\|\cdot\|\boldsymbol{U}_{1}\|_{2,\max}\|\boldsymbol{U}_{2}\|_{2,\max}\geq\frac{1}{8}\|\boldsymbol{S}_{a,:}-\boldsymbol{S}_{b,:}\|)\\
		& \leq P\left(\|\mathcal{M}_{(3)}(\widehat{\Theta}-\Theta)_{l,:}\|^{2}\geq\frac{1}{64}\frac{\|\boldsymbol{S}_{a,:}-\boldsymbol{S}_{b,:}\|^{2}}{\|\boldsymbol{U}_{1}\|_{2,\max}^{2}\|\boldsymbol{U}_{2}\|_{2,\max}^{2}}\right)\\
		& \leq P\left(\sum_{l=1}^{L}\|\mathcal{M}_{(3)}(\widehat{\Theta}-\Theta)_{l,:}\|^{2}\geq\frac{1}{64}\frac{L\Delta_{\min}^{2}}{\|\boldsymbol{U}_{1}\|_{2,\max}^{2}\|\boldsymbol{U}_{2}\|_{2,\max}^{2}}\right)\\
		& \leq P\left(\|\widehat{\Theta}-\Theta\|_{F}^{2}\geq\frac{1}{64}\frac{NTL\Delta_{\min}^{2}}{\mu_{0}^{2}r_{1}r_{2}}\right),
	\end{align*}
	where under Assumption R2):
	\[
	\|\boldsymbol{U}_{1}\|_{2,\max}^{2}\leq\frac{\mu_{0}r_{1}}{N},\quad\|\boldsymbol{U}_{2}\|_{2,\max}^{2}\leq\frac{\mu_{0}r_{2}}{T}.
	\]
	To ensure the event $\|\widehat{\Theta}-\Theta\|_{F}^{2}\leq NTL\Delta_{\min}^{2}/(64\mu_{0}^{2}r_{1}r_{2})$
	holds with high probability, we need to have 
	\begin{align}
		\frac{NTL\Delta_{\min}^{2}}{\mu_{0}^{2}r_{1}r_{2}} & \geq\frac{c_{0}(N\vee T)r_{1}r_{2}r_{3}\log(N+T+L)\|\Theta\|_{\max}^{2}}{p_{\min}^{2}\max(r_{1},r_{2},r_{3})},\label{eq:SNR_1}\\
		\frac{NTL\Delta_{\min}^{2}}{\mu_{0}^{2}r_{1}r_{2}} & \geq\frac{c_{0}w_{\max}^{2}(N\vee T)r_{1}r_{2}r_{3}\log^{2}(N+T+L)(L_{\alpha}^{2}/\gamma_{\alpha}^{2})}{w_{\min}^{2}p_{\min}^{2}\gamma_{\alpha}^{2}\max(r_{1},r_{2},r_{3})},\label{eq:SNR_2}
	\end{align}
	under the estimation error in Theorem \ref{thm:estimation}. Thus,
	we claim that 
	\[
	P\left(\|\widehat{\Theta}-\Theta\|_{F}^{2}\geq\frac{1}{64}\frac{NTL\Delta_{\min}^{2}}{\mu_{0}^{2}r_{1}r_{2}}\right)\leq\frac{c}{(N+T+L)^{2}},
	\]
	under the conditions (\ref{eq:SNR_1}) and (\ref{eq:SNR_2}). Next,
	we show that $P\left(\langle\mathcal{M}_{(3)}(\widehat{\Theta}-\Theta)_{l,:}\boldsymbol{V},\tilde{\boldsymbol{S}}_{a,:}-\boldsymbol{S}_{a,:}\rangle\leq-\frac{1}{8}\|\boldsymbol{S}_{a,:}-\boldsymbol{S}_{b,:}\|^{2}\right)$
	is negligible
	\begin{align}
		& P\left(\langle\mathcal{M}_{(3)}(\widehat{\Theta}-\Theta)_{l,:}\boldsymbol{V},\tilde{\boldsymbol{S}}_{a,:}-\boldsymbol{S}_{a,:}\rangle\leq-\frac{1}{8}\|\boldsymbol{S}_{a,:}-\boldsymbol{S}_{b,:}\|^{2}\right)\nonumber \\
		& \leq P(-\|\mathcal{M}_{(3)}(\widehat{\Theta}-\Theta)_{l,:}\|\cdot\|(\tilde{\boldsymbol{S}}_{a,:}-\boldsymbol{S}_{a,:})\boldsymbol{V}^{\intercal}\|_{\max}\leq-\frac{1}{8}\|\boldsymbol{S}_{a,:}-\boldsymbol{S}_{b,:}\|^{2})\nonumber \\
		& \leq P(\|\mathcal{M}_{(3)}(\widehat{\Theta}-\Theta)_{l,:}\|\cdot\|\tilde{\boldsymbol{S}}_{a,:}-\boldsymbol{S}_{a,:}\|\cdot\|\boldsymbol{V}\|_{2,\max}\geq\frac{1}{8}\|\boldsymbol{S}_{a,:}-\boldsymbol{S}_{b,:}\|^{2})\nonumber \\
		& \leq P(\|\mathcal{M}_{(3)}(\widehat{\Theta}-\Theta)_{l,:}\|\cdot\|\tilde{\boldsymbol{S}}_{a,:}-\boldsymbol{S}_{a,:}\|\cdot\|\boldsymbol{U}_{1}\|_{2,\max}\|\boldsymbol{U}_{2}\|_{2,\max}\geq\frac{1}{8}\|\boldsymbol{S}_{a,:}-\boldsymbol{S}_{b,:}\|).\label{eq:error_2}
	\end{align}
	One can observe that 
	\begin{align*}
		\|\tilde{\boldsymbol{S}}_{a,:}-\boldsymbol{S}_{a,:}\| & =\|\boldsymbol{M}_{:,a}^{\intercal}\mathcal{M}_{(3)}(\widehat{\Theta}-\Theta)\boldsymbol{V}\|\\
		& \leq\|\boldsymbol{M}_{:,a}^{\intercal}\mathcal{M}_{(3)}(\widehat{\Theta}-\Theta)\|\cdot\|\boldsymbol{V}\|_{\max}\\
		& \leq\|\boldsymbol{M}_{:,a}^{\intercal}\|\cdot\|\mathcal{M}_{(3)}(\widehat{\Theta}-\Theta)\|_{F}\cdot\max_{i,j}\|e_{i}^{\intercal}\boldsymbol{V}e_{j}\|\\
		& \leq\sqrt{\frac{r_{3}}{L}}\|\boldsymbol{U}_{1}\|_{2,\max}\|\boldsymbol{U}_{2}\|_{2,\max}\cdot\|\mathcal{M}_{(3)}(\widehat{\Theta}-\Theta)\|_{F}.
	\end{align*}
	Plug it back to (\ref{eq:error_2}), we obtain 
	\begin{align*}
		& P\left(\langle\mathcal{M}_{(3)}(\widehat{\Theta}-\Theta)_{l,:}\boldsymbol{V},\tilde{\boldsymbol{S}}_{a,:}-\boldsymbol{S}_{a,:}\rangle\leq-\frac{1}{8}\|\boldsymbol{S}_{a,:}-\boldsymbol{S}_{b,:}\|^{2}\right)\\
		& \leq P\left(\|\mathcal{M}_{(3)}(\widehat{\Theta}-\Theta)_{l,:}\|\|\mathcal{M}_{(3)}(\widehat{\Theta}-\Theta)\|_{F}\geq\frac{\|\boldsymbol{S}_{a,:}-\boldsymbol{S}_{b,:}\|^{2}}{8\sqrt{\frac{r_{3}}{L}}\|\boldsymbol{U}_{1}\|_{2,\max}^{2}\|\boldsymbol{U}_{2}\|_{2,\max}^{2}}\right)\\
		& \leq P\left(\sum_{l=1}^{L}\|\mathcal{M}_{(3)}(\widehat{\Theta}-\Theta)_{l,:}\|\|\mathcal{M}_{(3)}(\widehat{\Theta}-\Theta)\|_{F}\geq\frac{L\|\boldsymbol{S}_{a,:}-\boldsymbol{S}_{b,:}\|^{2}}{8\sqrt{\frac{r_{3}}{L}}\|\boldsymbol{U}_{1}\|_{2,\max}^{2}\|\boldsymbol{U}_{2}\|_{2,\max}^{2}}\right)\\
		& \leq P\left(\|\widehat{\Theta}-\Theta\|_{F}^{2}\geq\frac{NTL^{3/2}\Delta_{\min}^{2}}{8\mu_{0}^{2}r_{1}r_{2}\sqrt{r_{3}}}\right),
	\end{align*}
	which is negligible implied by the signal-to-noise ratio condition
	since $L\geq r_{3}$: 
	\begin{align*}
		\frac{NTL^{3/2}\Delta_{\min}^{2}}{\mu_{0}^{2}r_{1}r_{2}\sqrt{r_{3}}} & \geq\frac{NTL\Delta_{\min}^{2}}{\mu_{0}^{2}r_{1}r_{2}}\geq\frac{c_{0}L_{\alpha}^{2}(N\vee T)r_{1}r_{2}r_{3}(\|\Theta\|_{\max}^{2}\vee\sigma^{2})\log^{2}(N+T+L)}{p_{\min}^{4}\gamma_{\alpha}^{2}\max(r_{1},r_{2},r_{3})}.
	\end{align*}
	Thus, which completes our proof of Lemma \ref{lem:E1_decomposed}.
	\begin{lemma}
		\label{lem:Davis-Kahan} Let 
		\begin{align*}
			\lambda_{\max} & =\max\{\|\mathcal{M}_{(1)}(\mathcal{Y})\|,\|\mathcal{M}_{(2)}(\mathcal{Y})\|,\|\mathcal{M}_{(3)}(\mathcal{Y})\|\},\\
			\lambda_{\min} & =\min\left[\lambda_{r_{1}}\{\mathcal{M}_{(1)}(\mathcal{Y})\},\lambda_{r_{2}}\{\mathcal{M}_{(2)}(\mathcal{Y})\},\lambda_{r_{3}}\{\mathcal{M}_{(3)}(\mathcal{Y})\}\right],
		\end{align*}
		and $\kappa=\lambda_{\max}/\lambda_{\min}$, then we have: 
		
		\[
		\min_{\boldsymbol{R}_{1}\in\mathbb{O}_{r_{1}}}\|\widehat{\boldsymbol{U}}_{1}-\boldsymbol{U}_{1}\boldsymbol{R}_{1}\|_{F}\leq C\frac{\kappa\|\widehat{\Theta}-\Theta\|_{F}}{\lambda_{\min}},\quad\min_{\boldsymbol{R}_{2}\in\mathbb{O}_{r_{2}}}\|\widehat{\boldsymbol{U}}_{2}-\boldsymbol{U}_{2}\boldsymbol{R}_{2}\|\leq C\frac{\kappa\|\widehat{\Theta}-\Theta\|_{F}}{\lambda_{\min}}.
		\]
	\end{lemma}
	
	The proof of Lemma \ref{lem:Davis-Kahan} is provided in \citet{yu2015useful},
	Theorem 4.
	\begin{lemma}
		\label{lemma:upper_lower_MW} Under condition (\ref{eq:loss_condition}),
		we have $r_{3}/L\lesssim|l\in[L]:(\widehat{z})_{l}=a|\lesssim r_{3}/L$
		for any $a\in[r_{3}]$. Moreover, 
		\begin{align*}
			& \sqrt{L/r_{3}}\lesssim\lambda_{r_{3}}(\boldsymbol{M})\leq\|\boldsymbol{M}\|\lesssim\sqrt{L/r_{3}},\\
			& \sqrt{r_{3}/L}\lesssim\lambda_{r_{3}}(\boldsymbol{W})\leq\|\boldsymbol{W}\|\lesssim\sqrt{r_{3}/L}.
		\end{align*}
		The above inequalities also hold by replacing $\boldsymbol{M}$, $\boldsymbol{W}$
		with $\widehat{\boldsymbol{M}}$ and $\widehat{\boldsymbol{W}}$,
		respectively. 
	\end{lemma}
	
	The proof of Lemma \ref{lemma:upper_lower_MW} is provided in \citet{han2022exact},
	Lemma 4.
	
\end{document}